\documentclass{article}

\usepackage{microtype}
\usepackage{graphicx}
\usepackage{subfigure}
\usepackage{booktabs} 

\usepackage[citecolor=green]{hyperref}
\usepackage{url}
\usepackage{graphicx}
\usepackage{subfigure}
\usepackage{soul}
\newcommand{\ie}{\textit{i.e.}}
\newcommand{\eg}{\textit{e.g.}}

\newcommand{\etc}{\textit{etc}}
\usepackage{multirow}
\newtheorem{theorem}{Theorem}

\soulregister\cite7
\soulregister\citep7
\soulregister\citet7
\soulregister\ref7
\soulregister\pageref7

\usepackage{array}
\newcolumntype{+}{>{\global\let\currentrowstyle\relax}}
\newcolumntype{^}{>{\currentrowstyle}}
\newcommand{\rowstyle}[1]{\gdef\currentrowstyle{#1}#1\ignorespaces}

\usepackage{amsmath,amsfonts,bm}

\def\eqref#1{equation~\ref{#1}}

\def\1{\bm{1}}

\DeclareMathAlphabet{\mathsfit}{\encodingdefault}{\sfdefault}{m}{sl}
\SetMathAlphabet{\mathsfit}{bold}{\encodingdefault}{\sfdefault}{bx}{n}

\def\bx{\mathbf{x}}

\def\X{\mathcal{X}}

\def\Z{\mathcal{Z}}

\def\P{\mathcal{P}}

\newcommand{\comment}[1]{}

\usepackage{hyperref}

\usepackage[accepted]{icml2020}

\icmltitlerunning{Effective and Robust Detection of Adversarial Examples via Benford-Fourier Coefficients} 

\begin{document}

\twocolumn[
\icmltitle{Effective and Robust Detection of Adversarial Examples via \\ Benford-Fourier Coefficients}



\icmlsetsymbol{equal}{*}

\begin{icmlauthorlist}
\icmlauthor{Chengcheng Ma}{1,2}
\icmlauthor{Baoyuan Wu}{3}
\icmlauthor{Shibiao Xu}{1,2}
\icmlauthor{Yanbo Fan}{3}
\icmlauthor{Yong Zhang}{3}
\icmlauthor{Xiaopeng Zhang}{1,2}
\icmlauthor{Zhifeng Li}{3}
\end{icmlauthorlist}

\icmlaffiliation{1}{School of Artificial Intelligence, Chinese Academy of Sciences and University of Chinese Academy of Sciences}
\icmlaffiliation{2}{National Laboratory of Pattern Recognition, Institute of Automation, Chinese Academy of Sciences}
\icmlaffiliation{3}{Tencent AI Lab, China}

\icmlcorrespondingauthor{Shibiao Xu}{shibiao.xu@nlpr.ia.ac.cn}
\icmlcorrespondingauthor{Xiaopeng Zhang}{xiaopeng.zhang@ia.ac.cn}

\icmlkeywords{Machine Learning, ICML}

\vskip 0.3in
]



\printAffiliationsAndNotice{\icmlEqualContribution} 

\begin{abstract}
Adversarial examples have been well known as a serious threat to deep neural networks (DNNs). \comment{To ensure successful and safe operations of DNNs on real-world tasks, it is urgent to equip DNNs with effective defense strategies.}In this work, we study the detection of adversarial examples, based on the assumption that the output and internal responses of one DNN model for both adversarial and benign examples follow the generalized Gaussian distribution (GGD), but with different parameters (\ie, shape factor, mean, and variance). GGD is a general distribution family to cover many popular distributions (\eg, Laplacian, Gaussian, or uniform). It is more likely to approximate the intrinsic distributions of internal responses than any specific distribution.
  Besides, since the shape factor is more robust to different databases rather than the other two parameters, we propose to construct discriminative features via the shape factor for adversarial detection, employing the magnitude of Benford-Fourier coefficients (MBF), which can be easily estimated using responses.
  Finally, a support vector machine is trained as the adversarial detector through leveraging the MBF features.
  \comment{\hl{Through the Kolmogorov-Smirnov (KS) test \cite{ks-test-1951}, We empirically verify that: 1) the posterior vectors of both adversarial and benign examples follow GGD; 2) the extracted MBF features of adversarial and benign examples follow different distributions.} }
  Extensive experiments in terms of image classification demonstrate that the proposed detector is much more effective and robust on detecting adversarial examples of different crafting methods and different sources, compared to state-of-the-art adversarial detection methods.
\end{abstract}

\vspace{-2em}
\section{Introduction}
\label{submission}

Deep neural networks (DNNs) have achieved a remarkable success in many important applications, such as image classification, face recognition, object detection, \etc.
In the meanwhile, DNNs have been shown to be very vulnerable to adversarial examples.
However, many real-world scenarios have very restrictive requirements about the robustness of DNNs, such as face verification for login, or semantic segmentation in autonomous driving.
Adversarial examples are a serious threat to the applications of DNNs to these important tasks.
Since many kinds of adversarial attack methods have been proposed to fool DNNs, it is more urgent to equip effective defensive strategies to ensure the safety of deep models in real-world applications.
However, defense seems to be more challenging than attack, as it has to face adversarial examples from unknown crafting methods and unknown data sources.
Typical defensive strategies include adversarial training, adversarial de-noising, and adversarial detection. Compared to the former two strategies, adversarial detection is somewhat more cost-effective, as it often needs no re-training or modifying the original model.

There are two main challenges for adversarial detection.
{\bf (1)} The adversarial examples are designed to camouflage themselves to be close to the corresponding benign examples in the input space. Then, where and how to extract the discriminative information to train the detector?
{\bf (2)} The data sources and the generating methods of adversarial examples are often inaccessible to the detector. In this case, the detector can be stably effective across different data sources and different attack methods?
In other words, a good adversarial detector is required to be not only {\it effective} to distinguish adversarial and benign examples, but also {\it robust} to different data sources and attack methods.

To satisfy the first requirement of effectiveness, we utilize the other principle of crafting adversarial examples that the outputs between benign and adversarial examples should be large, to encourage the change of the final prediction. It means that the imperceptible difference between benign and adversarial examples in the input space is enlarged along the DNN model, leading to the significant difference in the output space.
Inspired by this fact, we assume that the output or the responses of internal layers of the DNN model should include the discriminative information for benign and adversarial examples.
A few works have attempted to extract different types of discriminative features from the output or the internal responses, such as kernel density (KD)  \cite{KD-BU-arxiv-2017} and the local intrinsic dimensionality (LID)  \cite{LID-iclr-2018}, \etc.
To achieve the robustness, the extracted discriminative features should model the intrinsic difference between adversarial and benign examples, rather than the difference from the changes of data sources or attack methods.
Many existing methods have shown the effectiveness to some extent on detecting adversarial examples of specific data sources and attack methods. However, their robustness, especially across different data sources, has not been well studied and verified.

In this work, we propose a novel detection method based on the assumption that the internal responses of both adversarial and benign examples follow the generalized Gaussian distribution (GGD) \cite{GGD-1989}, but with different parameters, including the shape factor, mean, and variance.
The rationale behind this assumption is that GGD covers many popular distributions with varied shape factors (such as Laplacian, Gaussian, or uniform distribution), such that GGD is more likely to approximate the intrinsic response distributions rather than one specific distribution.
Moreover, mean and variance of GGD may vary significantly with respect to different classes and data sources, even for benign examples, while the shape factor is more robust. For example, the mean and variance of two Gaussian distributions could be totally different, but their shape factors are the same (\ie, 2). Thus, we propose to use the shape factor as an effective and robust discrimination between adversarial and benign examples.
However, it is difficult to exactly estimate the shape factor in practice.
We resort to the magnitude of Benford-Fourier coefficients  \cite{bf-icip-2014}, which is a function of the shape factor. It can be easily estimated using internal responses, according to the definition of Fourier transform.
Then, the magnitudes estimated from internal responses of different convolutional layers are concatenated as a novel representation.
Finally, a support vector machine (SVM) \cite{svm-2013} is trained using the new representations as the adversarial detector.
Extensive experiments carried out on several databases verify the effectiveness and robustness of the proposed detection method. 
To further verify the rationale of our assumption, we present the empirical analysis through the Kolmogorov-Smirnov test (KS test) \cite{ks-test-1951}.
The KS test verifies that 1) the posterior vectors of both adversarial and benign examples predicted by the CNN model follow the distribution of GGD, but with different parameters, and 2) the MBF features of adversarial and benign examples follow different distributions, and the MBF features of adversarial examples crafted from different attack methods follow the same distribution, as well as that the MBF features of adversarial/benign examples from different data sources follow the same distribution.

Moreover, we visualize the statistics (\ie, mean $\pm$ standard deviation) of the extracted MBF features for adversarial and benign examples. The visualization reveals the distinct difference between adversarial and benign examples.
These empirical analysis demonstrates the effectiveness and robustness of the proposed MBF detector.

\vspace{-1em}
\section{Related work}
\label{Related work}
\vspace{-.2em}

The general idea of most existing detection methods is learning or constructing a new representation to discriminate adversarial and benign examples, utilizing the outputs or immediate responses of an original classification network.
\cite{detection-pca-iccv-2017} trained a cascading classifier based on the principal component analysis (PCA) \cite{pca-1901} statistics of responses from each convolutional layer of the defended convolutional neural network (CNN) model. An example is recognized as benign if all single classifiers of the cascade predict it as benign, otherwise adversarial.
\cite{safetynet-iccv-2017} proposed SafetyNet by adding a RBF-SVM classifier to detect adversarial examples, at the end of the original classification network.
\cite{subnetwork-iclr-2017} proposed a detection network along with an original classification network, which takes the internal responses of the original network as inputs. It shows effectiveness on detecting adversarial examples generated by simple attacks (such as FGSM \cite{fgsm} and JSMA \cite{jsma}), while performs much worse when facing more advanced attacks (such as C\&W \cite{CW}). It tells that this method is sensitive to attack methods.
\cite{detection-statistical-2017} attempted to detect adversarial examples using the statistical test of maximum mean discrepancy (MMD).
Although above detection methods show effectiveness on some attack methods and some databases, but a thorough evaluation presented in \cite{bypass-10-detection-2017} has shown that these methods are sensitive to attack methods or databases, and they can be somewhat easily invaded by new attacks.

Some recent works proposed to utilize neighboring samples in the same database to construct a better representation of a current sample.
\cite{KD-BU-arxiv-2017} defined two metrics based on the responses of the final hidden layer of the classification neural network, including kernel density estimation (KDE) and Bayesian neural network uncertainty (BU). If the metric score of KDE/BU is lower/higher than a pre-defined threshold, then the example is predicted as adversarial.
\cite{LID-iclr-2018} utilized the local intrinsic dimensionality (LID) to measure the characterization of adversarial regions of DNNs. LID describes the distance between one example and its $k$-nearest neighboring sample in the feature space of immediate responses of the original classification network. The distances computed from different layers are concatenated as the example representation, which is then used to train a shallow classifier to discriminate adversarial and benign examples.
\cite{I-defender-nips-2018} defined the intrinsic hidden state distribution (IHSD) of the responses of the original classification network to model different classes.
The Gaussian mixture model (GMM) was used to approximate IHSD of each class. Then, the posterior probability of one sample assigned to GMM is computed as the metric. If the probability is lower than a pre-defined threshold, then it is recognized as adversarial.
\cite{M-distance-nips-2018} computed the class-conditional Gaussian distribution of the responses of the original classification network based on the whole training set. Then, the Mahalanobis distance between one sample and its nearest class-conditional Gaussian distribution is used as the metric for detection. If the distance is larger than a pre-defined threshold, then it is detected as adversarial.
Compared to some aforementioned single-representation-based detection methods, these joint-representation-based methods showed better performances on some databases.
However, the detection cost for each example is much higher, as the responses of its neighboring samples should also be computed. Besides, since the representation is highly dependent on the neighbors or all training examples, the detection performance may be sensitive to data sources, which will be studied in later experiments.

There are also some other approaches that do not construct representations from the responses of the original classification network.
\cite{image-pca-2016} adopted PCA statistics to discriminate adversarial and benign images, independent of any DNN model.
However, the study presented in \cite{bypass-10-detection-2017} has demonstrated that this method works for MNIST but not for CIFAR-10, and PCA statistics are not robust features to detect adversarial images.
\cite{rce-nips-2018} proposed a novel loss called reverse cross-entropy (RCE) to train the classification network, such that the distance measured by kernel density \cite{KD-BU-arxiv-2017} between adversarial and benign examples could be enlarged.
\cite{defensegan-iclr-2018} proposed Defense-GAN to model the distribution of benign examples using a generative adversarial network (GAN). If the Wasserstein distance between one example and its corresponding example generated by GAN is larger than a threshold, then it is detected as adversarial.
However, above three methods are much more costly than other methods.

\vspace{-.8em}
\section{Preliminaries}
\label{Preliminaries}

\vspace{-.4em}
\subsection{Generalized Gaussian Distribution
\label{sec: subsec GGD}}

Assume that a random variable $\X \in \mathbb{R}^d$ follows the generalized Gaussian distribution (GGD) \cite{GGD-1989}. Then, its probability density function (PDF) is formulated with two positive parameters, including the shape factor $c$ and the standard deviation $\sigma$, as follows
\vspace{-.6em}
\begin{flalign}
\vspace{-.9em}
\P_{\X}(x) = A \cdot e^{-|\beta x|^c},
\label{eq: GGD pdf}
\vspace{-2.7em}
\end{flalign}
where $\beta = \frac{1}{\sigma} \big( \frac{\Gamma(3/c)}{\Gamma(1/c)} \big)^{\frac{1}{2}}$
and $A = \frac{\beta c}{2 \Gamma(1/c)}$, with $\Gamma(\cdot)$ being the Gamma function.
Note that the mean parameter $\mu$ is omitted above, as $\mu$ has no relation with the shape of distribution and we set it as $0$ without loss of generality.
A nice characteristic of GGD is that it covers many popular distributions with varied shape factors.
For example, when $c=1$, then it becomes the Laplacian distribution; when $c=2$, then it is the Gaussian distribution with a variance of $\sigma^{2}$; when $c \rightarrow +\infty$, then it is specified as a uniform distribution on $(-\sqrt{2}\sigma,\sqrt{2}\sigma)$.

\vspace{-.3em}
\subsection{Benford-Fourier Coefficients}
\label{sec: subsec BF}
\vspace{-0.3em}

Although generalized Gaussian distribution (GGD) is able to cover a bunch of distributions, it is hard to depict the exact forms of GGD precisely.
To this end, we further define a random variable $\Z = \log_{10} \left | \X \right |\mod 1$ for detecting and distinguishing different form of GGD, of which the PDF is formulated by means of Fourier Series as \cite{bf-icip-2007}, with the fundamental period being fixed as $2\pi$,
\begin{flalign}
\vspace{-2mm}
\P_{\Z}(z) &= 1+2\sum_{n=1}^{+\infty}\left [ A_n\cos\left ( 2\pi nz \right ) + B_n\sin\left ( 2\pi nz\right )\right ]\nonumber\\
&= 1 + 2 \sum_{n=1}^{+\infty} |a_n| \cos(2\pi n z + \phi_n),
\vspace{-2mm}
\end{flalign}
 where $z \in [0, 1)$ corresponds to the domain of random variable $\Z$, the phase of Fourier Series is explained as $\phi_n=\arctan\big( -\frac{B_n}{A_n} \big)$, and the magnitude denotes $|a_n|=\sqrt{A_n^2+B_n^2}$. $a_n = |a_n| \cdot e^{j \phi_n}$ denotes the $n$-th Fourier coefficient of $\mathcal{P}_\Z(z)$ evaluated at $2\pi n$, and its definition is
\vspace{-0.35 em}
\begin{flalign}
\label{eq: exact value of an}
a_n &= \int_{-\infty}^{+\infty} \mathcal{P}_\Z(z) \cdot e^{-j 2\pi n \log_{10} z} \mathrm{d} z \nonumber\\
&=
\frac{2A e^{\frac{j 2\pi n \log \beta}{\log 10}}}{\beta c}  \cdot \Gamma\bigg( \frac{-j2\pi n + \log 10}{c \log 10} \bigg) .
\end{flalign}
$a_n$ is also called as Benford-Fourier coefficient. Note that $a_n$ is a complex number, and its magnitude is calculated as
\begin{flalign}
\vspace{-1.5em}
| a_n | = \bigg( \prod_{k=0}^{+\infty} \bigg[ 1+ \big(\frac{2\pi n}{\log 10 (ck+1)}\big)^2 \bigg]^{-1} \bigg)^{\frac{1}{2}} .
\label{eq: an percise}
\vspace{-1.5em}
\end{flalign}
Note that $| a_n |$ gets smaller as $n \in \mathbb{N}$ increases. And, an interesting property of $| a_n |$ is that it only depends on the shape factor $c$, while is independent of the parameter $\sigma$.
Thus, one set of the absolute values of Benford-Fourier coefficients $\{ | a_n | \}_{n \in \mathbb{N}}$ correspond to one identical $c$, \ie, one identical special distribution of GGD. In other words, we could use $\{ | a_n | \}_{n \in \mathbb{N}}$ as features or representations to discriminate different special distributions of GGD.

However, if it is often difficult to know or even estimate the shape factor $c$, we cannot compute the value of $| a_n |$. But fortunately, recalling that $a_n$ is the $n$-th Fourier coefficient of $\mathcal{P}_\Z(z)$ evaluated at $2\pi n$, we can derive an easy estimation. Specifically, assume that $\bx = \{ x_1, \ldots, x_M \}$ is a set of $M$ i.i.d. points sampled from GGD with the same shape factor $c$. Then, the corresponding Benford-Fourier coefficients can be estimated as follows \cite{bf-icip-2014}:
\vspace{-0.65 em}
\begin{flalign}
\vspace{-1.0 em}
\label{eq: estimation of an}
&\hat{a}_n = \frac{\sum_{m=1}^M e^{-j 2\pi n \log_{10} |x_m|} }{M} =
\\
&\frac{1}{M}  \sum_{m=1}^{M} \bigg[\cos{(2\pi n \log_{10} |x_m|)} -j \sin{(2\pi n \log_{10} |x_m|)}
\bigg]. \nonumber
\vspace{-2.0 em}
\end{flalign}
The gap between $\hat{a}_n$ and $a_n$ is analyzed in Theorem \ref{theorem: estimation error of an}. It tells that $\hat{a}_n$ gets closer to $a_n$ as $M$ increases.
For clarity, we firstly introduce a few notations: $\mathcal{T}=e^{-j 2\pi n \log_{10}|\mathcal{X}|}$ is a random variable with $\mathcal{X}$ obeying the generalized Gaussian distribution, and $\hat{a}_n$ is an observation of the random variable $\mathcal{Y}=\frac{1}{M}\sum_{m=1}^{M}\mathcal{T}_m$.
Due to the space limit, the proof will be presented in {\bf supplementary material A}.

\vspace{-2.5 mm}
\begin{theorem} \label{theorem: estimation error of an}
\vspace{-2.5 mm}
Assume that the estimation error $\varepsilon_n= \hat{a}_n -a_n$ is an observation of the random variable $\mathcal{E} = \mathcal{Y}-a_n$. $\left | \mathcal{E}\right |$ follows the Rayleigh distribution \cite{rayleigh}, of which the probability density function is formulated as
\begin{flalign}
\vspace{-2.5 mm}
\P_{\left | \mathcal{E}\right |}\left ( r \right ) = 2Mre^{-Mr^2}.
\label{eq: error of an}
\vspace{-1.5 mm}
\end{flalign}
And the expectation and variance are
\vspace{-2.5 mm}
 \begin{flalign}
\vspace{-2.5 mm}
E\big ( \left | \mathcal{E}\right | \big )=\frac{1}{2}\sqrt\frac{\pi}{M},\quad
 D(\big | \mathcal{E}\big |)=\frac{4-\pi}{4M},
\label{eq: error of an}
\nonumber
\vspace{-0.8 em}
\end{flalign}
which implies that the estimation error $\varepsilon_n$ gets closer to 0 as the number of samples $M$ increases.
\vspace{-1.0 em}
\end{theorem}

\section{Adversarial Detection via Benford-Fourier Coefficients}
\label{others}

\subsection{Training Procedure of Adversarial Detector}
\label{sec: subsec train detector}
There are three stages to train the proposed adversarial detector, including: 
{\bf 1)} building a training set based on benign images;
{\bf 2)} extracting novel representations of the training set via Benford-Fourier coefficients;
{\bf 3)} training a SVM classifier as the adversarial detector.
They will be explained in details sequentially.
And, the overall training procedure is briefly summarized in Algorithm \ref{alg: train detector}.

\vspace{-.1em}
\noindent
{\bf Build a training set}.
Firstly, we collect $N$ clean images $\{ \bx_1, \ldots, \bx_N \}$, which can be correctly predicted by $f_{\boldsymbol{\theta}}$.
Then, we adopt one adversarial attack method (\eg, C\&W \cite{CW} or BIM  \cite{BIM}) to generate one adversarial image corresponding to each clean image. The crafted $N$ adversarial examples are denoted as $\{ \hat{\bx}_1, \ldots, \hat{\bx}_N \}$.
Besides, to avoid that the noisy image (polluted by some kind of non-malicious noises but still can be correctly predicted by $f_{\boldsymbol{\theta}}$) is incorrectly detected as adversarial, we also craft one noisy image by adding small random Gaussian noises onto each clean image. These $N$ noisy examples are denoted as $\{ \bar{\bx}_1, \ldots, \bar{\bx}_N \}$.
Note that, hereafter benign examples include both clean and Gaussian noisy examples.
Consequently, we obtain one training set with $3N$ examples, denoted as
$\mathcal{D}_{tr} = \{ (\bx_i, -1), (\bar{\bx}_i, -1), (\hat{\bx}_i, +1)\}_{i = 1, \ldots, N}$.

\vspace{-.1em}
\noindent
{\bf Extract novel representations}.
We firstly feed the $i$-th training image from $\mathcal{D}_{tr}$ into $f_{\boldsymbol{\theta}}$.
We concatenate all response entries of the $l$-th layer in $f_{\boldsymbol{\theta}}$ to obtain one vector $\mathbf{r}_i^l$.
Then, we estimate the corresponding Benford-Fourier coefficients according to Eq. (\ref{eq: estimation of an}), as follows
\vspace{-0.6em}
\begin{flalign}
\vspace{-0.6em}
(\hat{a}_n)_i^l = \frac{1}{M_i^l} \sum_{m=1}^{M_i^l} e^{-j 2\pi n \log_{10} |(r_m)_i^l|} ,
\vspace{-0.4em}
\end{flalign}
where $M_i^l$ indicates the length of $\mathbf{r}_i^l$.
%
The magnitude of $(\hat{a}_n)_i^l$ is computed as follows
\vspace{-0.6em}
\begin{flalign}
\label{eq: |a_n| in CNN}
\vspace{-0.6em}
| (\hat{a}_n)_i^l |
=
\frac{1}{M_i^l}   &\bigg(\sum_{m=1}^{M_i^l}\cos{(2\pi n \log_{10} |(r_m)_i^l|)^2} \\\nonumber&+ \sum_{m=1}^{M_i^l}\sin{(2\pi n \log_{10} |(r_m)_i^l|)^2} \bigg)^{\frac{1}{2}}
\vspace{-1em}
\end{flalign}
Then, we extract one $T$-dimensional 
feature vector $\mathbf{a}_i^l = [|(\hat{a}_1)_i^l|, \ldots, |(\hat{a}_{T})_i^l|] \in \mathbb{R}_+^{T}$ for the $i$-th training image from the $l$-th layer.
We set $T=16$ in experiments, as $| (\hat{a}_n)_i^l |$ of larger $n$ is too small for discrimination.
Finally, we concatenate the feature vectors of all layers to form a long vector $\hat{\mathbf{a}}_i = [\hat{\mathbf{a}}_i^1; \ldots; \hat{\mathbf{a}}_i^L]  \in \mathbb{R}_+^{T L}$, with $L$ being the number of layers in $f_{\boldsymbol{\theta}}$.
Consequently, we obtain a novel representation of training images, denoted as
$\hat{\mathcal{A}}_{tr} = \{ (\hat{\mathbf{a}}_i, \pm 1) \}_{i = 1, \ldots, 3N}$, where the label $+1$ or $-1$ is directly obtained from $\mathcal{D}_{tr}$.

\vspace{-.1em}
\noindent
{\bf Train an adversarial detector}.
Finally, we train a binary SVM classifier based on $\hat{\mathcal{A}}_{tr}$.
The trained SVM classifier will serve as the adversarial detector for the CNN model $f_{\boldsymbol{\theta}}$.

\vspace{-.1em}
\noindent
{\bf Testing}.
One novel testing example is firstly predicted as adversarial or not by the trained adversarial detector . If adversarial, then it is rejected; otherwise, it is fed into $f_{\boldsymbol{\theta}}$ to predict its class label.

\noindent
{\bf Remark}.
Note that in the derivation of $a_n$ (see Eq. (\ref{eq: exact value of an})), the mean parameter of GGD is set to $0$. In experiments, we calculate the mean values of internal-layer responses of all networks for every image, and find that mean parameters at most layers are close to 0, while the mean parameters at a few layers could be large. However, the mean value is subtracted from each response entry when we extract the MBF features in our experiments. Thus, the derived $a_n$ is applicable to our task.\comment{most mean values are very close to $0$ (around $10^{-2}$)}
Besides, according to Theorem \ref{theorem: estimation error of an}, the estimation error of $(\hat{a}_n)_i^l$ is inversely proportional to $M_i^l$. It tells that the coefficient estimated from the larger-sized layer is more accurate. In many neural networks (\eg, AlexNet \cite{alexnet}), the response sizes of high layers get smaller, which means the less accurate estimation. However, we believe that the discrimination between adversarial and benign examples in higher layers is more evident than that in lower layers. There is a trade-off between estimation accuracy and discrimination. This is why we concatenate the estimated magnitudes of all layers together to construct the novel representation.
\setlength{\textfloatsep}{10pt}
\begin{algorithm}[t]
\small
\caption{~ Training the adversarial detector via the magnitude of Benford-Fourier coefficients.}
\label{alg: train detector}
\begin{algorithmic}
\STATE {\bfseries Input:} The trained CNN model $f_{\boldsymbol{\theta}}$ with $L$ layers, and the training set $\mathcal{D}_{tr}$
   \FOR{$i=1$ {\bfseries to} $|\mathcal{D}_{tr}|$}
   \FOR{$i=1$ {\bfseries to} $L$}
   \STATE Compute $| (\hat{a}_n)_i^l |$ as Eq. (\ref{eq: |a_n| in CNN}), with $n = 1, \ldots, T$
   \STATE Concatenate $\{ | (\hat{a}_n)_i^l | \}_{n = 1,\ldots, T}$ to obtain a vector $\hat{\mathbf{a}}_i^l$
   \ENDFOR
   \STATE Concatenate $\{ \hat{\mathbf{a}}_i^l \}_{l=1,\ldots, L}$ to obtain a long vector $\hat{\mathbf{a}}_i$
   \ENDFOR
   \STATE Build a novel representation of the training set, denoted as $\hat{\mathcal{A}}_{tr} = \{ (\hat{\mathbf{a}}_i, \pm 1) \}_{i = 1, \ldots, |\mathcal{D}_{tr}|}$
   \STATE Train a binary SVM classifier based on $\hat{\mathcal{A}}_{tr}$
   \STATE {\bfseries Return} The trained binary SVM classifier
\end{algorithmic}

\end{algorithm}

\vspace{-3.0 mm}

\subsection{Experimental Settings}
\label{sec: subsec experimental settings}

\vspace{-0.3em}
\noindent
{\bf Databases and network architectures}
We conduct experiments on three databases, including CIFAR-10 \cite{cifar10}, SVHN \cite{svhn}, and a subset of ImageNet \cite{imagenet}.
In terms of CIFAR-10 and SVHN, we adopt the same settings as the compared method LID \cite{LID-iclr-2018}. 
Specifically, a 33-layer network pre-trained on the training set of CIFAR-10 achieves $82.37\%$ accuracy on the testing set with $10,000$ benign images; a 19-layer network pre-trained on the training set of SVHN achieves $92.6\%$ accuracy on the testing set with $26,032$ benign images.
Then, we add a small noise drawn from $\mathcal{N}(0, \sigma^2)$ on each testing image, with $\sigma$ being the similar level of the $\ell_2$ norm of adversarial perturbations on the same database. If both the benign and its noisy image can be correctly predicted by the classification network, then it is picked out for training the detection.
We finally collect $8,175$ and $23,862$ benign images from CIFAR-10 and SVHN, respectively.
These images are randomly partitioned to the $80\%$ training set and the $20\%$ testing set, used for the training and testing of the detector.
We also collect a subset from ImageNet, including 800 benign images of 8 classes ({\it snowbird, spoonbill, bobtail, Leonberg, hamster, proboscis monkey, cypripedium calceolus, and earthstar}). The 100 images of each class contain 50 testing images and 50 randomly selected training images.
We fine-tune the checkpoints of both AlexNet and VGG pre-trained on ImageNet\footnote{https://pytorch.org/docs/robust/torchvision/models.html} on these 800 images to achieve $100\%$ accuracy.
Then, 785 benign images are kept for detection, as both their noisy images and  themselves can be correctly predicted by both the fine-tuned AlexNet and VGG models.
These 785 images are then randomly partitioned to 480 training and 305 testing images used for detection.
For each database, as described in Section \ref{sec: subsec train detector}, one noisy and one adversarial image are generated for each benign image; then, all of benign, noisy, and adversarial images are used for detection.

\vspace{-0.1em}
\noindent
{\bf Attack methods}
We adopt four popular adversarial attack methods to craft adversarial examples, including basic iterative method (BIM \cite{BIM}), CarliniWagnerL2Attack (CW-L2  \cite{CW}), DeepFool \cite{DeepFool}, and random projected gradient descent (R-PGD \cite{RandomPGD}).
They are implemented by Foolbox\footnote{https://foolbox.readthedocs.io/en/v1.8.0/}.
The hyper-parameters of these methods will be presented in {\bf supplementary material D}.

\vspace{-0.1em}
\noindent
{\bf Compared detection methods}
We compare with three state-of-the-art and open-sourced adversarial detection methods, including KD+BU\footnote{
\text{https://github.com/rfeinman/detecting-adversarial-samples/}
} \cite{KD-BU-arxiv-2017},
Mahalanobis distance\footnote{
\text{https://github.com/pokaxpoka/deep\_Mahalanobis\_detector/}
} (M-D) \cite{M-distance-nips-2018},
and LID\footnote{
\text{https://github.com/xingjunm/lid\_adversarial\_subspace\_detection/}
} \cite{LID-iclr-2018}.
Note that another recent work called I-defender \cite{I-defender-nips-2018} is not compared, as its code is not available.
To ensure the fair comparison, the SVM classifier is trained with all compared methods, implemented by the {\it fitcsvm}\footnote{https://www.mathworks.com/help/stats/fitcsvm.html} function in MATLAB.
There are two important hyper-parameters in LID, \ie, the size of mini-batch and the number of neighbors.
On CIFAR-10 and SVHN, they are respectively set as 100 and 20, as suggested in \cite{LID-iclr-2018};
on ImageNet, as there are only 400 benign training images, they are respectively set as 50 and 20 in experiments.
Moreover, we find that there are some unfair settings in the implementations of compared methods.
For example, KD+BU utilizes the extra $50,000$ images of CIFAR-10 to compute the kernel density of each training and testing image; M-D also uses these extra images to compute the mean and co-variance of
GMM. Since extra images of the similar distribution with the training images are often unavailable, we believe that extra images should not be used to ensure the fair comparison. Thus, extra images are not used for KD+BU and M-D in our experiments.
Besides, LID utilizes other benign testing images as neighborhoods to extract features for each testing image. It is unfair to utilize the information of benign or adversarial for neighboring testing images.
In our experiments, we use benign training images as neighborhoods.
Moreover, we also compare with Defense-GAN \cite{defensegan-iclr-2018} and C1$\&$C2t/u \cite{NIPS2019_turning}, and show the computational complexities of all compared methods. Due to the space limit, these results will be reported in {\bf supplementary material C} and {\bf E}.

\begin{figure}[t]
\centering
\scalebox{0.45}
{
\subfigure[ clean]{
\begin{minipage}[b]{0.14\textwidth}
\includegraphics[width=1\linewidth]{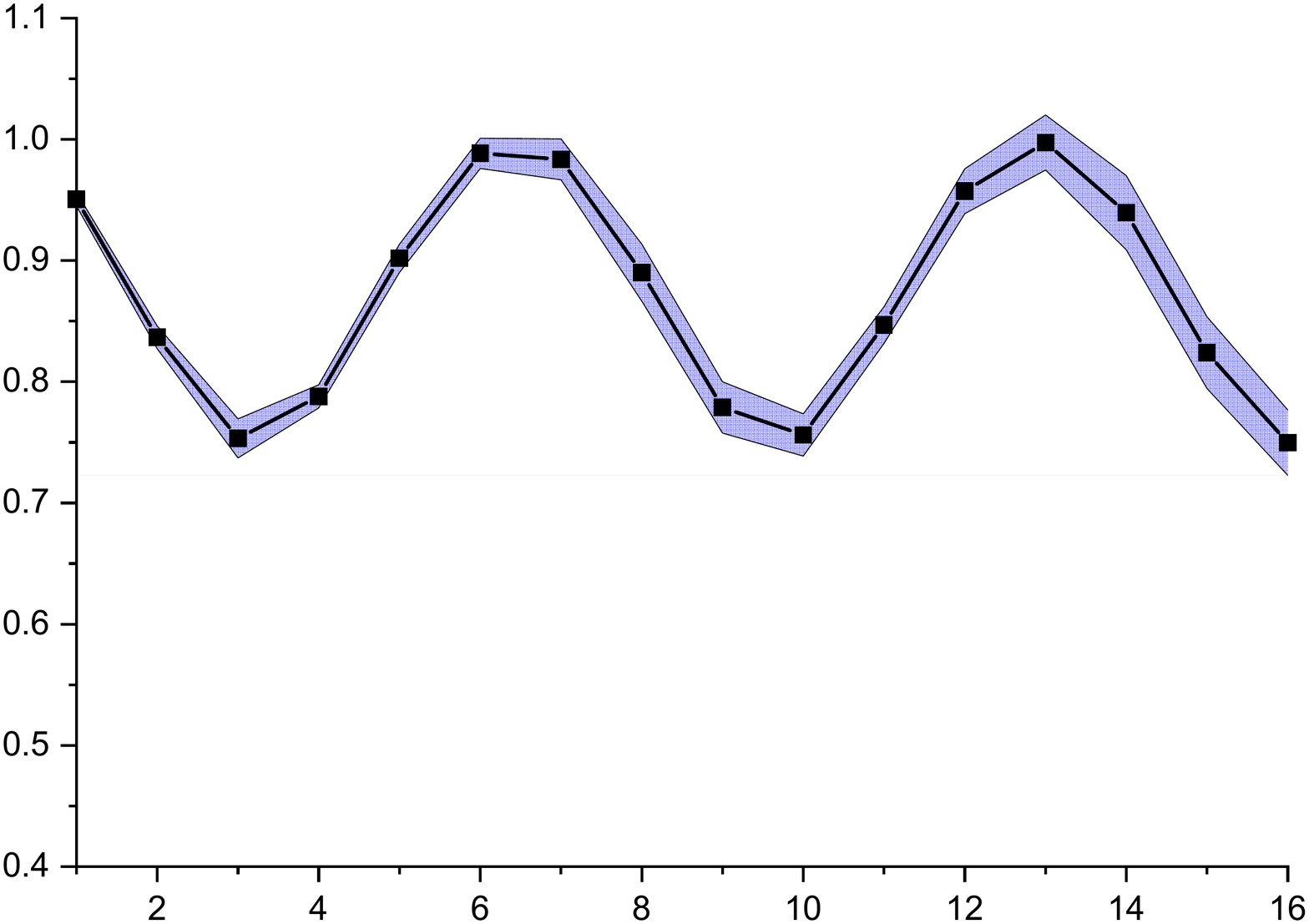}\vspace{4pt}
\includegraphics[width=1\linewidth]{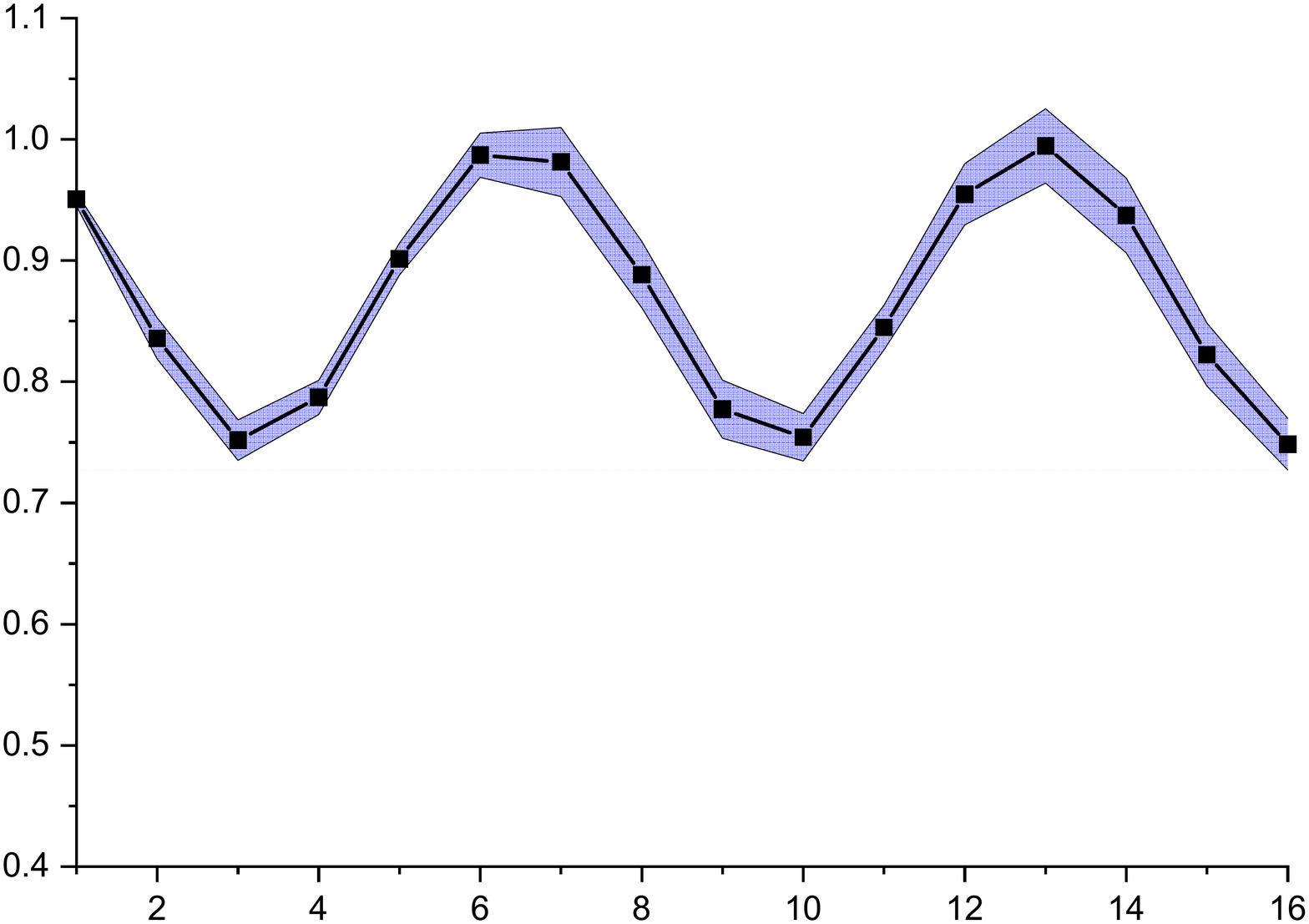}\vspace{4pt}
\includegraphics[width=1\linewidth]{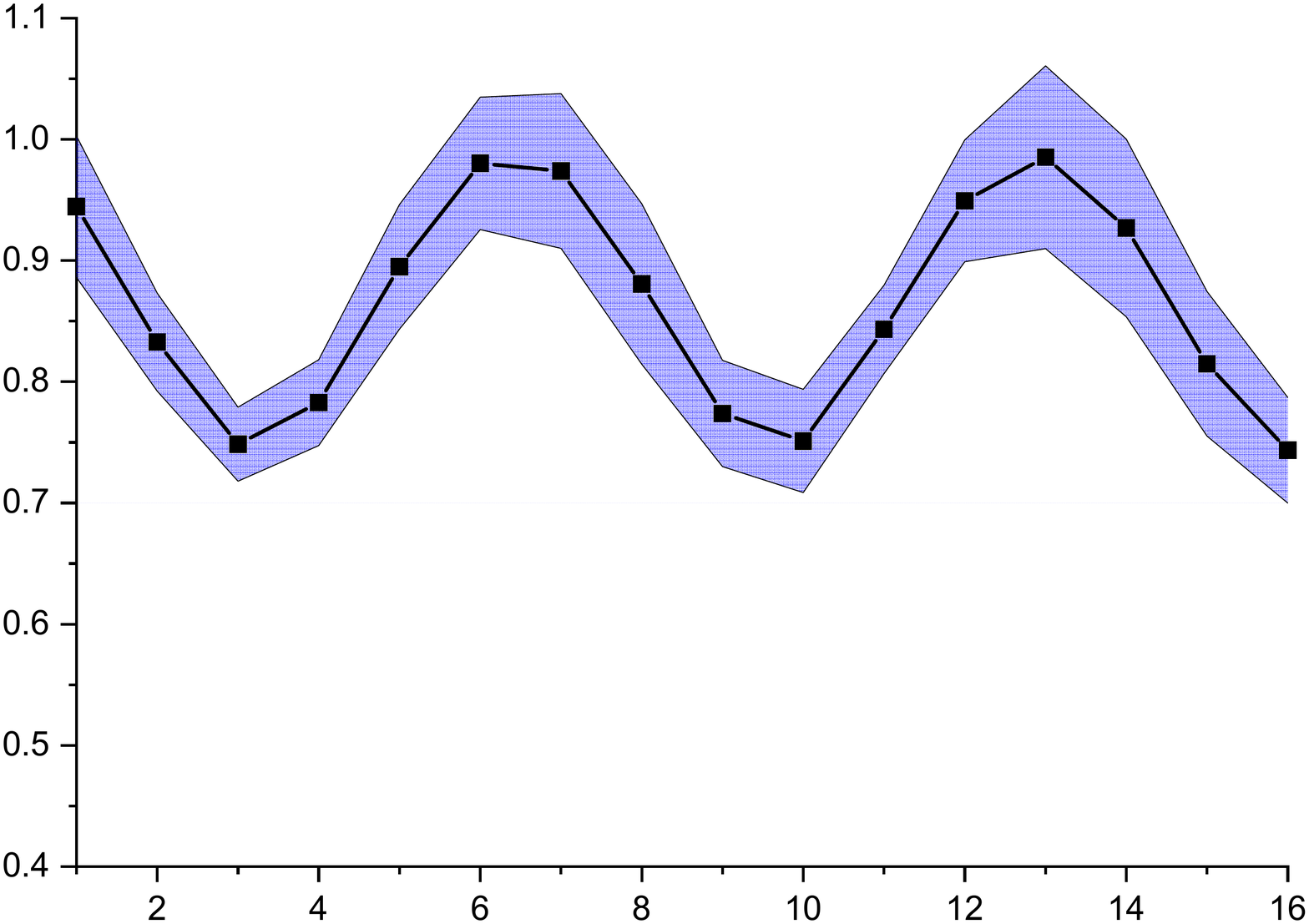}
\end{minipage}
}
\subfigure[ noisy]{
\begin{minipage}[b]{0.14\textwidth}
\includegraphics[width=1\linewidth]{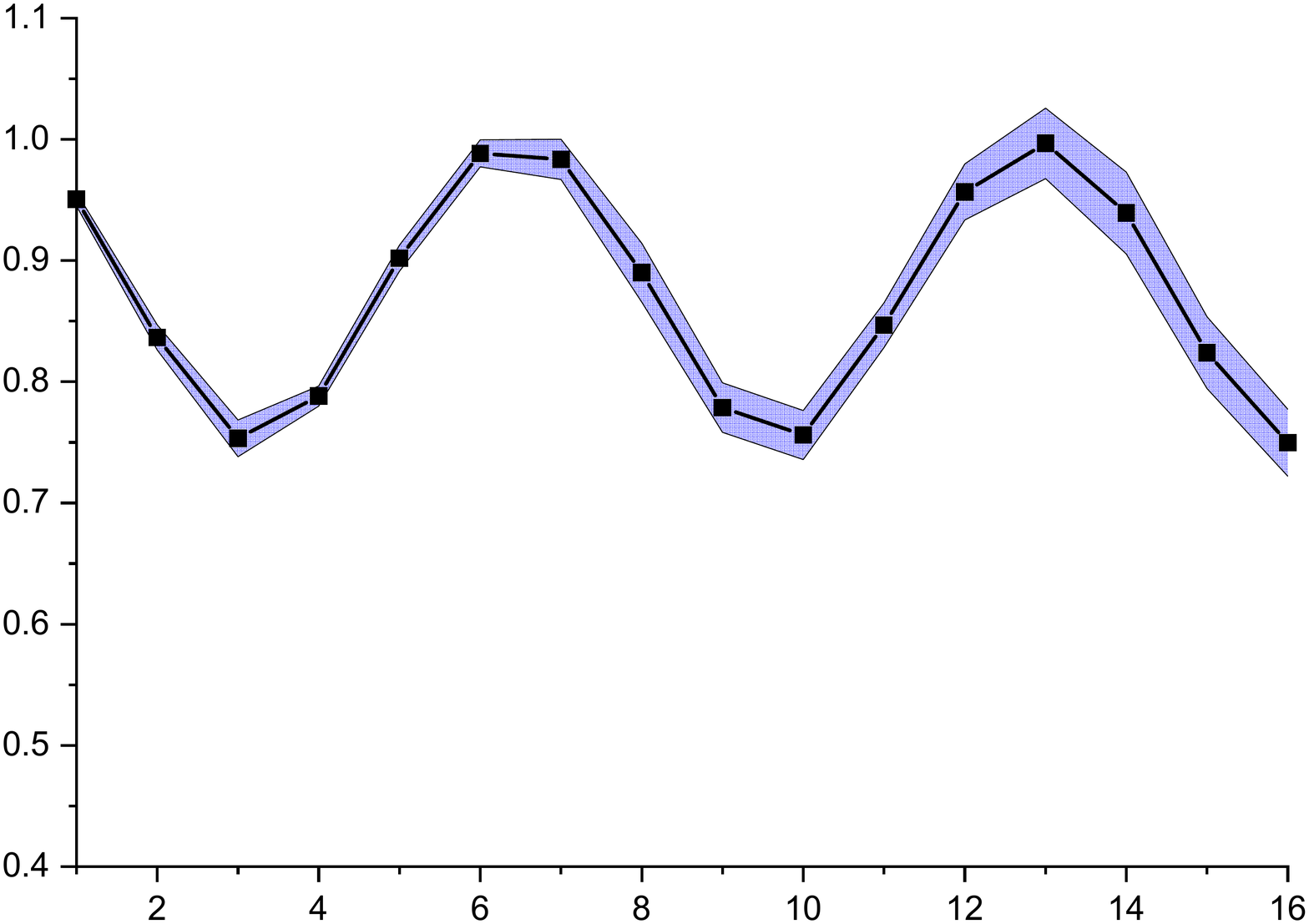}\vspace{4pt}
\includegraphics[width=1\linewidth]{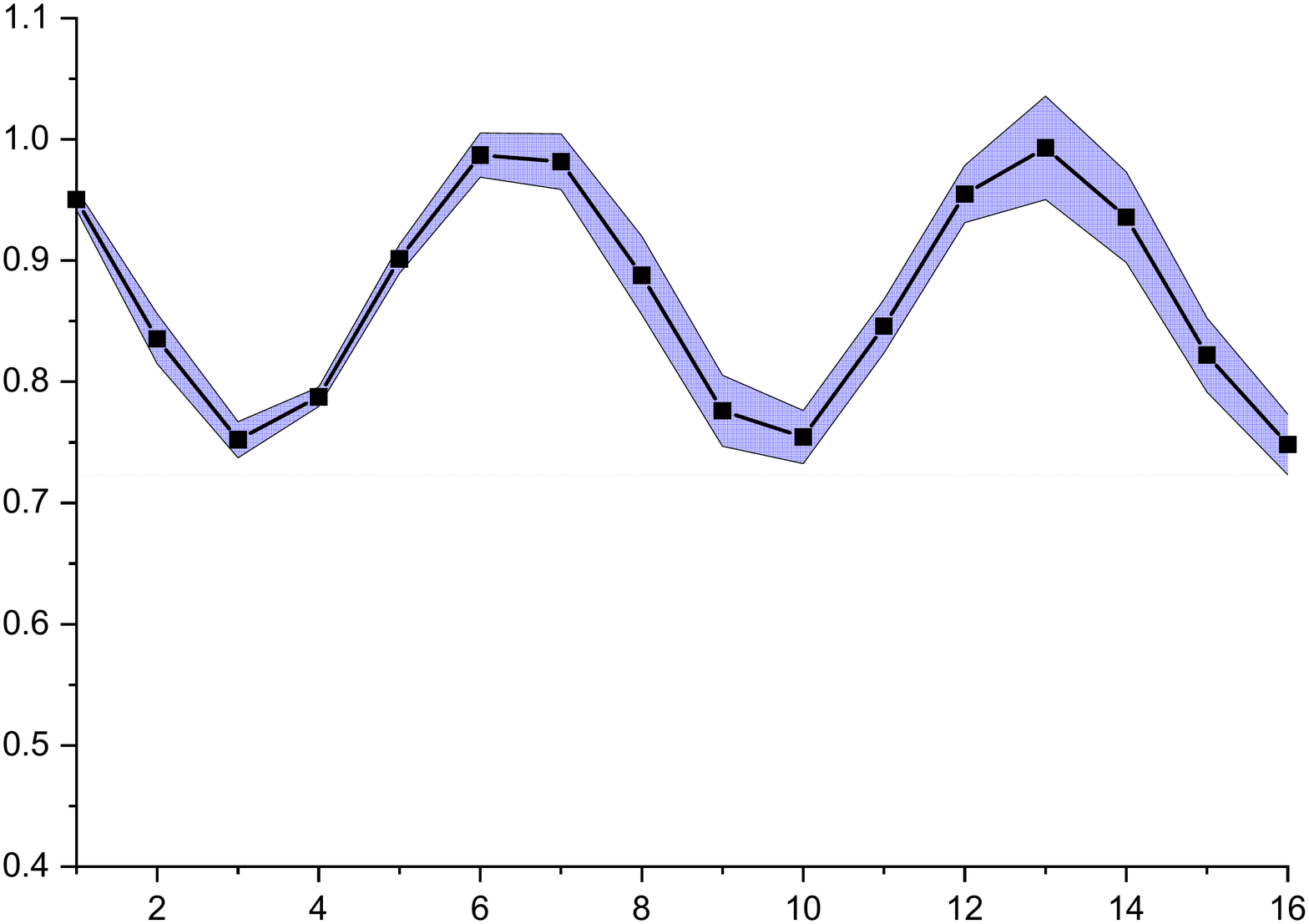}\vspace{4pt}
\includegraphics[width=1\linewidth]{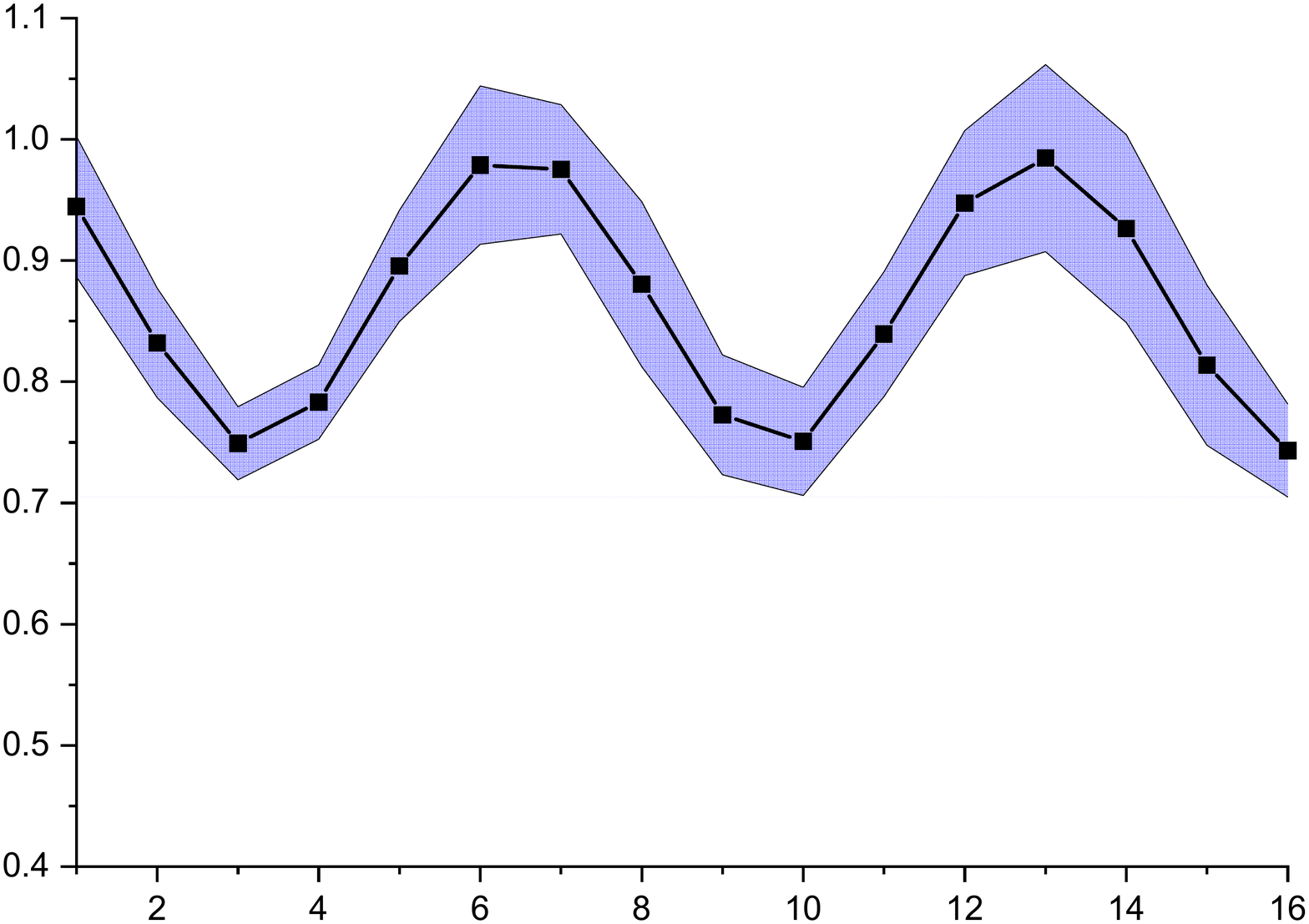}
\end{minipage}
}
\subfigure[  BIM]{
\begin{minipage}[b]{0.14\textwidth}
\includegraphics[width=1\linewidth]{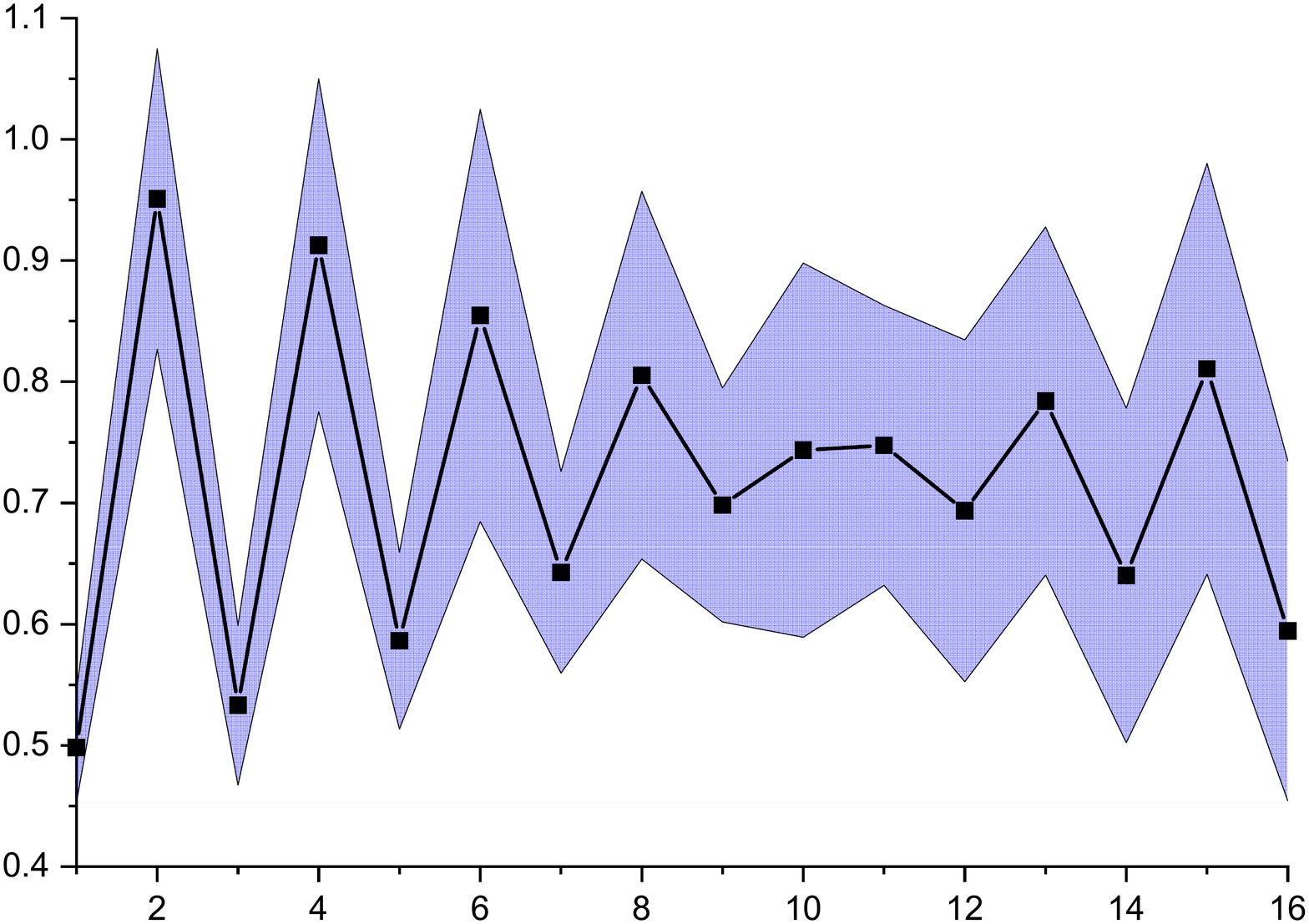}\vspace{4pt}
\includegraphics[width=1\linewidth]{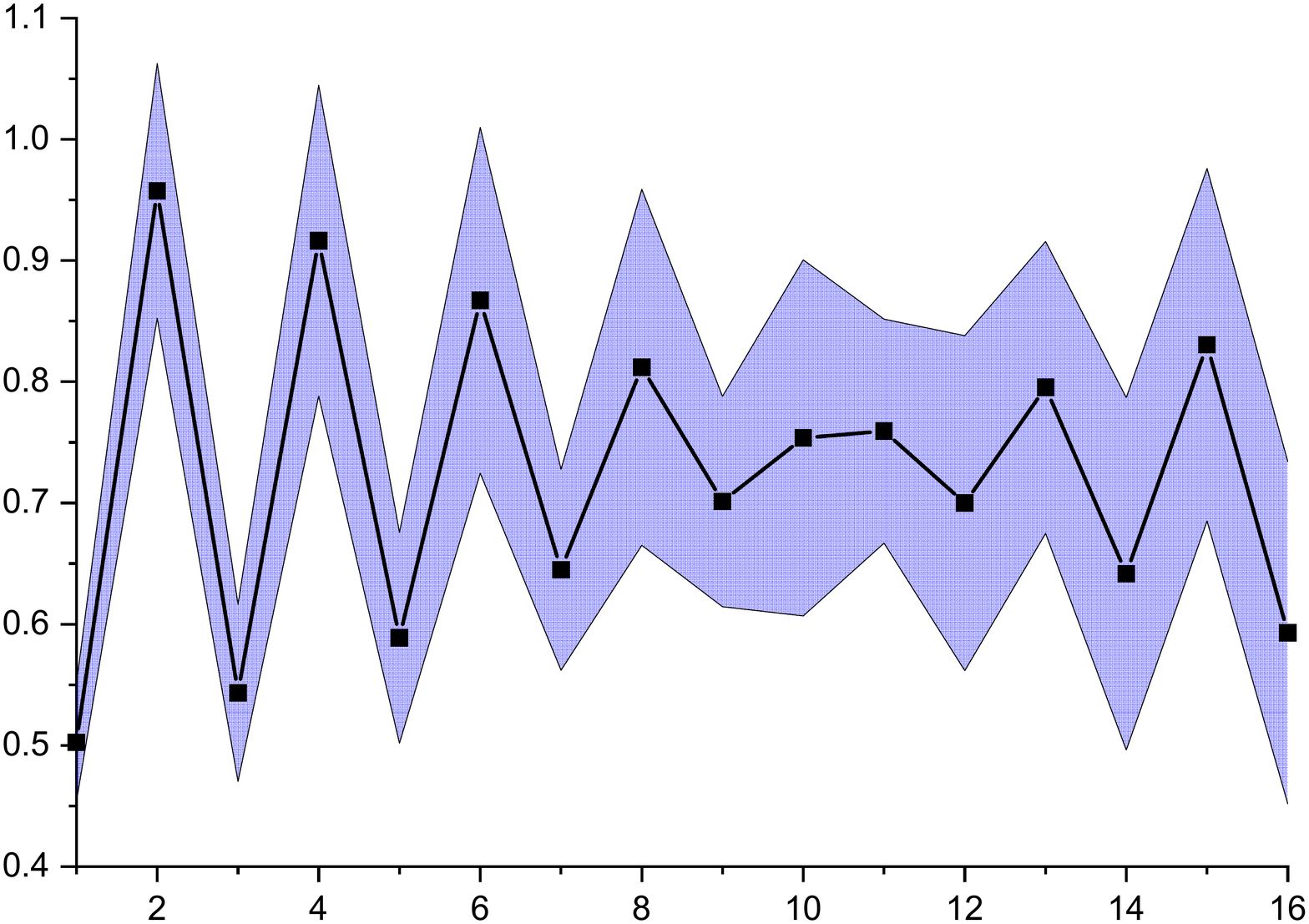}\vspace{4pt}
\includegraphics[width=1\linewidth]{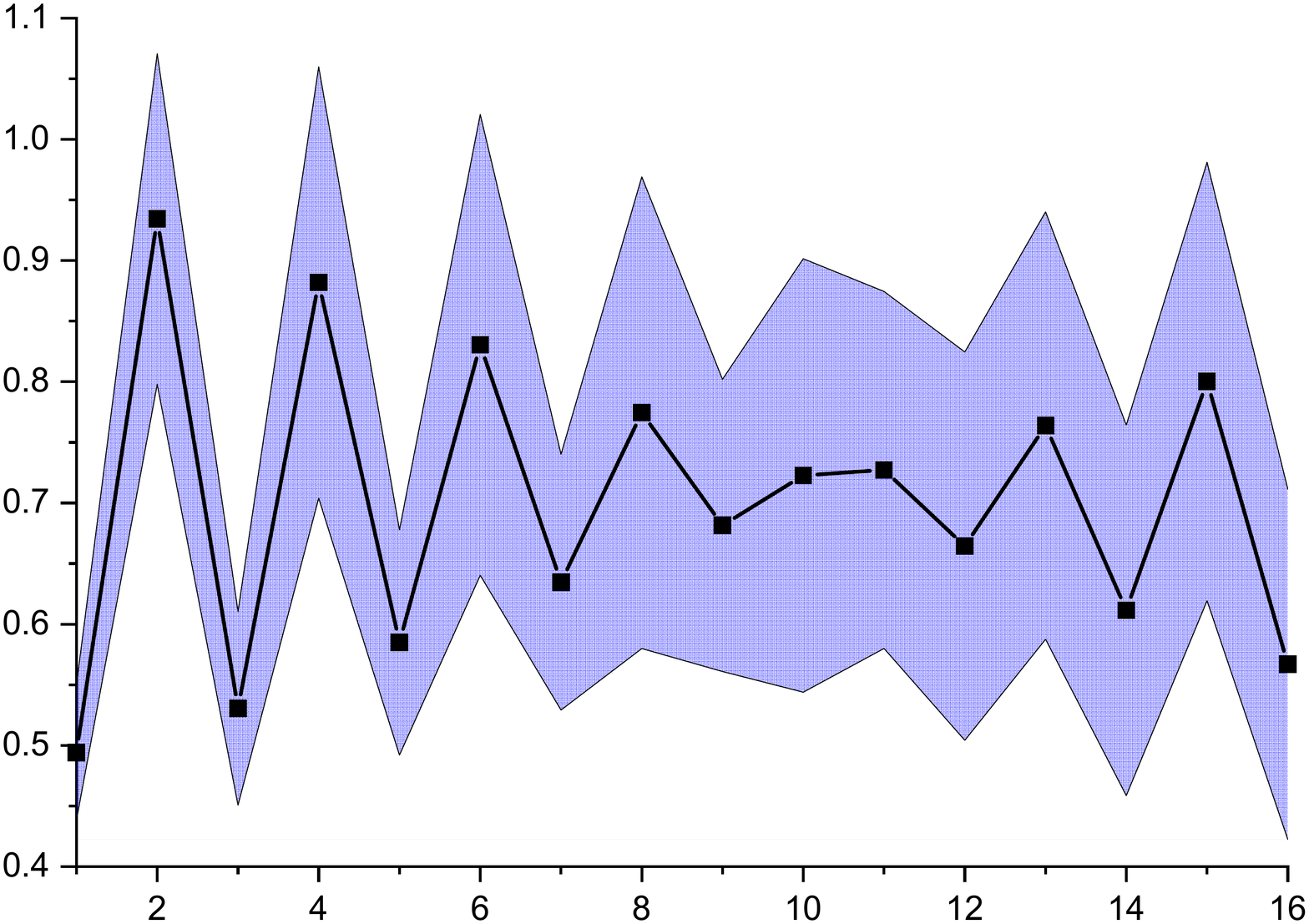}
\end{minipage}
}
\subfigure[  CW-L2]{
\begin{minipage}[b]{0.14\textwidth}
\includegraphics[width=1\linewidth]{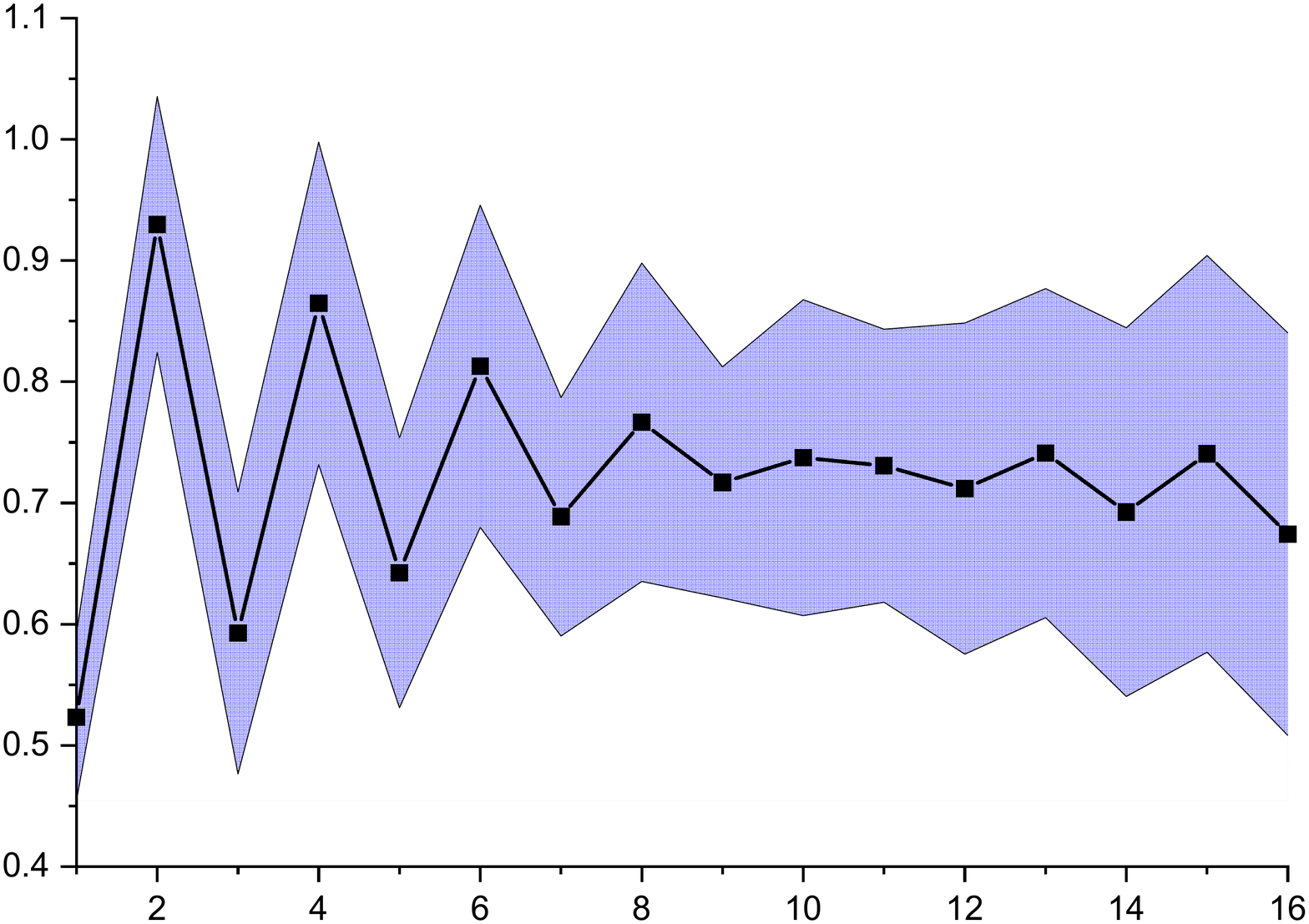}\vspace{4pt}
\includegraphics[width=1\linewidth]{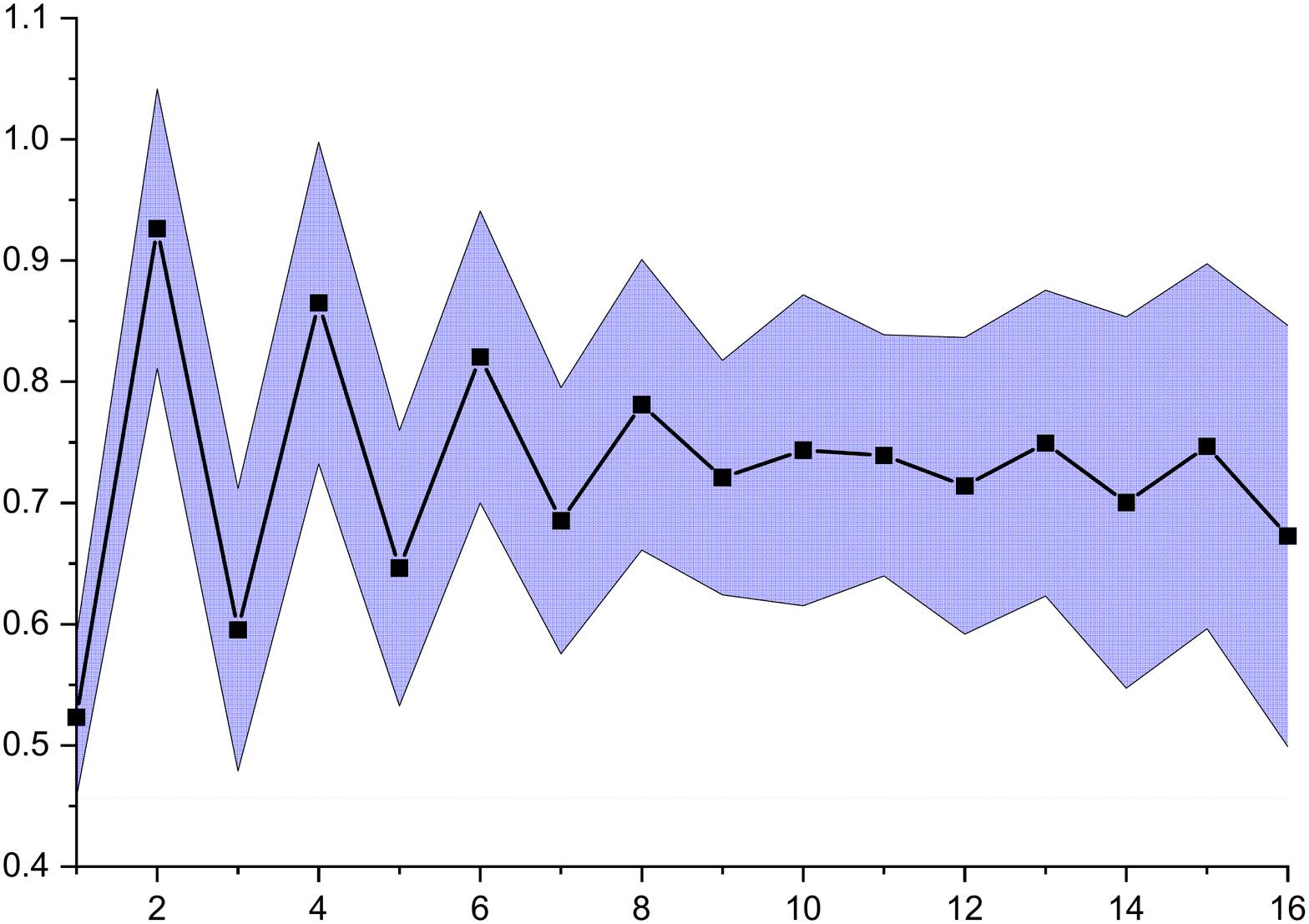}\vspace{4pt}
\includegraphics[width=1\linewidth]{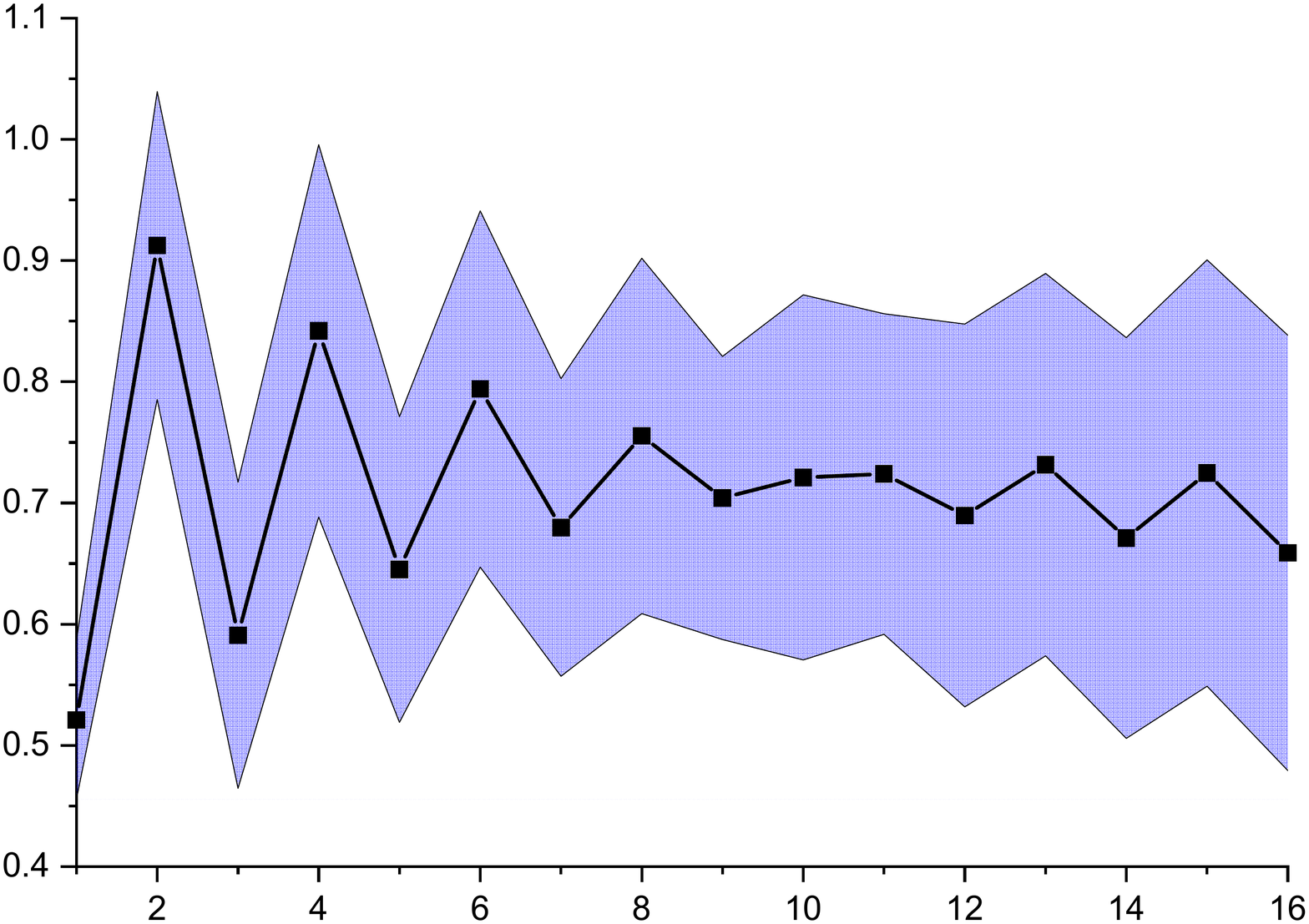}
\end{minipage}
}
\subfigure[  DeepFool]{
\begin{minipage}[b]{0.14\textwidth}
\includegraphics[width=1\linewidth]{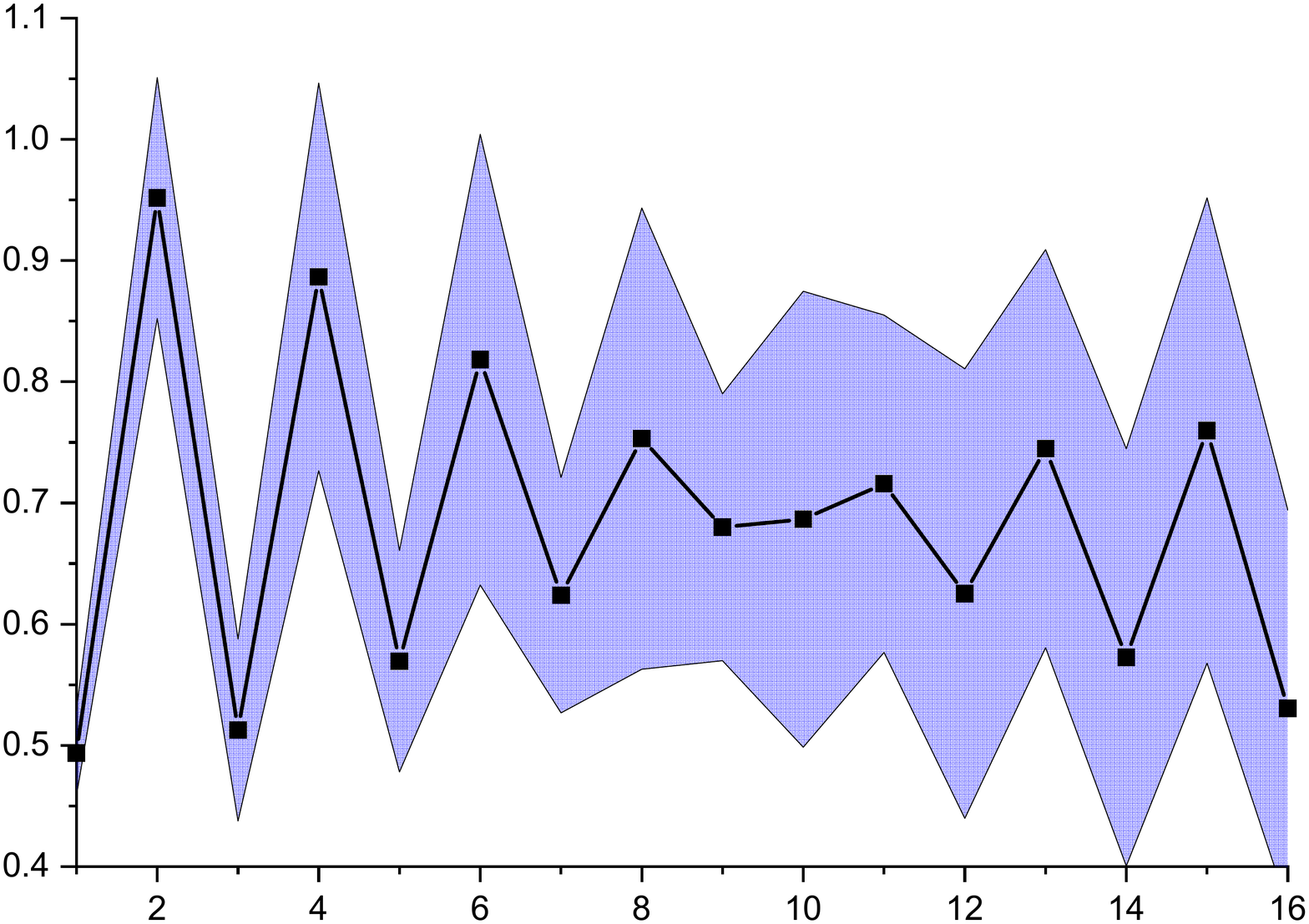}\vspace{4pt}
\includegraphics[width=1\linewidth]{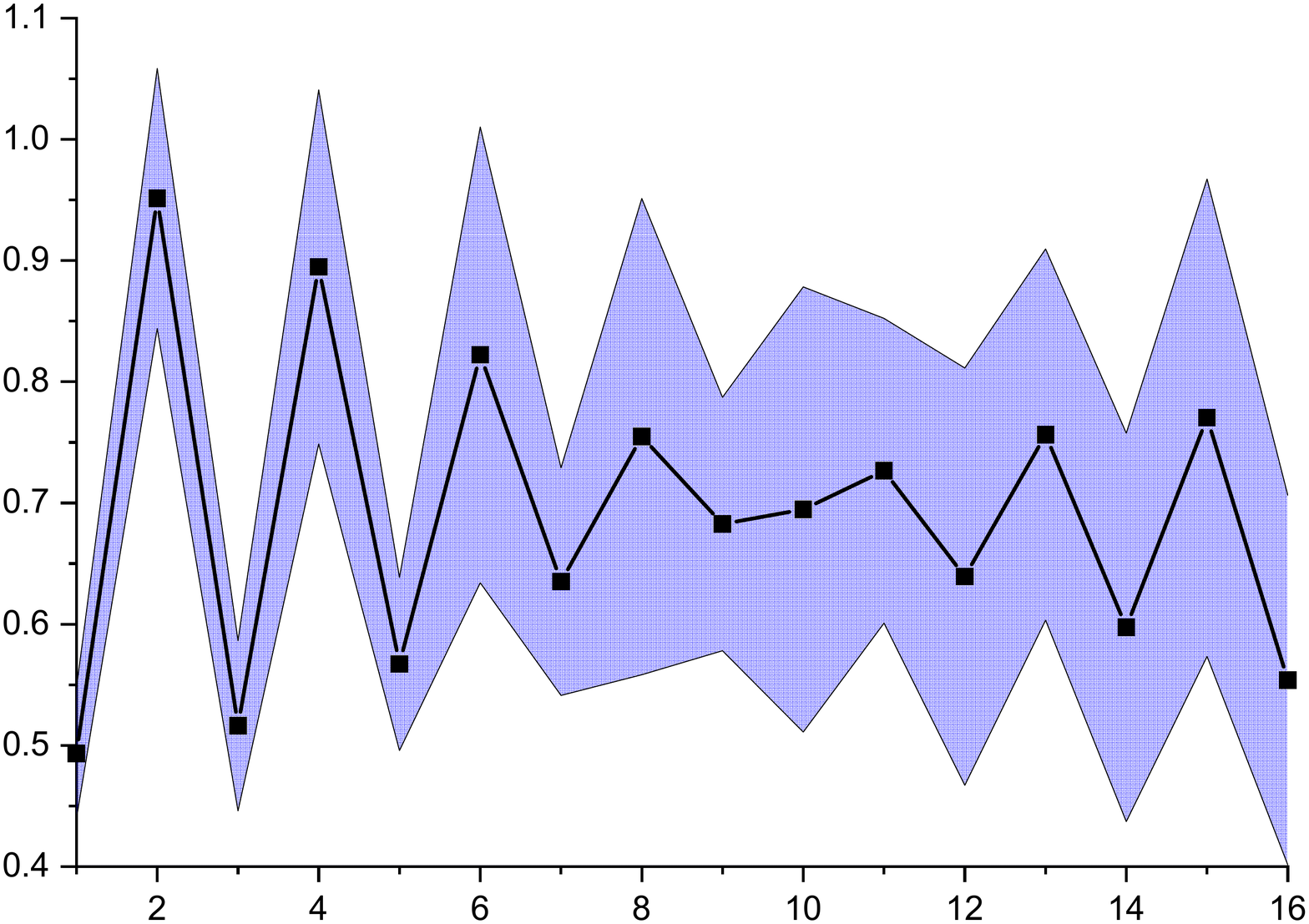}\vspace{4pt}
\includegraphics[width=1\linewidth]{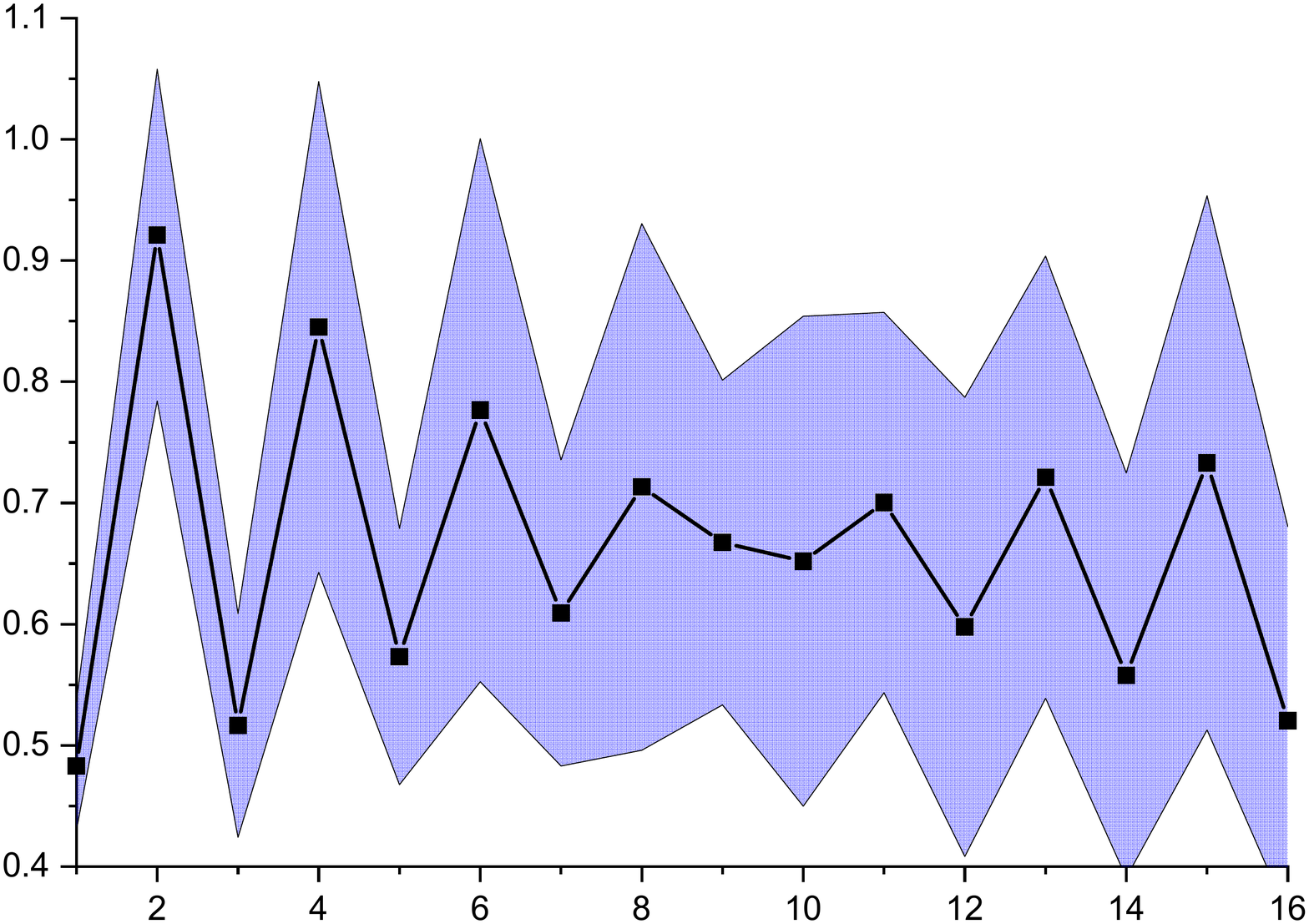}
\end{minipage}
}
\subfigure[  R-PGD]{
\begin{minipage}[b]{0.14\textwidth}
\includegraphics[width=1\linewidth]{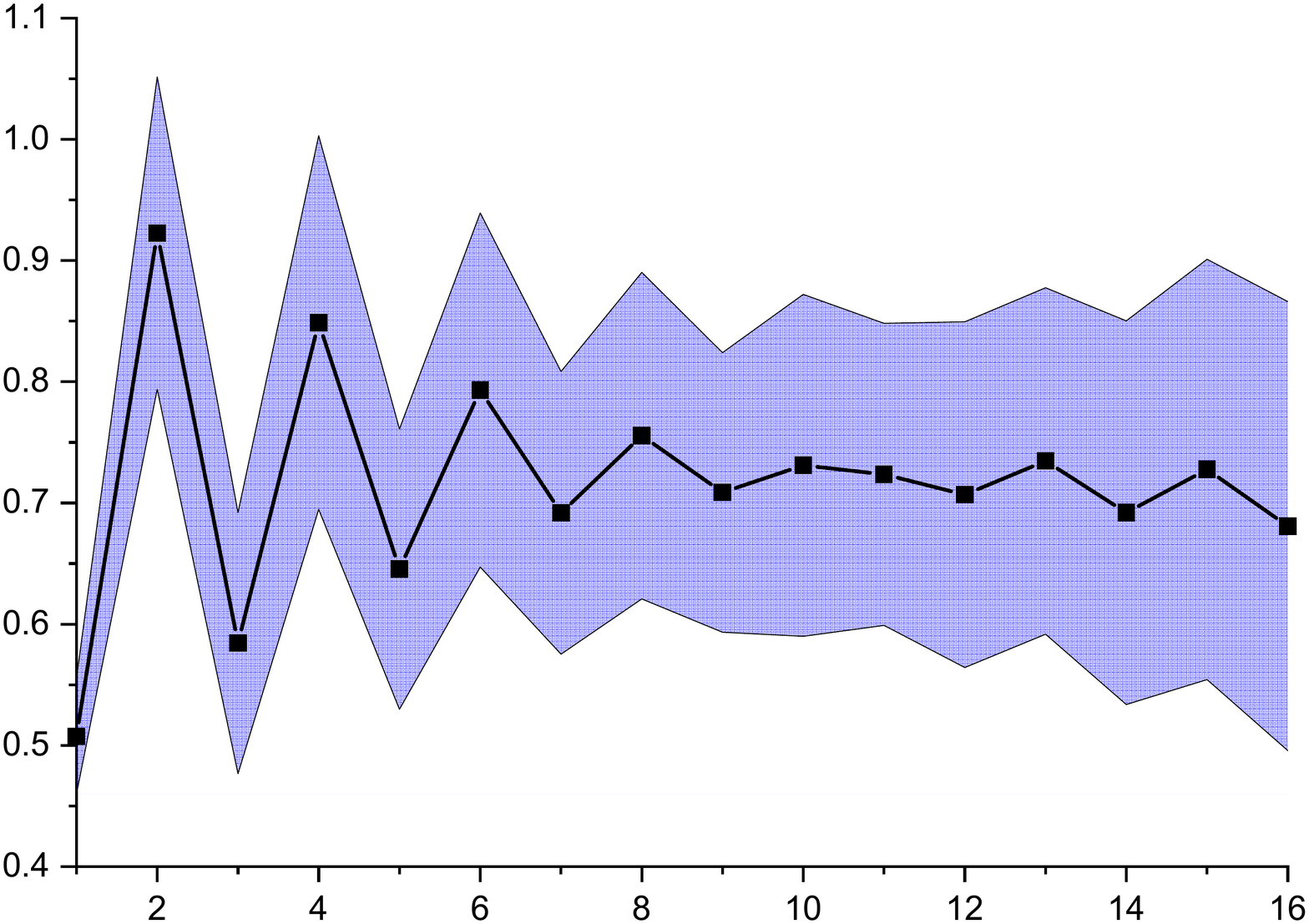}\vspace{4pt}
\includegraphics[width=1\linewidth]{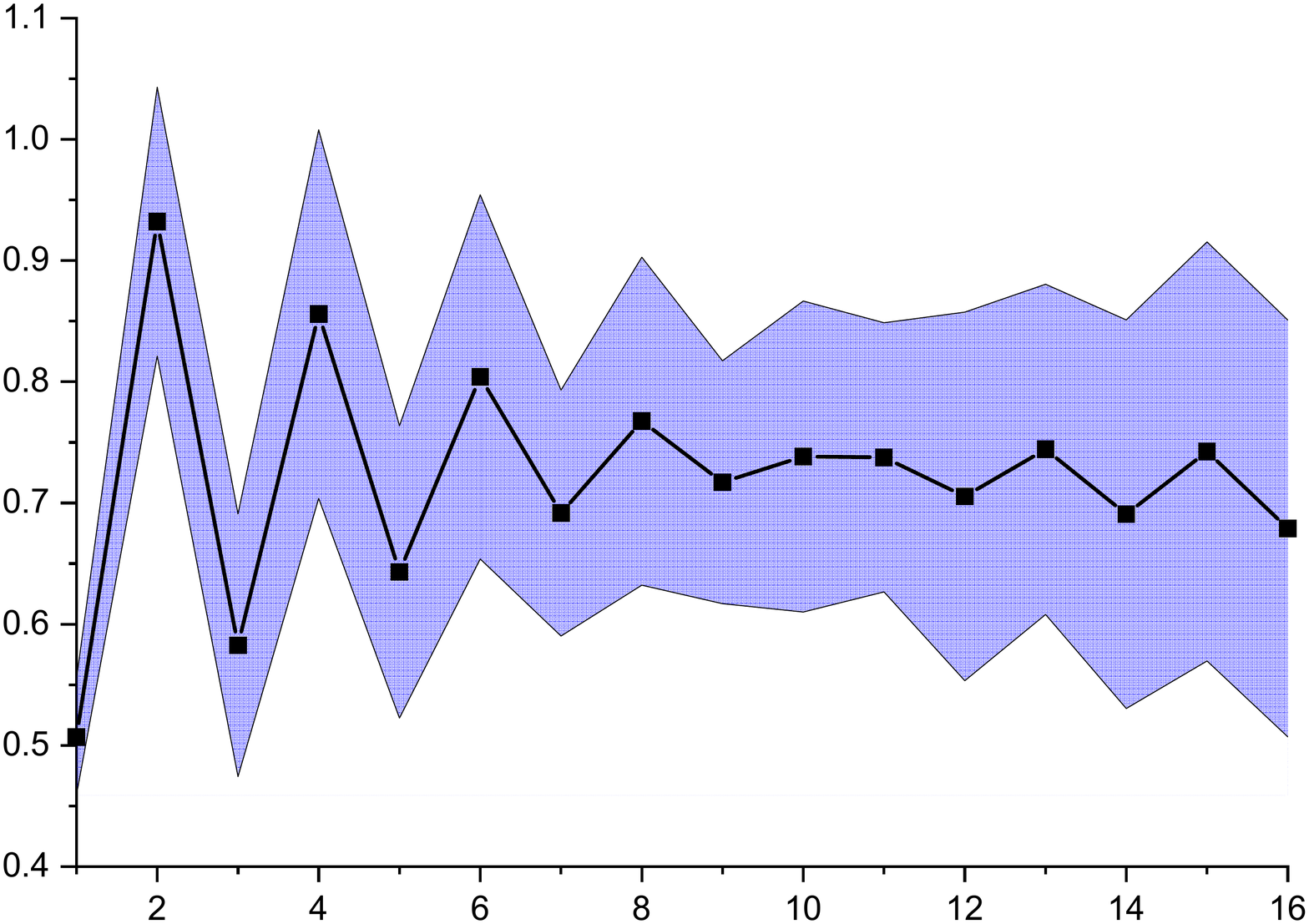}\vspace{4pt}
\includegraphics[width=1\linewidth]{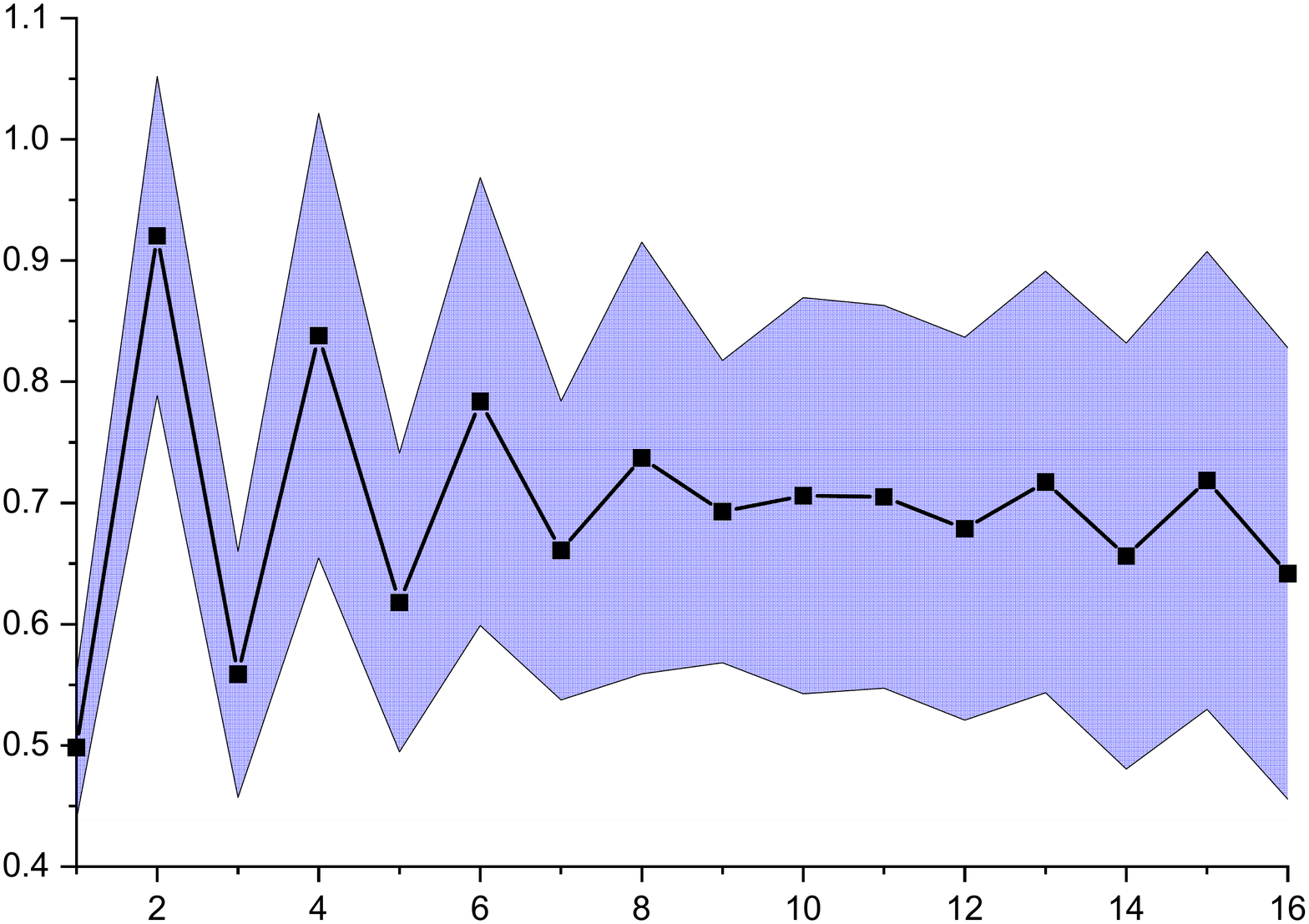}
\end{minipage}
}
}
\vspace{-3mm}
  \caption{ Statistics (mean $\pm$ standard deviation) of MBF coefficients on train (top row), test (median row), and out-of-sample (bottom row) set of ImageNet-VGG-16. 
  }
  \label{fig:vgg MBF}
  \vspace{-1.3mm}
\end{figure}

\vspace{-0.4em}
\noindent
{\bf Three comparison cases and evaluation metrics}
We conduct experiments of three cases, including:
{\bf 1)} \textit{non-transfer}, both training and testing adversarial examples are crafted by the same attack method;
{\bf 2)} \textit{attack-transfer}, both training and testing adversarial examples are crafted by different attack methods;
{\bf 3)} \textit{data-transfer}, both training and testing adversarial examples are crafted by the same attack method, but the data sources of training and testing benign examples are different.
Two widely used metrics are used to evaluate the detection performance, including area under the receiver operating characteristics (AUROC), and detection accuracy, which is the diagonal summation of the confusion matrix, using $0.5$ as the threshold of the posterior probability.
Higher values of both metrics indicate better performance.

\renewcommand{\arraystretch}{0.85}
\begin{table*}[t]
\centering
\caption{Detection results in the {\it non-transfer} case.}
\scalebox{0.66}{
\begin{tabular}{c|l|c c c c|c c c c}
\hline
\multirow{2}{*}{database} & \multirow{2}{*}{Detector} & \multicolumn{4}{c|}{AUROC $(\%)$} & \multicolumn{4}{c}{Accuracy $(\%)$}
\\
\cline{3-10}
 &  &  BIM & CW-L2 & DeepFool & R-PGD & BIM & CW-L2 & DeepFool & R-PGD
\\ \hline
\multirow{4}{*}{CIFAR-10}
 &KD+BU \cite{KD-BU-arxiv-2017}   &$79.0$  &$82.8$  &$80.6$  &$77.7$ &$74.2$  &$73.0$  &$71.7$  &$72.6$ \\
 &M-D \cite{M-distance-nips-2018}       &$51.7$  &$48.4$  &$53.7$  &$52.0$ &$66.7$  &$66.7$  &$66.7$  &$66.7$  \\
 &LID \cite{LID-iclr-2018}         &$87.9$  &$87.4$  &$86.6$  &$82.7$  &$72.6$  &$78.5$  &$78.0$  &$70.2$ \\
 &MBF           &\textbf{99.6} &\textbf{99.6} &\textbf{96.9} &\textbf{99.4} &\textbf{98.8} &\textbf{98.3} &\textbf{91.8} &\textbf{98.5}  \\ \hline
\multirow{4}{*}{SVHN}
 &KD+BU \cite{KD-BU-arxiv-2017}       &$81.5 $  &$85.1 $  &$84.1 $  &$82.5 $ &$78.6 $  &$80.0 $  &$79.3 $  &$78.6 $  \\
&M-D    \cite{M-distance-nips-2018}     &$49.7 $  &$50.0 $  &$49.6 $  &$50.1 $ &$66.7 $  &$66.7 $  &$66.7 $  &$66.7 $  \\
&LID \cite{LID-iclr-2018}        &$91.7 $  &$88.7 $  &$92.2 $  &$90.5 $ &$83.3 $  &$80.8 $  &$84.0 $  &$81.5 $  \\
 &MBF           &\textbf{99.7} &\textbf{99.9} &\textbf{99.3} &\textbf{99.5} &\textbf{98.8} &\textbf{98.3} &\textbf{91.8} &\textbf{98.5} \\ \hline
 &KD+BU \cite{KD-BU-arxiv-2017}       &$50.8$  &$52.6$    &$51.1$  &$51.5$  &$34.4$  &$36.8$    &$34.9$  &$35.3$  \\
ImageNet- &M-D \cite{M-distance-nips-2018} &$48.1$  &$60.1$    &$46.5$  &$54.1$ &$66.7$  &$66.6$    &$66.8$  &$66.7$  \\
AlexNet &LID \cite{LID-iclr-2018}&$71.7$  &$70.9$    &$71.9$  &$72.3$  &$68.7$  &$60.5$    &$65.8$  &$68.3$  \\
 &MBF           &\textbf{99.9} &\textbf{99.6} &\textbf{99.9} &\textbf{99.8}    &\textbf{98.6} &\textbf{97.8} &\textbf{98.8} &\textbf{98.4} \\ \hline
 &KD+BU \cite{KD-BU-arxiv-2017}       &$57.1$  &$57.7$    &$56.2$  &$58.6$  &$42.8$  &$43.6$    &$41.7$  &$44.8$  \\
ImageNet- &M-D \cite{M-distance-nips-2018}  &$63.7$  &$64.8$    &$45.9$  &$66.9$  &$67.1$  &$65.6$    &$66.7$  &$67.4$  \\
VGG16 &LID \cite{LID-iclr-2018}&$82.6$  &$84.2$    &$89.5$  &$84.2$  &$77.8$  &$76.7$    &$83.3$  &$76.4$ \\
 &MBF          &\textbf{99.8}  &\textbf{100.0}    &\textbf{100.0}  &\textbf{100.0} &\textbf{99.6}  &\textbf{99.6}    &\textbf{100.0}  &\textbf{100.0}
 \\ \hline
\end{tabular}
}
\label{table: non-transfer results}
\vspace{-5mm}
\end{table*}

\renewcommand{\arraystretch}{0.9}
\begin{table*}[!t]
\centering
\caption{Detection results evaluated by AUROC ($\%$) in the {\it attack-transfer} case.}
\vspace{1mm}
\scalebox{0.616}{
\begin{tabular}{c|c|c c c c}
\hline
\multirow{2}{*}{database} & Test attack $\rightarrow$ & BIM & CW-L2 & DeepFool & R-PGD
\\
\cline{2-2}
 & Train attack $\downarrow$ & \multicolumn{4}{c}{KD+BU \cite{KD-BU-arxiv-2017}  /   M-D \cite{M-distance-nips-2018}  /   LID \cite{LID-iclr-2018}  /   MBF}
 \\
 \hline
\multirow{4}{*}{CIFAR-10} & BIM & 79.0 / 51.7 / 87.9 / \textbf{99.6} & 78.1 / 52.2 / 85.7 / \textbf{99.4} & 76.4 / 52.4 / 86.4 / \textbf{87.8} & 78.6 / 50.9 / 87.6 / \textbf{99.4} \\
 & CW-L2 & 83.4 / 47.9 / 90.3 / \textbf{99.6} & 82.8 / 48.4 / 87.4 / \textbf{99.6} & 80.3 / 46.5 / 87.0 / \textbf{88.6} & 82.9 / 48.2 / 88.9 / \textbf{99.5} \\
& DeepFool & 83.6 / 53.5 / 89.7 / \textbf{99.3} & 83.1 / 51.4 / 86.1 / \textbf{99.1} & 80.6 / 53.7 / 86.6 / \textbf{96.9} & 83.1 / 53.6 / 87.9 / \textbf{98.9} \\
& R-PGD & 78.1 / 51.5 / 86.1 / \textbf{99.6} & 77.3 / 51.9 / 84.5 / \textbf{99.5} & 75.8 / 52.4 / 84.9 / \textbf{88.3} & 77.7 / 52.1 / 82.7 / \textbf{99.4} \\
 \hline
\multirow{4}{*}{SVHN} & BIM & 81.5 / 49.7 / 91.7 / \textbf{99.7} & 83.1 / 51.0 / 88.3 / \textbf{99.7} & 80.1 / 48.8 / 91.9 / \textbf{97.5} & 81.5 / 50.4 / 90.7 / \textbf{99.5} \\
 & CW-L2 & 83.6 / 50.8 / 91.6 / \textbf{99.7} & 85.1 / 50.0 / 88.7 / \textbf{99.9} & 82.5 / 49.7 / 92.0 / \textbf{97.3} & 83.6 / 49.1 / 90.8 / \textbf{99.7} \\
& DeepFool & 85.0 / 50.0 / 91.6 / \textbf{99.7} & 86.3 / 50.0 / 87.9 / \textbf{99.8} & 84.1 / 49.6 / 92.2 / \textbf{99.3} & 85.0 / 49.4 / 90.8 / \textbf{99.6} \\
& R-PGD & 82.5 / 49.9 / 91.2 / \textbf{99.6} & 84.0 / 49.5 / 87.8 / \textbf{99.6} & 81.2 / 50.2 / 91.6 / \textbf{97.4} & 82.5 / 50.1 / 90.5 / \textbf{99.5} \\
 \hline
 & BIM & 50.8 / 48.1 / 71.7 / \textbf{99.9} & 50.8 / 44.9 / 70.8 / \textbf{99.6} & 50.4 / 49.7 / 72.1 / \textbf{99.8} & 50.7 / 50.4 / 70.8 / \textbf{99.8} \\
ImageNet- & CW-L2 & 52.3 / 50.7 / 69.7 / \textbf{99.8} & 52.6 / 60.1 / 70.9 / \textbf{99.7} & 51.2 / 52.1 / 66.0 / \textbf{99.7} & 52.2 / 55.8 / 69.8 / \textbf{99.8} \\
AlexNet & DeepFool & 51.8 / 53.6 / 71.3 / \textbf{99.9} & 51.9 / 53.5 / 69.7 / \textbf{99.7} & 51.2 / 46.5 / 71.9 / \textbf{99.9} & 51.8 / 51.2 / 71.2 / \textbf{99.8} \\
& R-PGD & 51.6 / 52.4 / 72.5 / \textbf{99.8} & 51.3 / 50.4 / 72.3 / \textbf{99.6} & 50.3 / 50.5 / 72.2 / \textbf{99.7} & 51.5 / 54.1 / 72.3 / \textbf{99.8} \\
 \hline
 & BIM & 57.1 / 63.7 / 82.6 / \textbf{99.8} & 57.1 / 57.2 / 84.3 / \textbf{99.5} & 53.8 / 54.4 / 90.4 / \textbf{100.0} & 57.1 / 65.4 / 84.2 / \textbf{100.0} \\
ImageNet- & CW-L2 & 57.3 / 64.3 / 81.9 / \textbf{100.0} & 57.7 / 64.8 / 84.2 / \textbf{100.0} & 53.8 / 56.6 / 90.2 / \textbf{100.0} & 57.5 / 67.6 / 84.3 / \textbf{100.0} \\
VGG-16 & DeepFool & 58.7 / 45.5 / 80.2 / \textbf{99.7} & 59.4 / 46.1 / 79.1 / \textbf{99.3} & 56.2 / 45.9 / 89.5 / \textbf{100.0} & 59.0 / 44.0 / 80.9 / \textbf{99.7} \\
& R-PGD & 58.6 / 62.3 / 81.1 / \textbf{100.0} & 58.9 / 60.8 / 83.7 / \textbf{100.0} & 55.3 / 54.4 / 90.8 / \textbf{100.0} & 58.6 / 66.9 / 84.2 / \textbf{100.0} \\
\hline
\end{tabular}}
\label{table: attack-transfer AUROC results}
\vspace{-7mm}
\end{table*}

\renewcommand{\arraystretch}{0.9}
\begin{table*}[!t]
\centering
\caption{Detection results evaluated by accuracy ($\%$) in the {\it attack-transfer} case.}
\vspace{1mm}
\scalebox{0.63}{
\begin{tabular}{c|c|c c c c}
\hline
\multirow{2}{*}{database} & Test attack $\rightarrow$ & BIM & CW-L2 & DeepFool & R-PGD
\\
\cline{2-2}
 & Train attack $\downarrow$ & \multicolumn{4}{c}{KD+BU  \cite{KD-BU-arxiv-2017}  /   M-D \cite{M-distance-nips-2018}  /  LID  \cite{LID-iclr-2018}   /   MBF}
 \\
 \hline
\multirow{4}{*}{CIFAR-10} & BIM & 74.2 / 66.7 / 72.6 / \textbf{98.8} & 73.0 / 66.7 / 66.9 / \textbf{97.4} & 71.7 / 66.7 / 70.2 / \textbf{87.3} & 73.8 / 66.7 / 61.3 / \textbf{96.8} \\
 & CW-L2 & 74.2 / 66.7 / 80.3 / \textbf{98.1} & 73.0 / 66.7 / 78.5 / \textbf{98.3} & 71.7 / 66.7 / 77.9 / \textbf{89.5} & 73.8 / 66.7 / 73.6 / \textbf{98.2} \\
& DeepFool & 74.0 / 66.7 / 80.4 / \textbf{95.2} & 72.8 / 66.7 / 77.4 / \textbf{95.2} & 71.7 / 66.7 / 78.0 / \textbf{91.8} & 73.4 / 66.7 / 74.8 / \textbf{95.4} \\
& R-PGD & 73.1 / 66.7 / 77.7 / \textbf{98.6} & 71.9 / 66.7 / 75.1 / \textbf{98.5} & 71.0 / 66.7 / 75.4 / \textbf{89.7} & 72.6 / 66.7 / 70.2 / \textbf{98.5} \\
 \hline
\multirow{4}{*}{SVHN} & BIM & 78.6 / 66.7 / 83.3 / \textbf{98.7} & 80.5 / 66.7 / 79.4 / \textbf{98.9} & 77.7 / 66.7 / 83.5 / \textbf{95.8} & 78.6 / 66.7 / 82.9 / \textbf{98.2} \\
 & CW-L2 & 78.2 / 66.7 / 83.5 / \textbf{98.0} & 80.0 / 66.7 / 80.8 / \textbf{99.2} & 77.1 / 66.7 / 84.0 / \textbf{95.9} & 78.1 / 66.7 / 80.7 / \textbf{97.3} \\
& DeepFool & 79.8 / 66.7 / 83.9 / \textbf{98.0} & 81.8 / 66.7 / 79.9 / \textbf{98.4} & 79.3 / 66.7 / 84.0 / \textbf{97.3} & 79.9 / 66.7 / 80.7 / \textbf{98.0} \\
& R-PGD & 78.6 / 66.7 / 83.6 / \textbf{98.6} & 80.5 / 66.7 / 79.9 / \textbf{98.9} & 77.7 / 66.7 / 83.8 / \textbf{96.5} & 78.6 / 66.7 / 81.5 / \textbf{98.8} \\
 \hline
 & BIM & 34.4 / 66.7 / 68.7 / \textbf{98.6} & 34.4 / 66.6 / 68.6 / \textbf{97.3} & 34.1 / 66.7 / 66.7 / \textbf{98.5} & 34.3 / 66.7 / 65.0 / \textbf{98.4} \\
ImageNet- & CW-L2 & 36.5 / 66.2 / 64.6 / \textbf{98.0} & 36.8 / 66.6 / 60.6 / \textbf{97.8} & 35.0 / 66.1 / 60.7 / \textbf{98.1} & 36.4 / 66.1 / 62.6 / \textbf{98.0} \\
AlexNet & DeepFool & 36.0 / 66.7 / 69.4 / \textbf{98.8} & 36.0 / 66.8 / 65.8 / \textbf{97.9} & 34.9 / 66.8 / 65.8 / \textbf{98.8} & 35.9 / 66.6 / 68.3 / \textbf{98.6} \\
& R-PGD & 35.4 / 66.7 / 67.5 / \textbf{98.5} & 35.2 / 66.7 / 67.7 / \textbf{97.6} & 34.1 / 66.7 / 67.0 / \textbf{98.3} & 35.3 / 66.7 / 68.3 / \textbf{98.4} \\
 \hline
 & BIM & 42.8 / 67.1 / 77.8 / \textbf{98.6} & 42.8 / 66.8 / 74.5 / \textbf{97.3} & 38.4 / 66.0 / 82.7 / \textbf{98.5} & 42.7 / 67.9 / 76.9 / \textbf{98.4} \\
ImageNet- & CW-L2 & 43.1 / 64.8 / 75.2 / \textbf{98.0} & 43.6 / 65.6 / 76.7 / \textbf{97.8} & 38.4 / 62.5 / 82.5 / \textbf{98.1} & 43.3 / 67.2 / 73.9 / \textbf{98.0} \\
VGG-16 & DeepFool & 45.4 / 66.7 / 76.2 / \textbf{98.8} & 46.1 / 66.7 / 72.9 / \textbf{97.9} & 41.8 / 66.7 / 83.3 / \textbf{98.8} & 45.8 / 66.7 / 74.4 / \textbf{98.6} \\
& R-PGD & 44.8 / 66.3 / 76.3 / \textbf{98.5} & 45.4 / 66.6 / 74.1 / \textbf{97.6} & 40.6 / 64.9 / 82.4 / \textbf{98.3} & 45.8 / 67.4 / 76.4 / \textbf{98.4} \\
\hline
\end{tabular}
}
\label{table: attack-transfer accuracy results}
\vspace{-5.5mm}
\end{table*}

\vspace{-1.5mm}
\subsection{Results}
\label{sec: subsec results}
\vspace{-1.5mm}


Detection results in the {\it non-transfer} case are shown in Table \ref{table: non-transfer results}.
The proposed MBF method shows the best performance in all cases, and is much superior to all compared methods.
LID performs the second-best in most case.
KD+BU gives somewhat good detection performance on CIFAR-10 and SVHN, but performs very poor on ImageNet.
It tells that KD+BU is very sensitive to different databases and networks.
M-D gives almost the degenerate results. Note that the results of M-D on CIFAR-10 and SVHN reported in \cite{M-distance-nips-2018} are very high, but their networks (\ie, ResNet \cite{resnet} and DenseNet \cite{densenet} are different from that used in our experiments. It implies that M-D may be suitable for very deep neural networks, but not for shallower networks.

Detection results in the {\it attack-transfer} case evaluated by AUROC and accuracy are presented in Tables \ref{table: attack-transfer AUROC results} and \ref{table: attack-transfer accuracy results}, respectively.
MBF still shows much better performance in all transfer cases than all compared methods, and the changes among detecting different attacks are very small. It verifies the robustness of MBF to different attacks.

We also conduct a {\it data-transfer} experiment on ImageNet. Specifically, we collect extra 365 images of 8 classes (same with the classes used for the detection training, see Section \ref{sec: subsec experimental settings}) through searching the class names in Baidu and Facebook. These 365 benign and their noisy images can be correctly predicted by both the fine-tuned AlexNet and VGG models. The results on these images are shown in Table \ref{ImageNet_database_transfer_auc}. MBF still shows the best performance.
And, compared to the corresponding results in Table \ref{table: non-transfer results}, the AUROC/accuracy scores of MBF on detecting different attacks change very gently, verifying its robustness to different data sources.

\vspace{-1mm}
\subsection{Hypothesis Test via Kolmogorov-Smirnov Test}
\vspace{-1mm}
\label{sec: ks test}
\noindent
{\bf Kolmogorov-Smirnov test} (KS test) \cite{ks-test-1951} is a non-parametric test method in statistics, to test whether a sample follows a reference probability distribution (one-sample KS test), or whether two samples follow the same distribution (two-sample KS test).
Specifically, in one-sample KS test, the distance between the empirical distribution function of one sample and the cumulative distribution function of the reference probability distribution is measured. Then, the $p$-value corresponding to the obtained distance is computed. If the $p$-value is larger than the significance level $\alpha$ (here we set $\alpha=0.05$), then the null hypothesis that the sample follows the reference distribution is accepted; otherwise, rejected.
Similarly, in two-sample KS test, the distance between the empirical distribution functions of two samples is computed. If the corresponding $p$-value is larger than $\alpha$, then it accepts that two samples follow the same distribution.
The KS test conducted below is implemented by the python function
$scipy.stats.ks\_2samp$\footnote{https://docs.scipy.org/doc/scipy-0.14.0/reference/generated/scipy.stats.ks\_2samp.html}.

\noindent
{\bf Hypothesis test 1}. Here we verify that whether the posterior vectors of both adversarial and benign examples follow GGD.
We denote the posterior vector of one adversarial example as $\mathbf{p}_{\text{adv}}$, and that of one benign example as $\mathbf{p}_{\text{ben}}$. The distribution of GGD is denoted as $\mathcal{P}_{\text{GGD}}$.
Then, we conduct two one-sample KS tests, including:
\vspace{-1.5 mm}
\begin{itemize}
    \vspace{-1.5 mm}
    \item {\bf H1.1} The test of adversarial examples: {\it H$_0$}: $\mathbf{p}_{\text{adv}} \sim \mathcal{P}_{\text{GGD-adv}}$; ~{\it H$_1$}: $\mathbf{p}_{\text{adv}} \not\sim \mathcal{P}_{\text{GGD-adv}}$.
    \vspace{-1.5 mm}
    \item {\bf H1.2} The test of benign examples: {\it H$_0$}: $\mathbf{p}_{\text{ben}} \sim \mathcal{P}_{\text{GGD-ben}}$; ~{\it H$_1$}: $\mathbf{p}_{\text{ben}} \not\sim \mathcal{P}_{\text{GGD-ben}}$.
    \vspace{-1.5 mm}
\end{itemize}
\vspace{-1.5 mm}
In {\bf H1.1}, the reference distribution $\mathcal{P}_{\text{GGD-adv}}$ is firstly estimated from $\mathbf{p}_{\text{adv}}$, using the estimated method proposed in \cite{estimateGGD}.
To alleviate the uncertainty of the estimation, we draw $500$ samples from the estimated $\mathcal{P}_{\text{GGD-adv}}$. Then, we conduct the two-sample KS test between $\mathbf{p}_{\text{adv}}$ and these $500$ samples respectively.
\comment{
\hl{However, when the parameters of GGD are determined from the given samples, the $p$-values determined by KS test are invalid \cite{ks-test-1951}.
As a result, we have to specify the shape parameter of GGD as $2$  (}\ie  \hl{Gaussian distribution) and choose the Anderson-Darling test (AD test) \cite{AD_test} for Gaussian distribution with unknown mean and variance. The Anderson-Darling test is a modification of KS test, which used to test if a sample of data came from a population with a specific distribution. We implement AD test by the python function $statsmodels.stats.diagnostic.normal\_ad$\footnote{https://www.statsmodels.org/stable/generated/statsmodels.stats.\\diagnostic.normal\_ad.html}, which only supports the test of Gaussian distribution.
}
}
The average $p$-value over $1000$ tests is recorded.
The mean of the average $p$-values over the whole database is reported.
{\bf H1.2} is conducted similarly.
The results tested on ImageNet-AlexNet are shown in Table \ref{Hypothesis test 1}.
Due to the space limit, results on other databases and models will be presented in the {\bf supplementary material B}.
In all cases, the $p$-values are larger than the significance level $0.05$.
Hence, we can conclude that {\it the posterior vectors of both adversarial and benign examples follow GGD}. However, note that the parameters of their corresponding GGD are different, which will be verified in the following test.

\comment{
\begin{itemize}
    \item {\it H$_0$}: $\mathbf{p}_{adv} \sim \mathcal{P}_{\text{GGD}}$;
    \item {\it H$_1$}: $\mathbf{p}_{adv} \not\sim \mathcal{P}_{\text{GGD}}$.
\end{itemize}

We denote the empirical distribution of the posterior vectors of adversarial examples as  $\mathcal{\hat{P}}_{\text{adv-posterior}}$, and that of benign examples as $\mathcal{\hat{P}}_{\text{benign-pos}}$. The distribution of GGD is denoted as $P_{\text{GGD}}$.
Then, the hypothesis test is
\vspace{-0.8em}
\begin{itemize}
    \item {\it H$_0$}: $\mathcal{\hat{P}}_{\text{adv-pos}} = P_{\text{GGD}};$
    \vspace{-0.2em}
    \item {\it H$_1$}: $\mathcal{\hat{P}}_{\text{adv-pos}} \neq P_{\text{GGD}}.$
    \vspace{-0.8em}
\end{itemize}
}

\noindent
{\bf Hypothesis test 2}. Here we verify that whether the extracted MBF features of adversarial and benign examples follow the same empirical distribution.
We denote the MBF feature vector of one adversarial example as $\mathbf{m}_{\text{adv}}$, and the corresponding empirical distribution is denoted as $\mathcal{\hat{P}}_{\text{adv-MBF}}$. Similarly, we define $\mathbf{m}_{\text{ben}}$ and $\mathcal{\hat{P}}_{\text{ben-MBF}}$ for benign examples.
Then, we conduct the following four two-sample KS tests:
\vspace{-0.8em}
\begin{itemize}
    \item {\bf H2.1} The test between adversarial and benign examples:
        {\it H$_0$}: $\mathcal{\hat{P}}_{\text{adv-MBF}} = \mathcal{\hat{P}}_{\text{ben-MBF}}$; ~
        {\it H$_1$}: $\mathcal{\hat{P}}_{\text{adv-MBF}} \neq \mathcal{\hat{P}}_{\text{ben-MBF}}$.
    \vspace{-0.2em}
    \item {\bf H2.2} The test between adversarial examples crafted from different attack methods:
        {\it H$_0$}: $\mathcal{\hat{P}}_{\text{adv-MBF}}^{\text{attack-1}} = \mathcal{\hat{P}}_{\text{adv-MBF}}^{\text{attack-2}}$; ~
        {\it H$_1$}: $\mathcal{\hat{P}}_{\text{adv-MBF}}^{\text{attack-1}} \neq \mathcal{\hat{P}}_{\text{adv-MBF}}^{\text{attack-2}}$.
    \vspace{-0.2em}
    \item {\bf H2.3} The test between adversarial examples from different data sources (\ie, train, test and out-of-sample set):
        {\it H$_0$}: $\mathcal{\hat{P}}_{\text{adv-MBF}}^{\text{source-1}} = \mathcal{\hat{P}}_{\text{adv-MBF}}^{\text{source-2}}$; ~
        {\it H$_1$}: $\mathcal{\hat{P}}_{\text{adv-MBF}}^{\text{source-1}} \neq \mathcal{\hat{P}}_{\text{adv-MBF}}^{\text{source-2}}$.
    \vspace{-0.2em}
    \item {\bf H2.4} The test between benign examples from different data sources (\ie, train, test and out-of-sample set):
        {\it H$_0$}: $\mathcal{\hat{P}}_{\text{ben-MBF}}^{\text{source-1}} = \mathcal{\hat{P}}_{\text{ben-MBF}}^{\text{source-2}}$; ~
        {\it H$_1$}: $\mathcal{\hat{P}}_{\text{ben-MBF}}^{\text{source-1}} \neq \mathcal{\hat{P}}_{\text{ben-MBF}}^{\text{source-2}}$.
        \vspace{-0.4em}
\end{itemize}

\renewcommand{\arraystretch}{0.95}
\begin{table}[!t]
\centering
\vspace{-2.5 mm}
\caption{Detection results evaluated by AUROC score (\%) in the {\it data-transfer} case.
All detectors are trained on the train set, and tested on the out-of-sample set (including $365$ images) of the ImageNet database.
The best results are highlighted in bold.}
\vspace{-0.5mm}
\scalebox{0.7}
{
\begin{tabular}{>{\centering}+p{0.063\textwidth}|>{\centering}^p{0.028\textwidth} >{\centering}^p{.058\textwidth} >{\centering}^p{.063\textwidth} >{\centering}^p{.058\textwidth}| >{\centering}^p{0.028\textwidth} >{\centering}^p{0.058\textwidth} >{\centering}^p{.063\textwidth} >{\centering\arraybackslash}^p{.058\textwidth}}
\hline
&\multicolumn{4}{c|}{ImageNet-AlexNet} & \multicolumn{4}{c}{ImageNet-VGG-16}
\\\hline
\rowstyle{\footnotesize }\normalsize
Detector & BIM & CW-L2 & DeepFool & R-PGD  & BIM & CW-L2 & DeepFool & R-PGD \\
\hline
 KD+BU        &$66.3$  &$69.4$    &$64.7$  &$68.8$  &$84.0$  &$85.0$    &$79.9$  &$85.2$ \\
 M-D    &$48.7$  &$49.2$    &$50.0$  &$51.0$    &$54.5$  &$50.0$    &$50.4$  &$56.1$ \\
 LID   &$69.1$  &$67.6$    &$72.9$  &$71.4$  &$84.2$  &88.4    &89.2  &$83.7$       \\
 MBF           &\textbf{99.1} &\textbf{99.0} &\textbf{99.6} &\textbf{99.4} &\textbf{99.3}  &\textbf{99.5}    &\textbf{99.3}  &\textbf{99.5} \\ \hline
\end{tabular}}
\label{ImageNet_database_transfer_auc}
\vspace{-2.5 mm}
\end{table}

\vspace{-2.5 mm}
\renewcommand{\arraystretch}{1}
\begin{table}[!t]
\centering
\vspace{-2.5 mm}
\caption{The $p$-value of KS hypothesis test among clean samples, noisy samples, and adversarial samples crafted by four methods, respectively, on ImageNet-AlexNet.
}
\vspace{-0.5mm}
\scalebox{0.815}
{
\begin{tabular}{>{\centering}+p{0.12\textwidth}|>{\centering}^p{0.034\textwidth} >{\centering}^p{.034\textwidth} >{\centering}^p{.034\textwidth} >{\centering}^p{.062\textwidth} >{\centering}^p{.06\textwidth} >{\centering\arraybackslash}^p{.062\textwidth}}
\hline
\rowstyle{\normalsize }\scriptsize
ImageNet-AlexNet & clean & noisy & BIM & CW-L2 & DeepFool & R-PGD \\
\hline
train set  &$0.116$ &$0.116$ &$0.253$ &$0.235$ &$0.242$ &$0.253$\\
test set   &$0.117$ &$0.117$ &$0.261$ &$0.236$ &$0.243$ &$0.259$\\  \hline
\end{tabular}}
\label{Hypothesis test 1}
\vspace{-2mm}
\end{table}

\renewcommand{\arraystretch}{1}
\begin{table}[!t]
\centering
\vspace{-1 mm}
\caption{The $p$-value of two-sample KS test among MBF coefficients of different types of examples.}
\vspace{0. mm}
\scalebox{0.62}
{
\begin{tabular}{>{\centering}+m{0.16\textwidth}|>{\centering}+m{0.11\textwidth} >{\centering}+m{0.11\textwidth} >{\centering}+m{0.14\textwidth} >{\centering\arraybackslash}+m{0.149\textwidth}}
\hline
ImageNet-VGG-16 & (clean, noisy) & (clean, BIM) & (BIM, DeepFool) & (CW-L2, R-PGD)  \\
\hline
{\fontsize{14}{14}\selectfont train set}      & {\fontsize{14}{14}\selectfont $0.998$}  & {\fontsize{14}{14}\selectfont $0.000$}  &{\fontsize{14}{14}\selectfont $0.000$}  &{\fontsize{14}{14}\selectfont$0.236$}\\
{\fontsize{14}{14}\selectfont test set}       & {\fontsize{14}{14}\selectfont $1.000$}  & {\fontsize{14}{14}\selectfont $0.000$}  &{\fontsize{14}{14}\selectfont $0.000$}  &{\fontsize{14}{14}\selectfont $0.225$}\\
\hline

\end{tabular}}
\label{Hypothesis test 2.1}
\vspace{-1.8 mm}
\end{table}

\renewcommand{\arraystretch}{1}
\begin{table}[!t]
\centering
\vspace{-5.5mm}
\caption{The $p$-value of KS hypothesis test among MBF coefficients of different data sources.}
\vspace{0.2 mm}
\scalebox{0.67}
{
\begin{tabular}{>{\centering}+p{0.23\textwidth}|>{\centering}+p{0.05\textwidth} >{\centering}+p{0.05\textwidth} >{\centering}+p{0.05\textwidth} >{\centering}+p{0.062\textwidth} >{\centering}+p{0.062\textwidth} >{\centering\arraybackslash}+p{0.077\textwidth}}
\hline
\large
ImageNet-VGG-16& clean & noisy & BIM & CW-L2 & DeepFool & R-PGD \\ \hline
(train set, test set) & {\fontsize{13}{13}\selectfont $0.839$} & {\fontsize{13}{13}\selectfont $0.858$} &{\fontsize{13}{13}\selectfont $0.368$} &{\fontsize{13}{13}\selectfont $0.842$} &{\fontsize{13}{13}\selectfont $0.246$} &{\fontsize{13}{13}\selectfont $0.221$}\\
(train set, out-of-sample set) & {\fontsize{13}{13}\selectfont $0.049$} & {\fontsize{13}{13}\selectfont $0.054$} &{\fontsize{13}{13}\selectfont $0.094$} &{\fontsize{13}{13}\selectfont $0.407$}  &{\fontsize{13}{13}\selectfont $0.737$} &{\fontsize{13}{13}\selectfont $0.088$}\\

\hline
\end{tabular}}
\label{Hypothesis test 2.2}
\vspace{-0.5 mm}
\end{table}

In above four two-sample KS tests, we test on the 16-dimensional MBF features extracted from the soft-max layer. Since the implementation $scipy.stats.ks\_2samp$ cannot compare two vectors, we compare the feature of each dimension separately, then report the average $p$-value over all dimensions.
Specifically, when comparing two sets of samples, we firstly concatenate the feature of each dimension across all samples in the same set, leading to 16 long vectors for each set. Then, each pair of two long vectors corresponding to the same dimension from two sets is compared by KS test. The average $p$-value over all 16 dimensions is reported.
%
%
The $p$-values of {\bf H2.1} are shown in Table \ref{Hypothesis test 2.1} (see the column ``(clean, BIM)"). The $p$-values on both train and test set are $0$. Thus, the hypothesis $H_0$ is rejected, \ie, the MBF features of adversarial and benign examples follow different distributions.
%
The $p$-values of {\bf H2.2} are shown in Table  \ref{Hypothesis test 2.1}. We pick two groups of attack methods, \ie, (BIM, DeepFool) and (CW-L2, R-PGD). The $p$-values of (BIM, DeepFool) are 0, while the $p$-values of (CW-L2, R-PGD) are larger than $0.05$.
It demonstrates that the distributions of adversarial examples crafted from different attack methods are possible to be different.
%
The $p$-values of {\bf H2.3} are shown in Table \ref{Hypothesis test 2.2}.
The $p$-values of all types of adversarial examples exceed 0.05. Thus, the MBF features of adversarial examples from different data sources follow the same distribution.
The $p$-values of {\bf H2.4} are shown in Table \ref{Hypothesis test 2.2}.
Only the $p$-value of ``(train set, out-of-sample set)" of clean examples is slightly lower than 0.05, while the values of other cases exceed 0.05. Thus, in most cases, the MBF features of benign examples from different data sources follow the same distribution.

From above analysis, we obtain the following conclusions:
{\bf 1)} The extracted MBF features of adversarial and benign examples follow different empirical distributions. It explains why MBF features are effective for detecting adversarial and benign examples;
{\bf 2)} The extracted MBF features of adversarial/benign examples from the train, test and out-of-sample sets follow the same empirical distribution.
Although the extracted MBF features of adversarial examples crafted from different attack methods may not follow the same empirical distribution, the significant difference between benign and different adversarial distributions can still lead to the good detection performance in the attack-transfer case.
It explains why MBF features are robust across different attack methods and different data sources.
Moreover, we visualize the statistics of each dimension of MBF features, \ie, mean and standard deviation, as shown in Fig. \ref{fig:vgg MBF}. These visualizations also support above conclusions.
Due to space limit, more KS tests and visualizations on different databases and networks will be presented in the {\bf supplementary material B}.

\vspace{-3mm}
\section{Conclusion}
\vspace{-2mm}
This work has proposed a novel adversarial detection method, dubbed MBF.  The assumption behind is that the internal responses of the classification network of both adversarial and benign examples follow the generalized Gaussian distribution (GGD), but with different shape factors. The magnitude of Benford-Fourier coefficient is a function w.r.t. the shape factor, and can be easily estimated based on responses. Thus, it can serve as the discriminative features between adversarial and benign examples.
The extensive experiments conducted on several databases, as well as the empirical analysis via KS test, demonstrate the superior effectiveness and robustness to different attacks and different data sources of the proposed MBF method, over state-of-the-art detection methods.
\nocite{langley00}

\bibliography{example_paper}
\bibliographystyle{icml2020}

\newpage
\twocolumn[
\icmltitle{Supplementary Material: \\ Effective and Robust Detection of Adversarial Examples via \\Benford-Fourier Coefficients}
\vskip 0.3in
]

\appendix
\section{Proof}
\subsection{Proof of Theorem 1}
 Assume that all the variables $\left \{ \mathcal{T}_m \right \}_{m=1,...,M}$ are independent and identically distributed. Applying the central limit theorem \cite{rosenblatt1956central} to the real and imaginary parts of $\mathcal{Y}$, we can obtain that both parts asymptotically follow the Gaussian distribution
 \begin{flalign}
\mathcal{Y}=\frac{1}{M}\sum_{m=1}^{M}\mathcal{T}_m\sim \mathcal{N}\big(E(\mathcal{T}_1),\;\frac{1}{M}D(\mathcal{T}_1) \big),
\label{eq: clt y}
\end{flalign}

where
\begin{flalign}
E(\mathcal{T}_1) &=E\left ( e^{-j2\pi n\log_{10}\left |\X _1 \right |} \right ) \nonumber\\
&=\int_{-\infty}^{+\infty} \mathcal{P}_{\X_{1}} (x) \cdot e^{-j 2\pi n \log_{10}\left | x \right |} \mathrm{d} x   \nonumber\\
&=a_n,
\label{eq: integral}
\\
\\
D(\mathcal{T}_1) &=E\big ( \left |\mathcal{T}_1\right|^2\big )-\left|E\left ( \mathcal{T}_1\right )\right|^2 \nonumber\\
&=E\big(\left| e^{-j2\pi n\log_{10}\left |\X_1 \right |} \right|^2\big)-|a_n|^2  \nonumber\\
&=1-|a_n|^2.
\label{eq: var integral}
\end{flalign}

The PDF of $\mathcal{Y}$ can be rewritten as follows:
\begin{flalign}
\mathcal{Y}\sim \mathcal{N}(a_n,\frac{1-|a_{n}|^{2}}{M}).
\label{eq: clt y}
\end{flalign}
Thus, we obtain
 \begin{flalign}
\mathcal{E}\sim \mathcal{N}(0,\frac{1-|a_{n}|^{2}}{M}).
\label{eq: e}
\end{flalign}

Besides, the pseudo variance \cite{pseudo} of $\mathcal{T} _1$ is
\begin{flalign}
\mathcal{J}_{\mathcal{T}_1,\mathcal{T}_1}
&=E(\mathcal{T}_1^2)-E(\mathcal{T}_1)^2   \nonumber\\
&=E\big(e^{-j2\pi n\log_{10}\left |\X _1^2 \right |} \big)-a_n^2 \nonumber\\
&=a_{2n}-a_n^2.
\label{eq: pvar integral}
\end{flalign}
Correspondingly,
 \begin{flalign}
\mathcal{J}_{\mathcal{E},\mathcal{E}}=\frac{1}{M}\mathcal{J}_{\mathcal{T} _1,\mathcal{T} _1}=\frac{a_{2n}-a_n^2}{M}.
\label{eq: Ew02}
\end{flalign}
Since $(a_{2n} - a_n^2)$ is bounded, we have $\lim_{M\rightarrow \infty}\mathcal{J}_{\mathcal{E},\mathcal{E}}=0$.
Thus, we obtain that the random variable $\mathcal{E}$ follows circularly-symmetric complex Gaussian distribution, because the sufficient and necessary condition is that mean value and pseudo variance equal zero \cite{pseudo}.
It implies that both the real and imaginary part of $\mathcal{E}$ follow the same Gaussian distribution and they are independent.
Thus, the magnitude of this complex random variable follows the Rayleigh distribution \cite{rayleigh}, and the probability density function of $\left | \mathcal{E} \right |$ can be formulated as
 \begin{flalign}
\P_{\left | \mathcal{E}\right |}\left ( r \right )=\frac{r}{s^2}e^{-r^2/2s^2} ,
\label{eq: pdf w0}
\end{flalign}
where $s$ is the scale parameter.
Knowing the properties of the Rayleigh distribution\cite{rayleigh}, we have:
 \begin{flalign}
s^2=\frac{1}{2}D(\mathcal{E})=\frac{1-|a_n|^2}{2M}.
\end{flalign}

Utilizing the fact that $|a_n|^2$ is close to $0$ when n is a modest number, we obtain that $D\left( \mathcal{E}\right ) \approx \frac{1}{M}$, leading to $s^2 = \frac{1}{2M}$. Then, we obtain
 \begin{flalign}
E\big ( \left | \mathcal{E}\right | \big )=\frac{1}{2}\sqrt\frac{\pi}{M},  D\big(\left | \mathcal{E}\right |\big)=\frac{4-\pi}{4M}.
\label{eq: mean and var for W0 of M}
\end{flalign}
It is easy to observe that both $E\big ( \left | \mathcal{E}\right | \big )$ and $D\big(\left | \mathcal{E}\right |\big)$ are close to zero when the number of samples M increases. It implies that the estimation error $\varepsilon_n$ gets closer to 0 as M increases.

\section{Additional Empirical Analysis}

\begin{figure}[htbp]
\centering
\scalebox{0.5}
{
\subfigure[ clean]{
\begin{minipage}[b]{0.14\textwidth}
\includegraphics[width=1\linewidth]{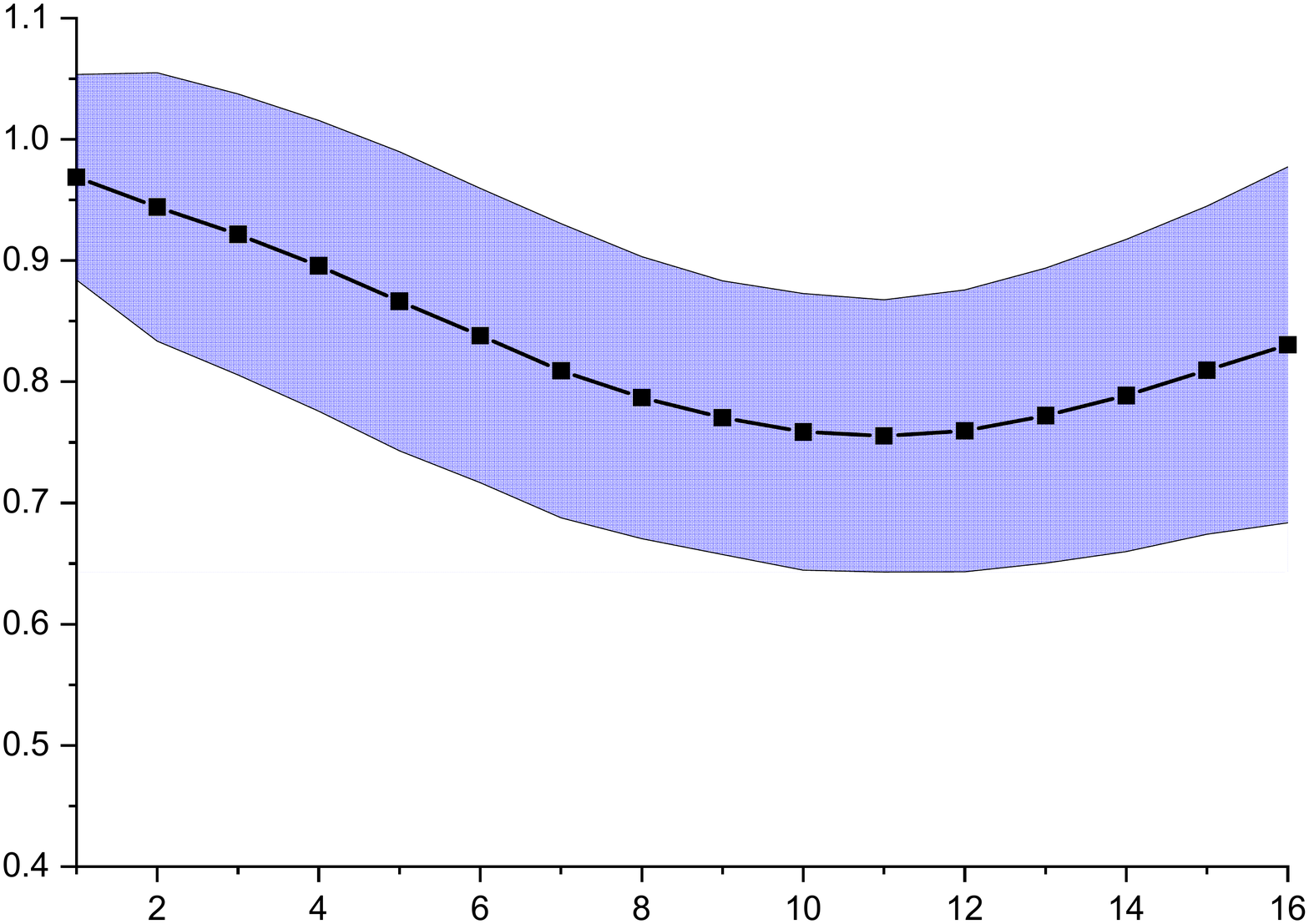}\vspace{4pt}
\includegraphics[width=1\linewidth]{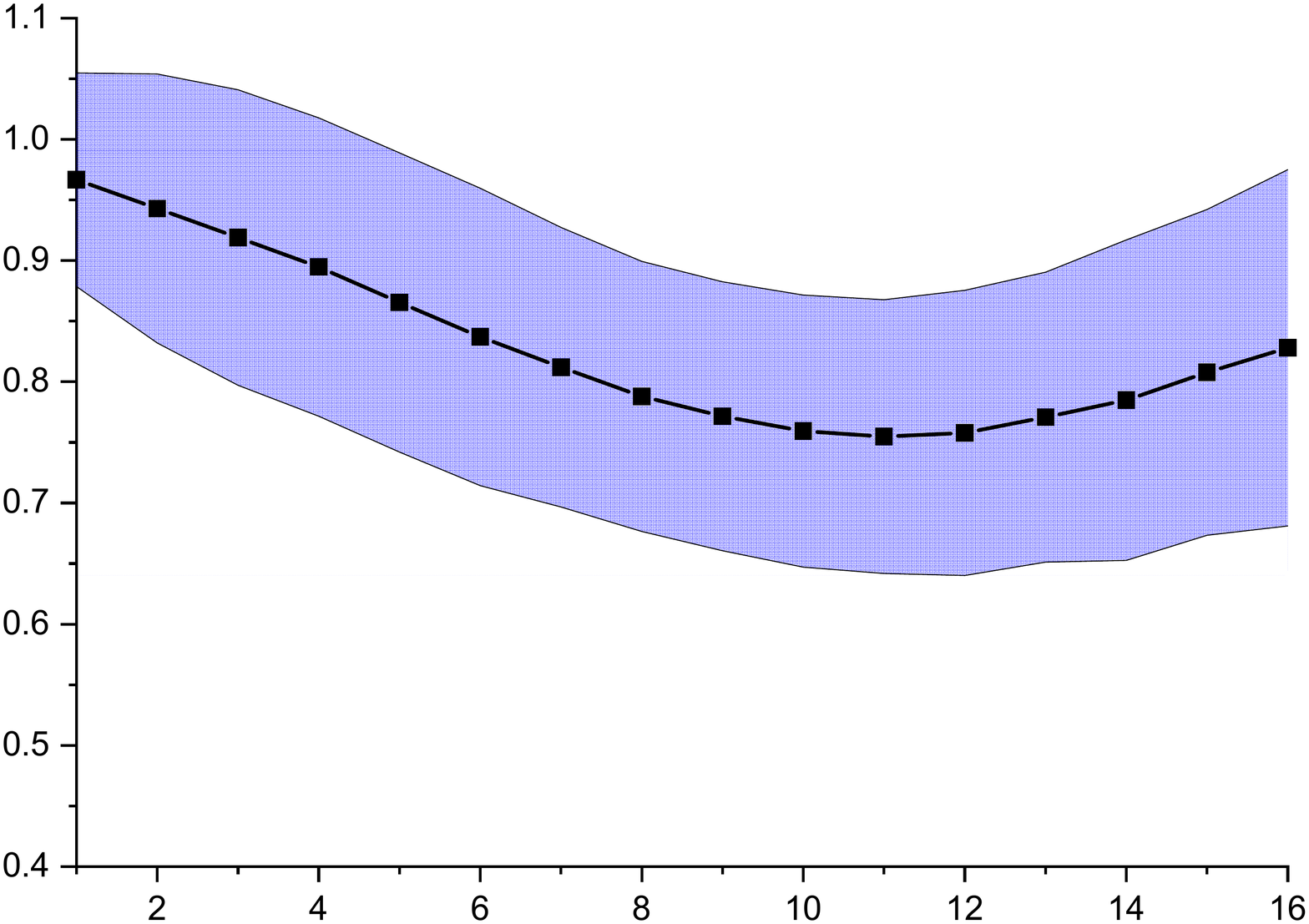}
\end{minipage}
}
\subfigure[ noisy]{
\begin{minipage}[b]{0.14\textwidth}
\includegraphics[width=1\linewidth]{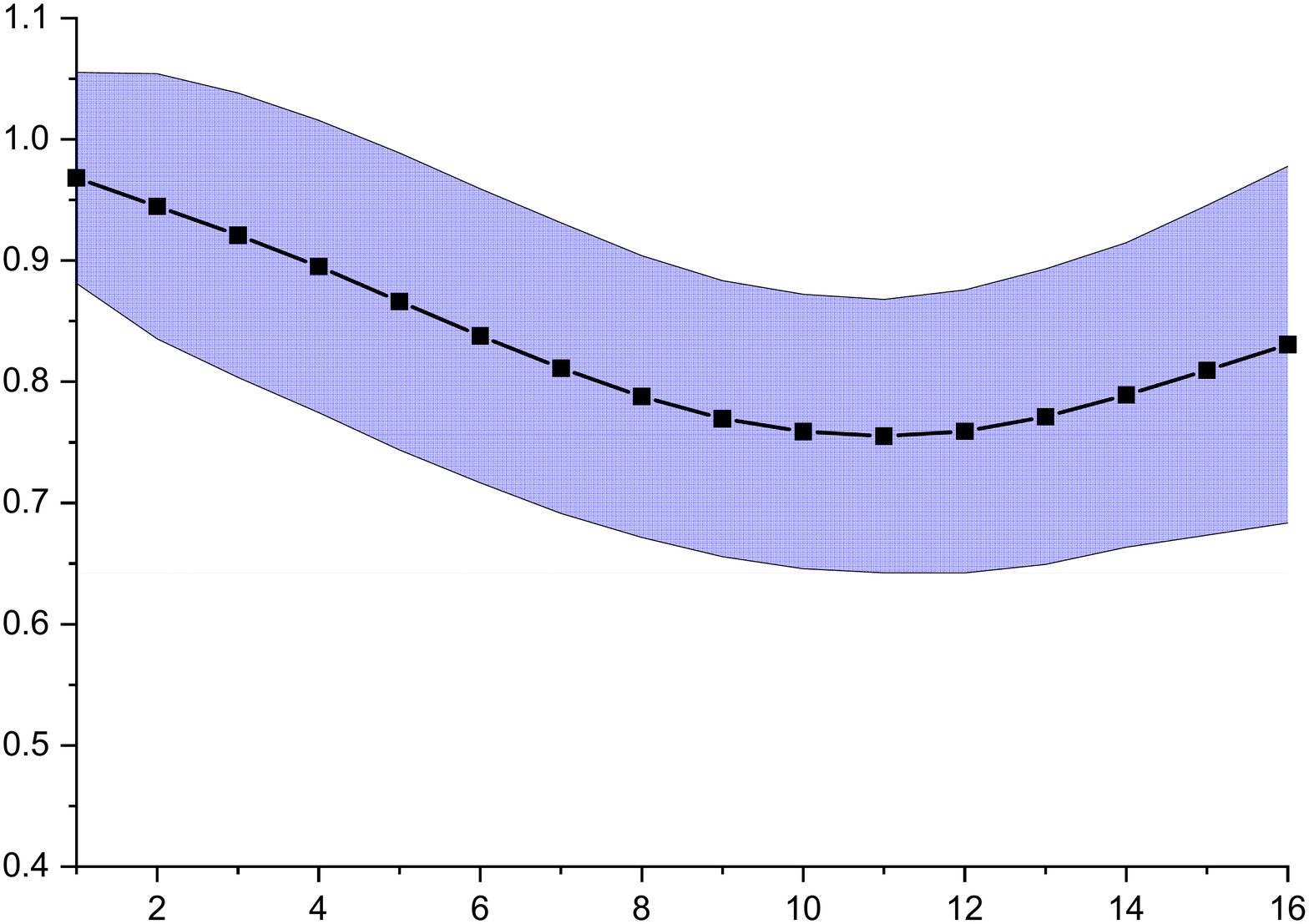}\vspace{4pt}
\includegraphics[width=1\linewidth]{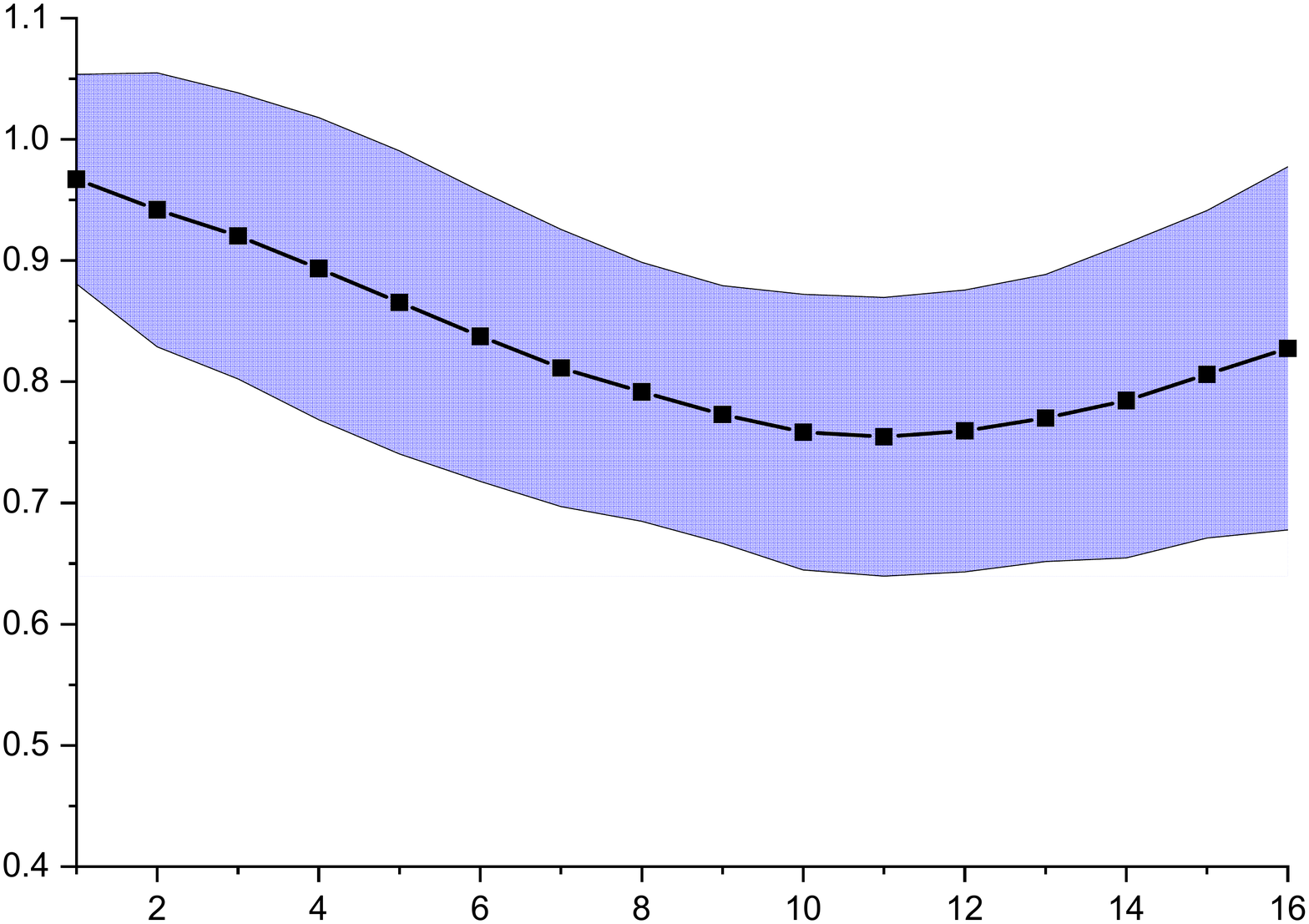}
\end{minipage}
}
\subfigure[  BIM]{
\begin{minipage}[b]{0.14\textwidth}
\includegraphics[width=1\linewidth]{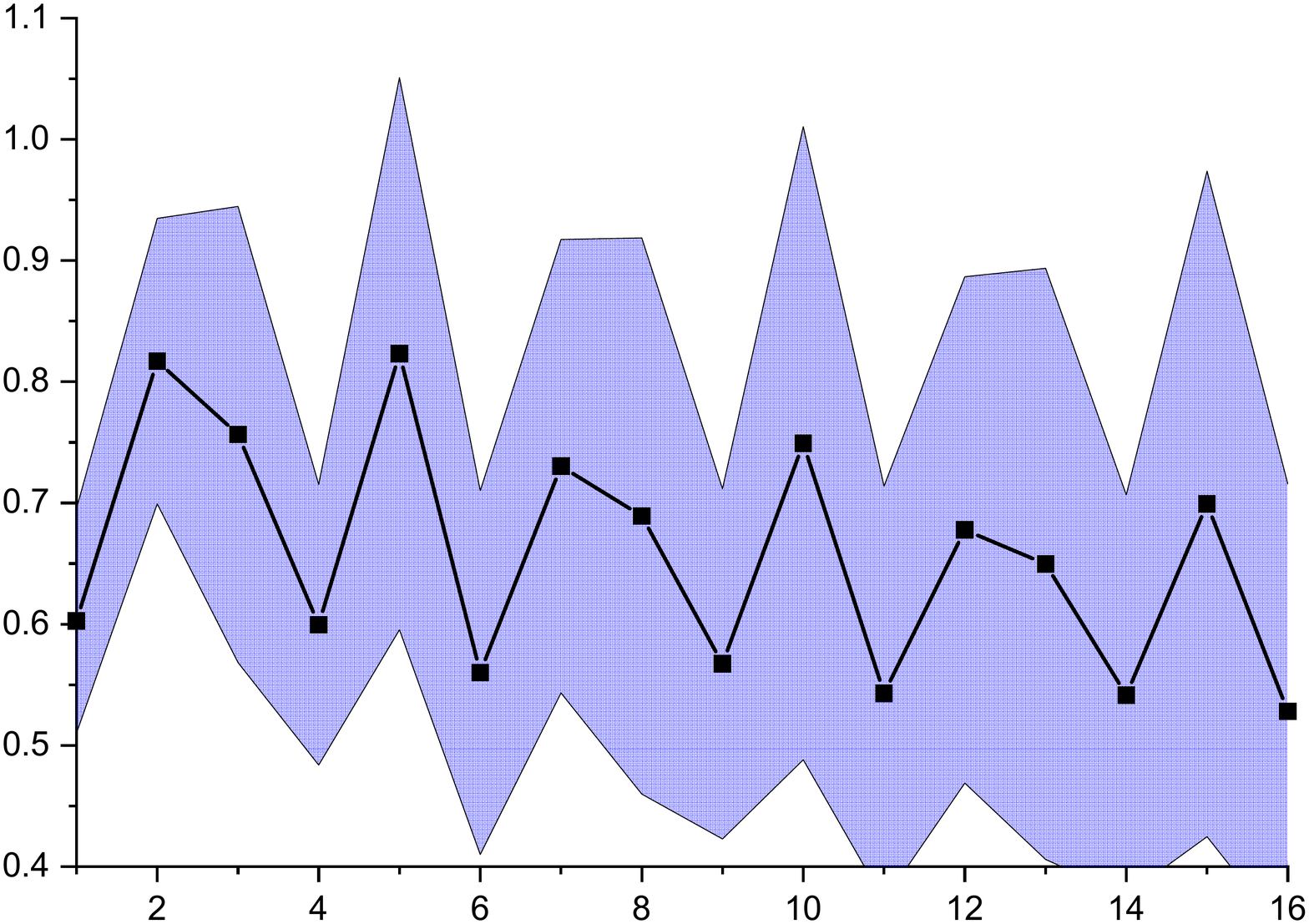}\vspace{4pt}
\includegraphics[width=1\linewidth]{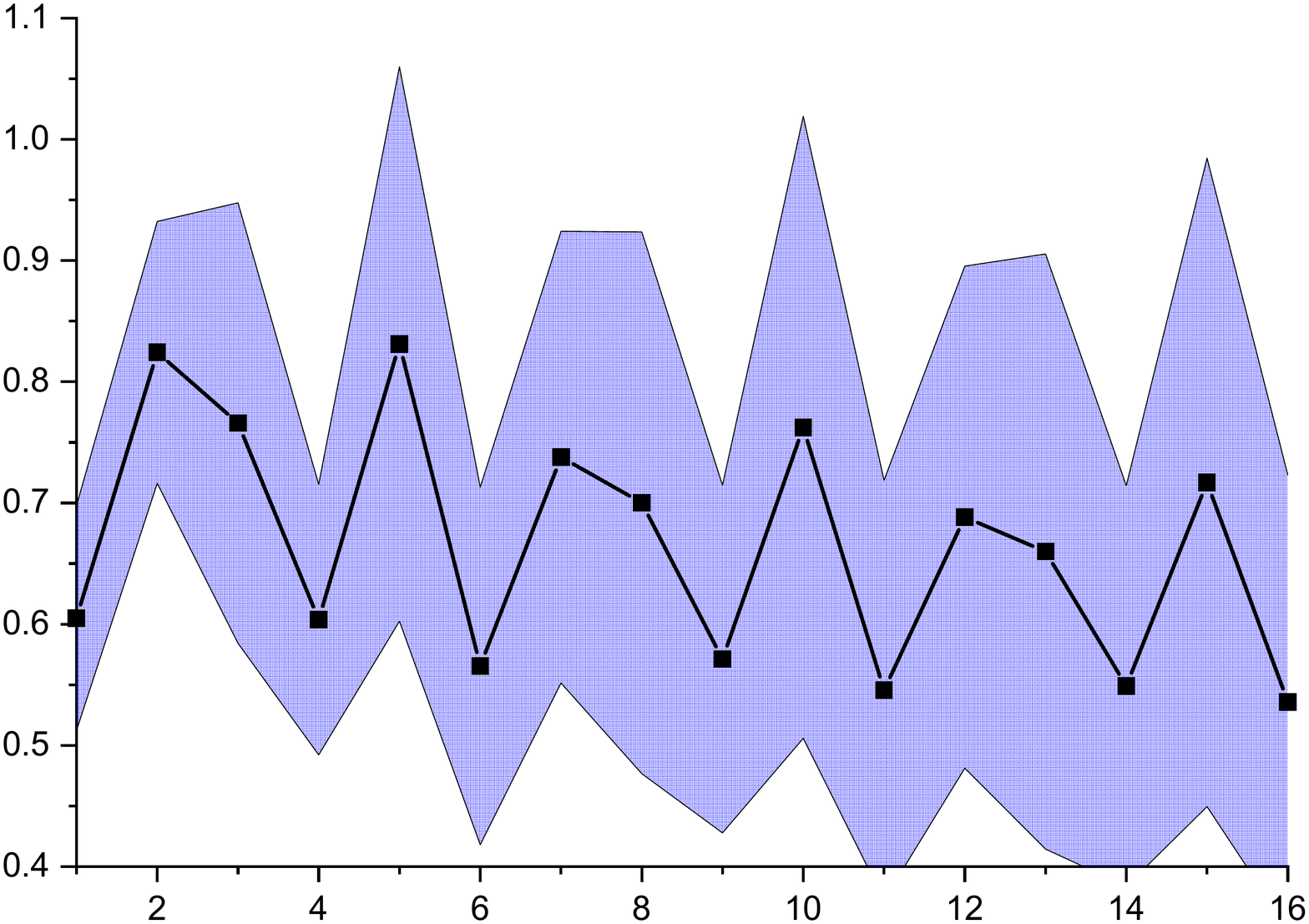}
\end{minipage}
}
\subfigure[  CW-L2]{
\begin{minipage}[b]{0.14\textwidth}
\includegraphics[width=1\linewidth]{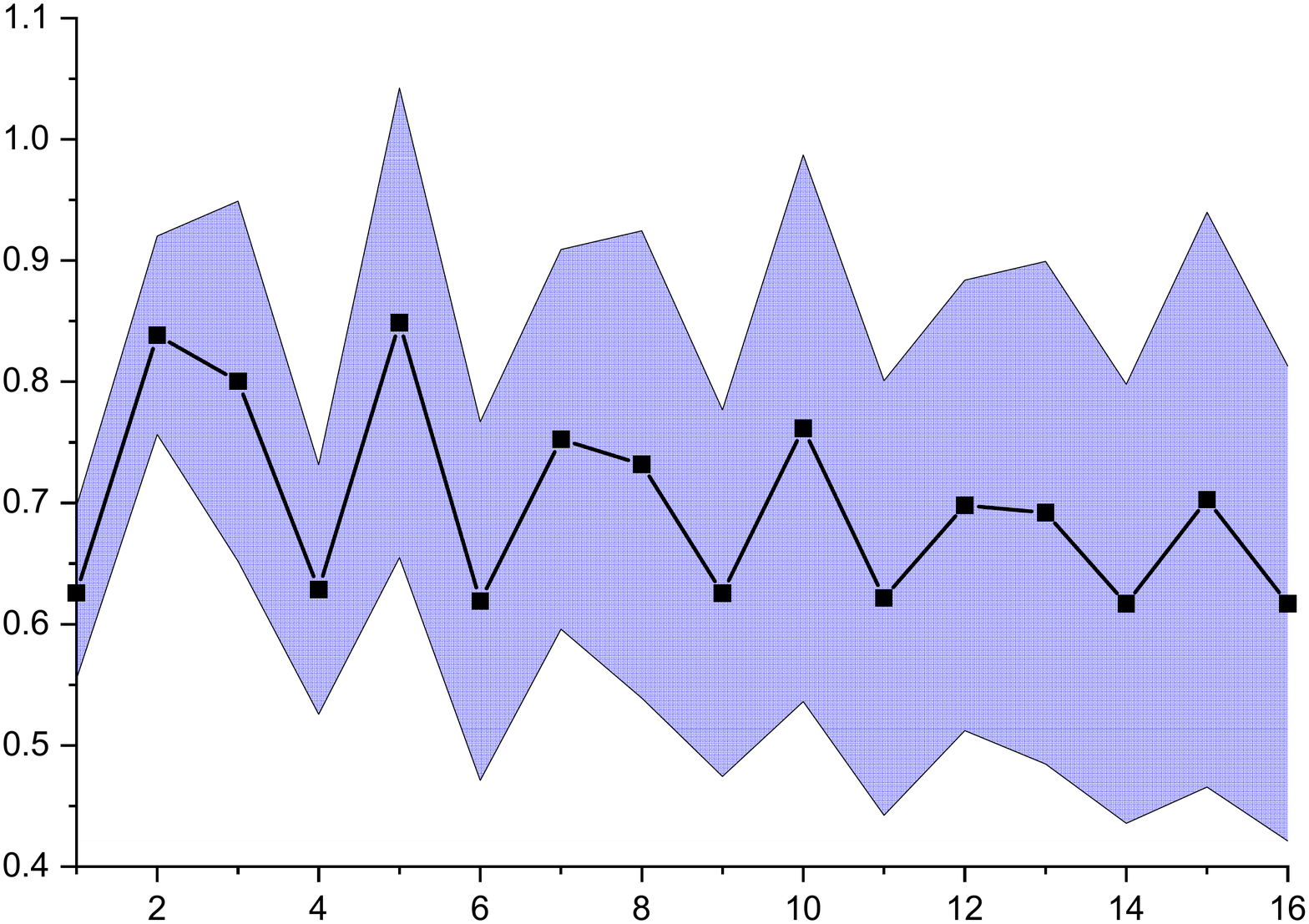}\vspace{4pt}
\includegraphics[width=1\linewidth]{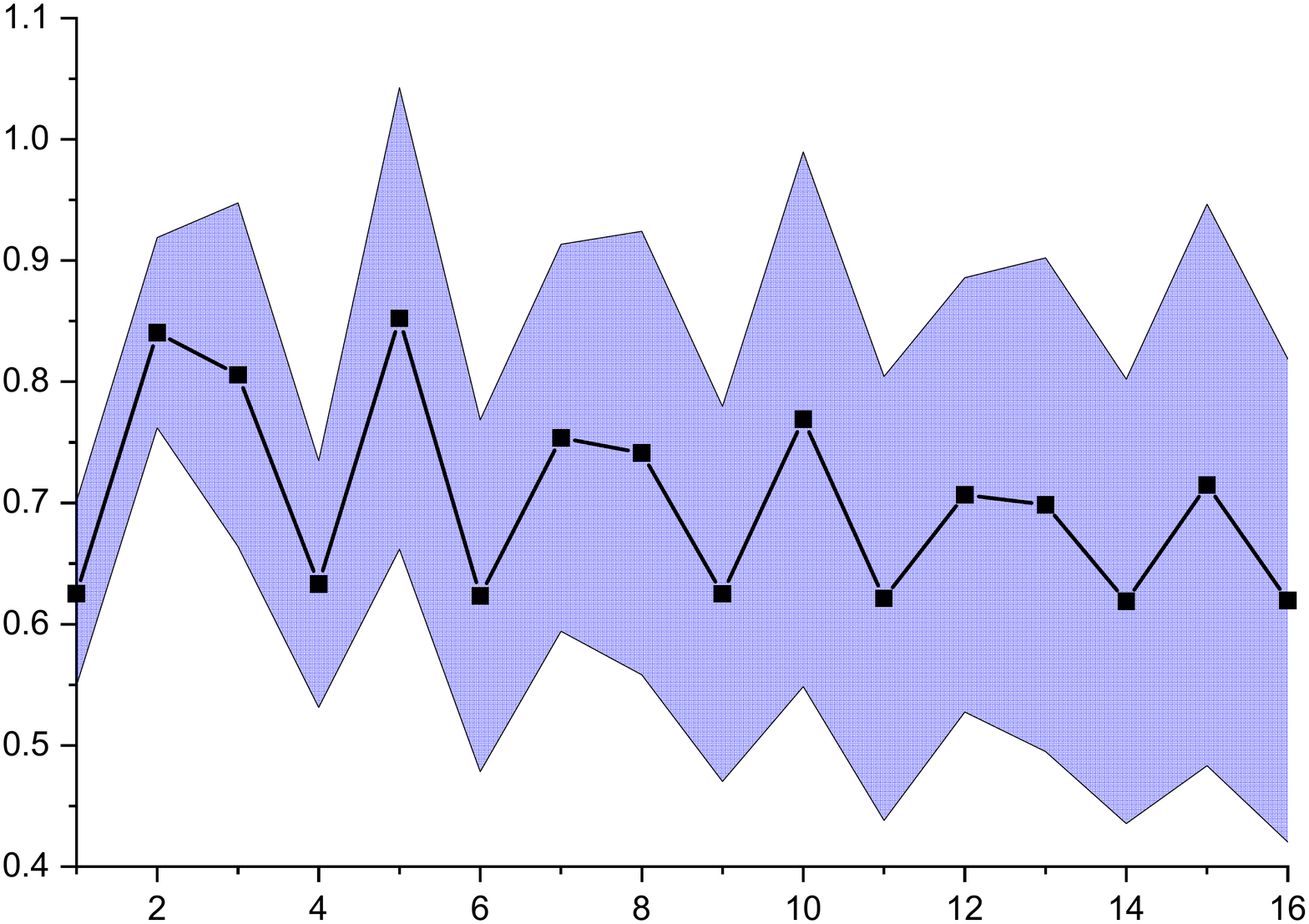}
\end{minipage}
}
\subfigure[  DeepFool]{
\begin{minipage}[b]{0.14\textwidth}
\includegraphics[width=1\linewidth]{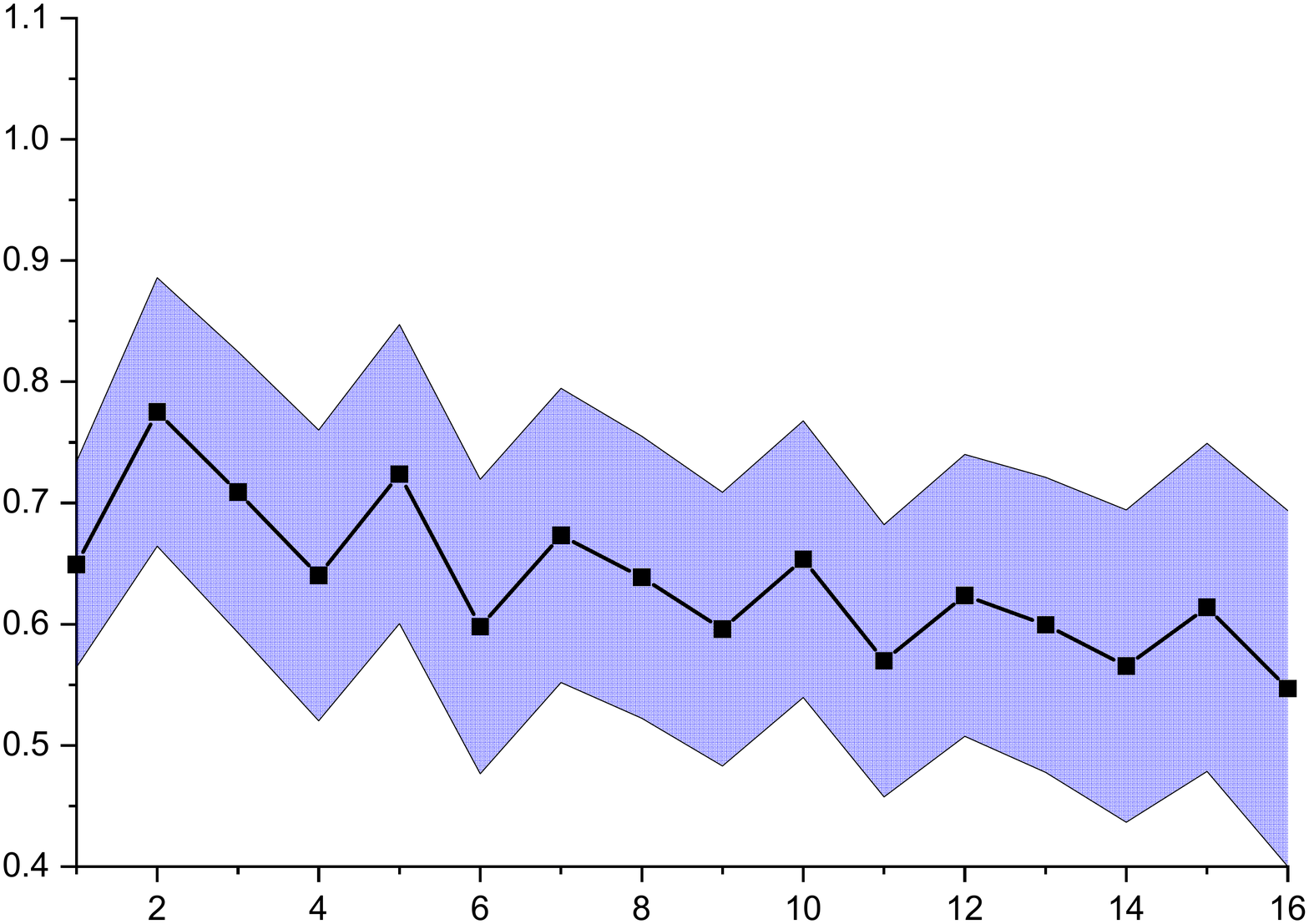}\vspace{4pt}
\includegraphics[width=1\linewidth]{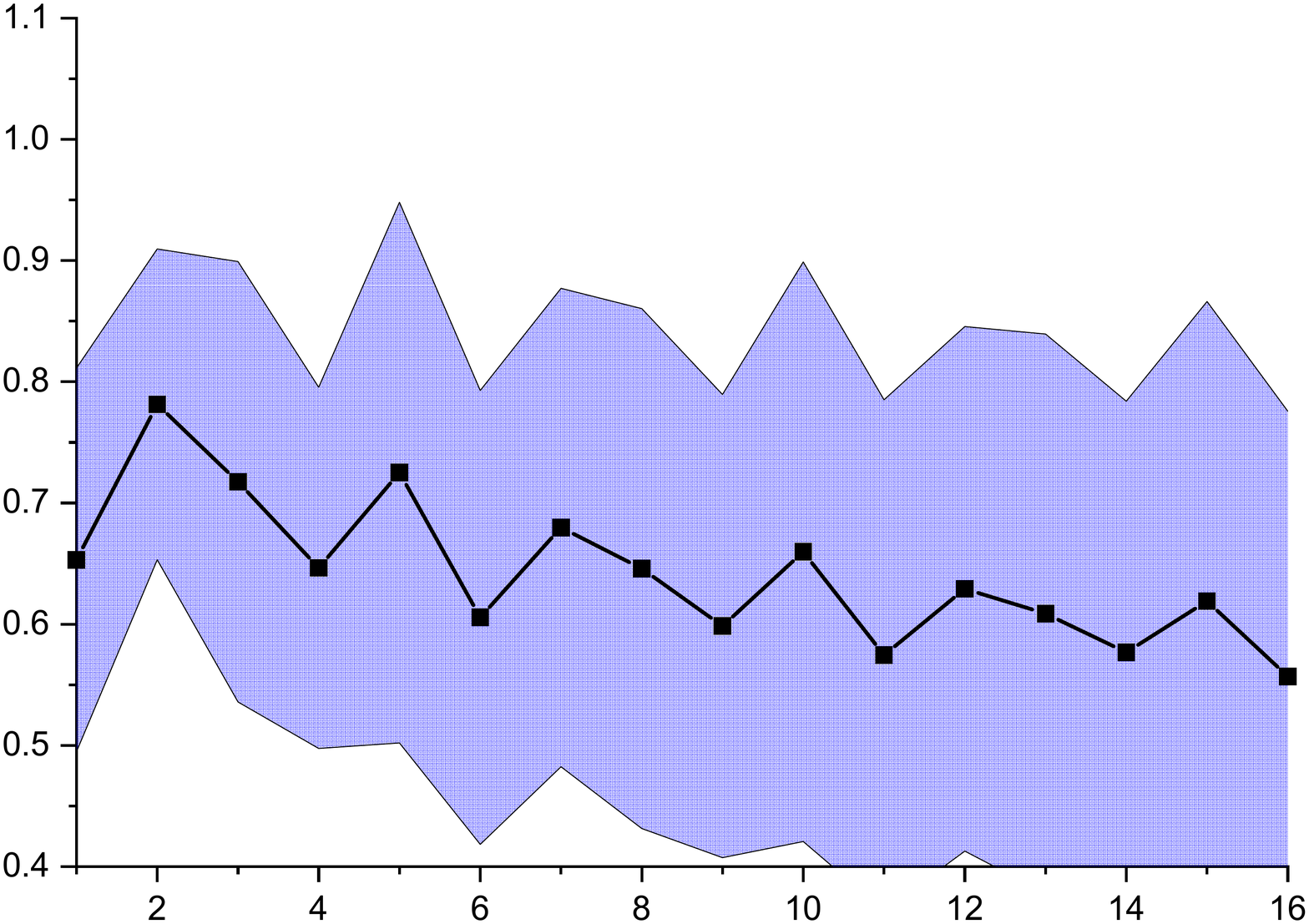}
\end{minipage}
}
\subfigure[  R-PGD]{
\begin{minipage}[b]{0.14\textwidth}
\includegraphics[width=1\linewidth]{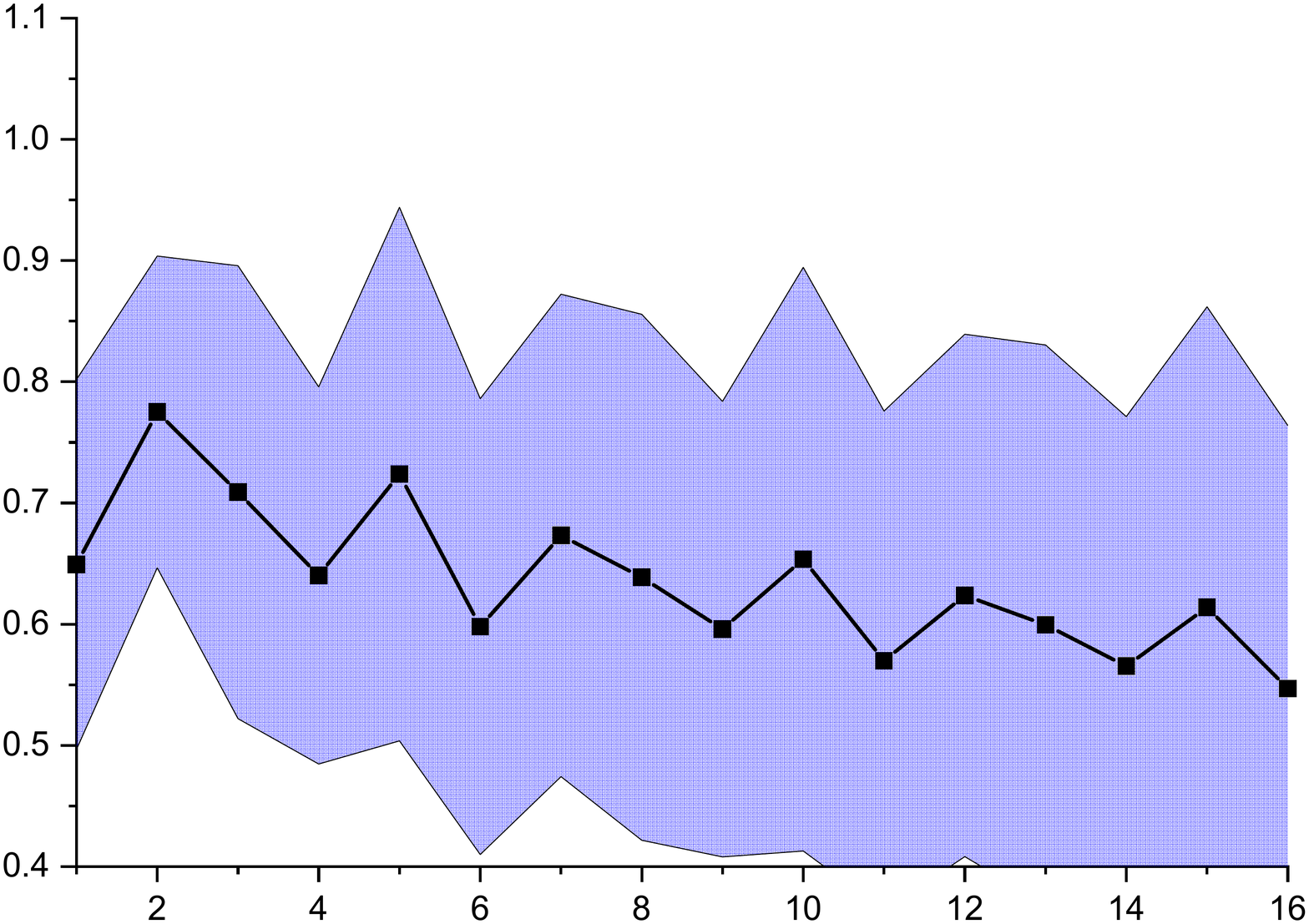}\vspace{4pt}
\includegraphics[width=1\linewidth]{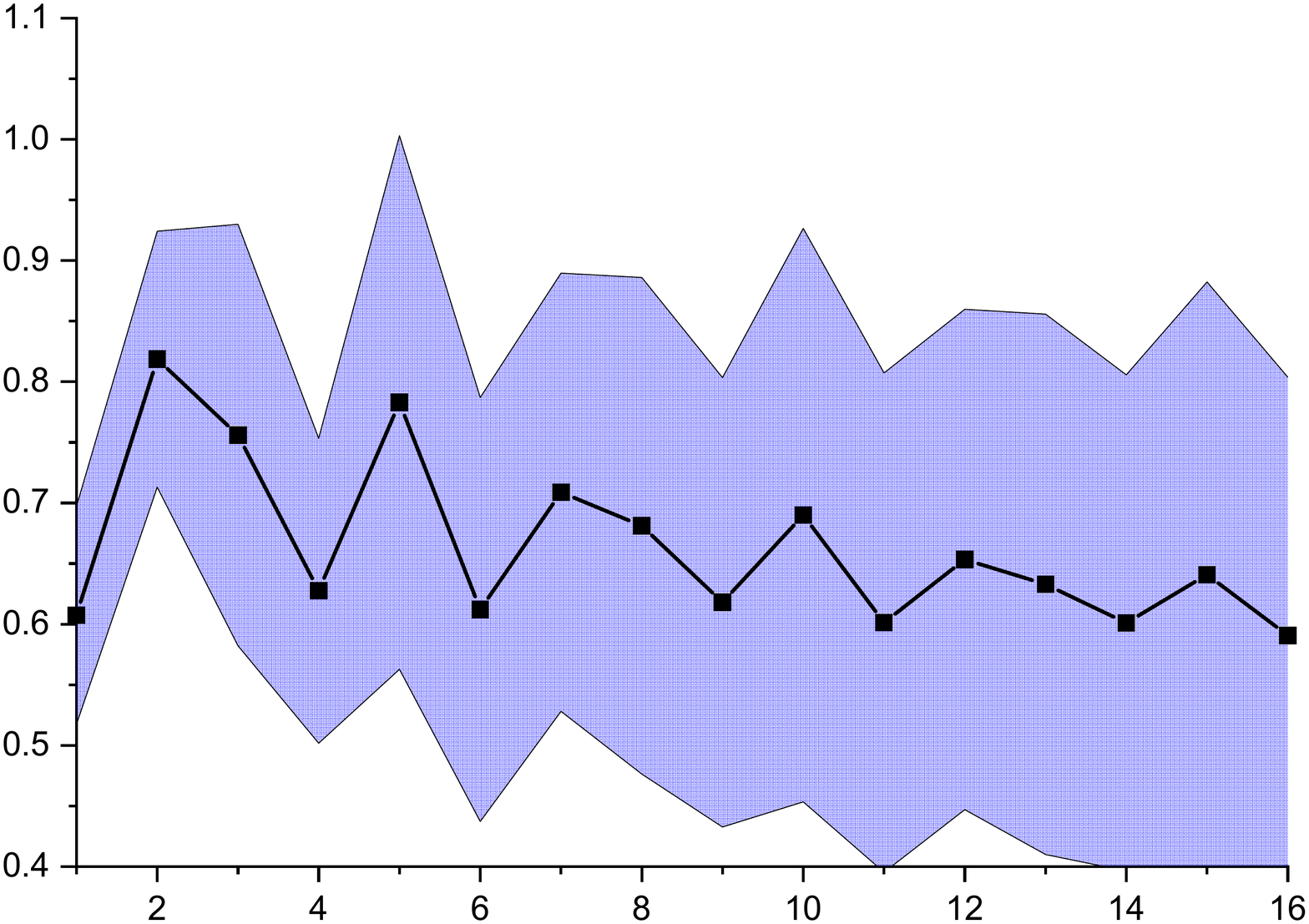}
\end{minipage}
}
}
\vspace{-3mm}
  \caption{ Statistics (mean $\pm$ standard deviation) of MBF coefficients on train (top row) and test (bottom row) set of CIFAR-10. 
  }
  \label{fig:cifar MBF}
  \vspace{-4mm}
\end{figure}

\begin{figure}[htbp]
\centering
\scalebox{0.5}
{
\subfigure[ clean]{
\begin{minipage}[b]{0.14\textwidth}
\includegraphics[width=1\linewidth]{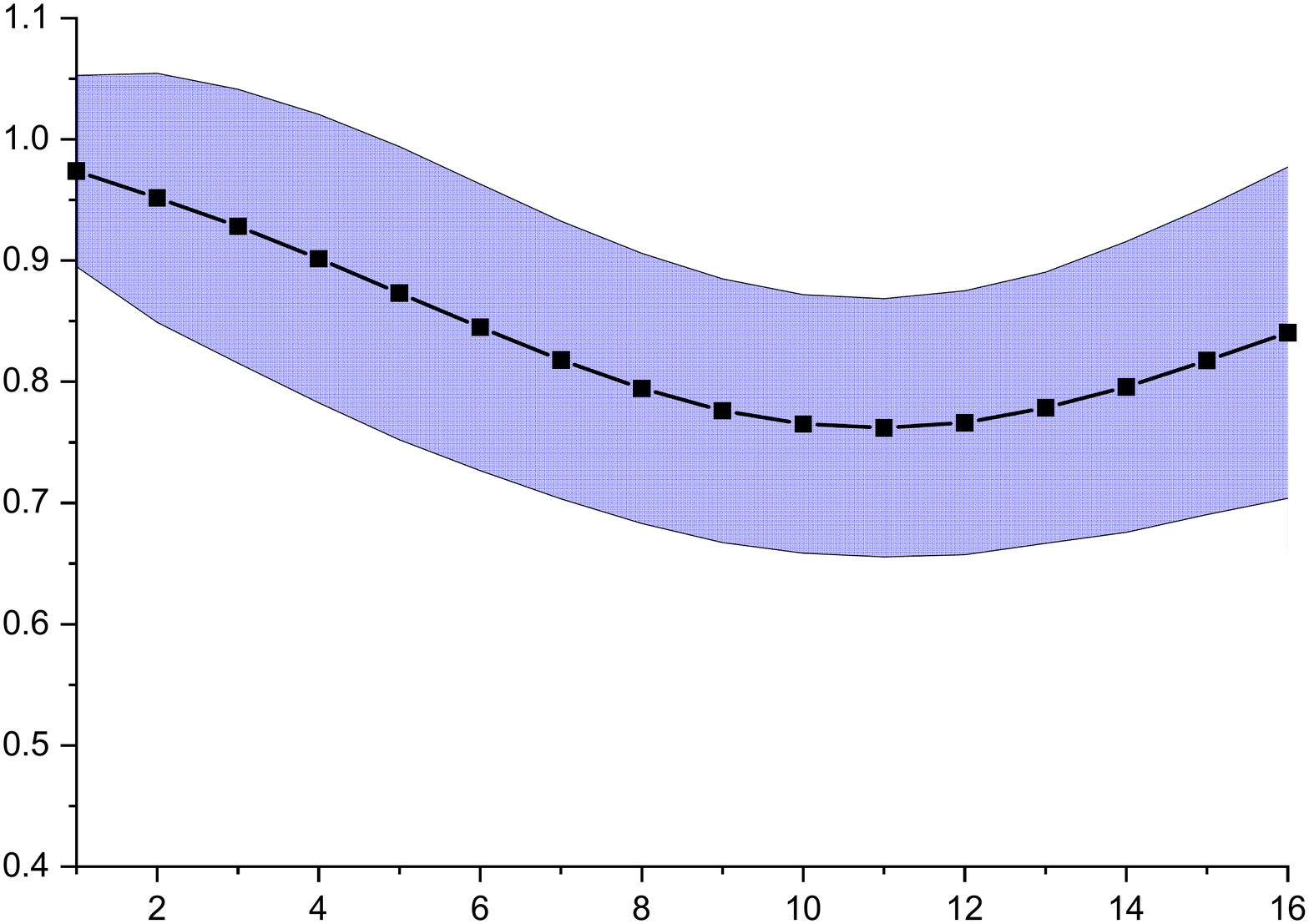}\vspace{4pt}
\includegraphics[width=1\linewidth]{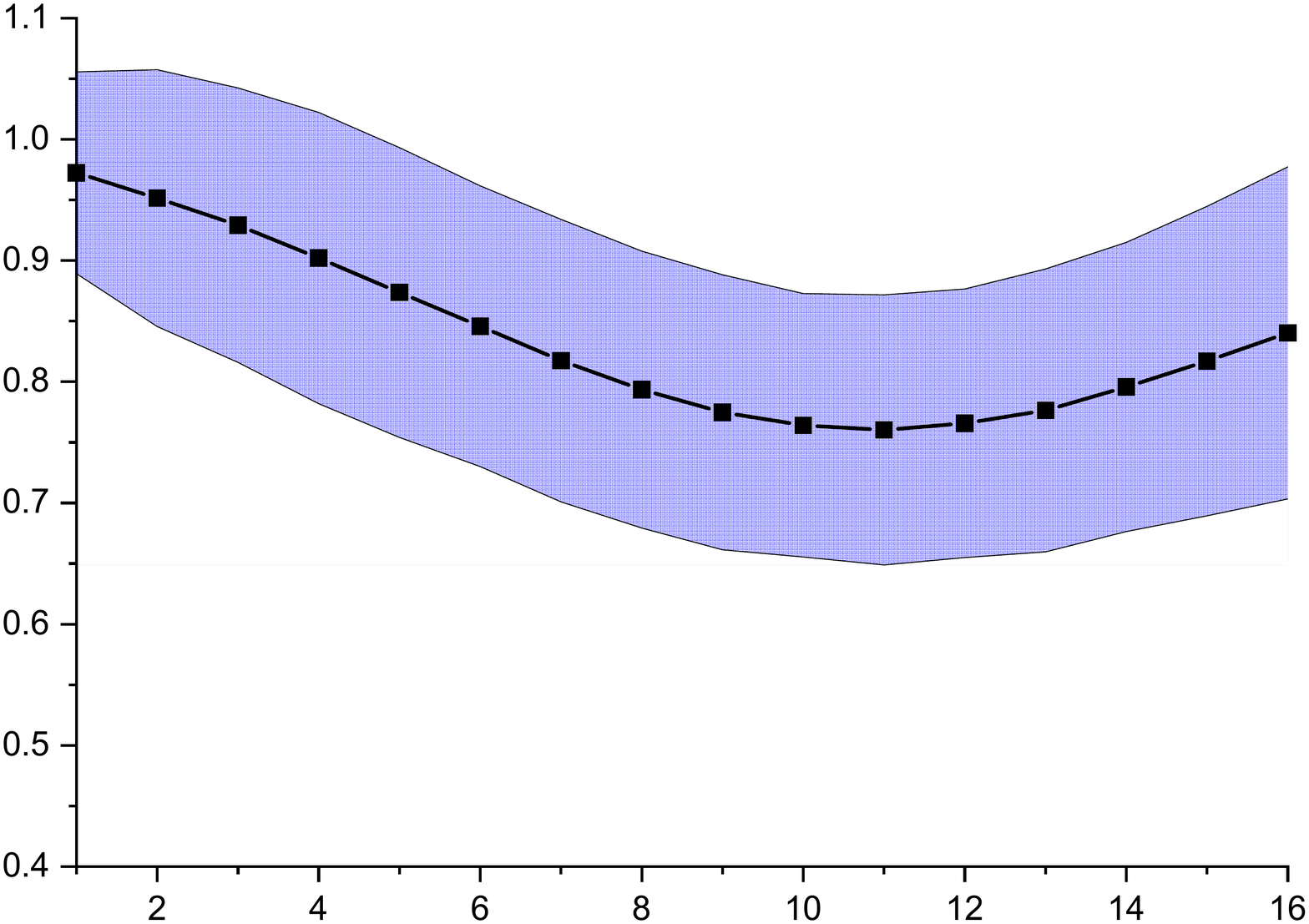}
\end{minipage}
}
\subfigure[ noisy]{
\begin{minipage}[b]{0.14\textwidth}
\includegraphics[width=1\linewidth]{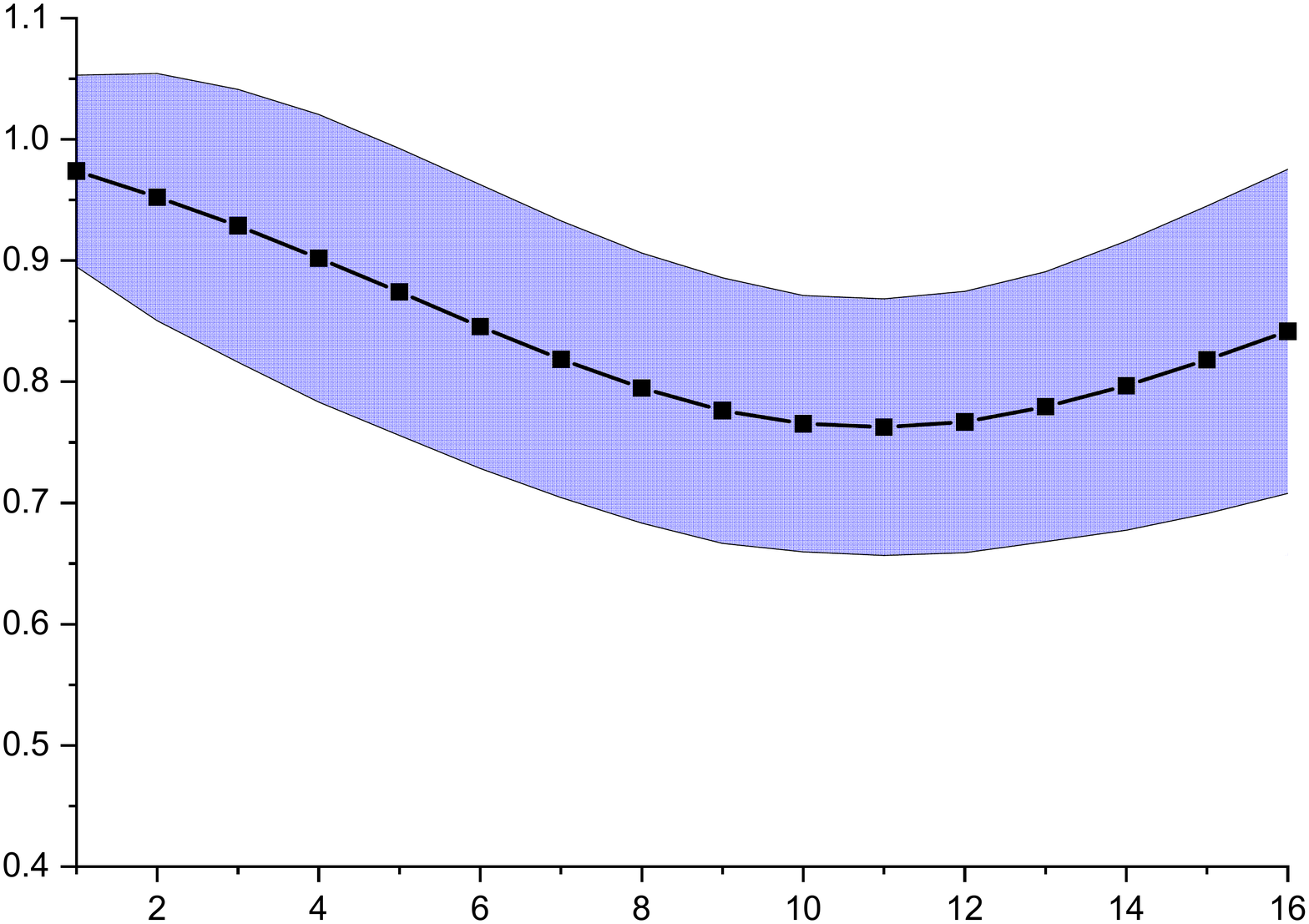}\vspace{4pt}
\includegraphics[width=1\linewidth]{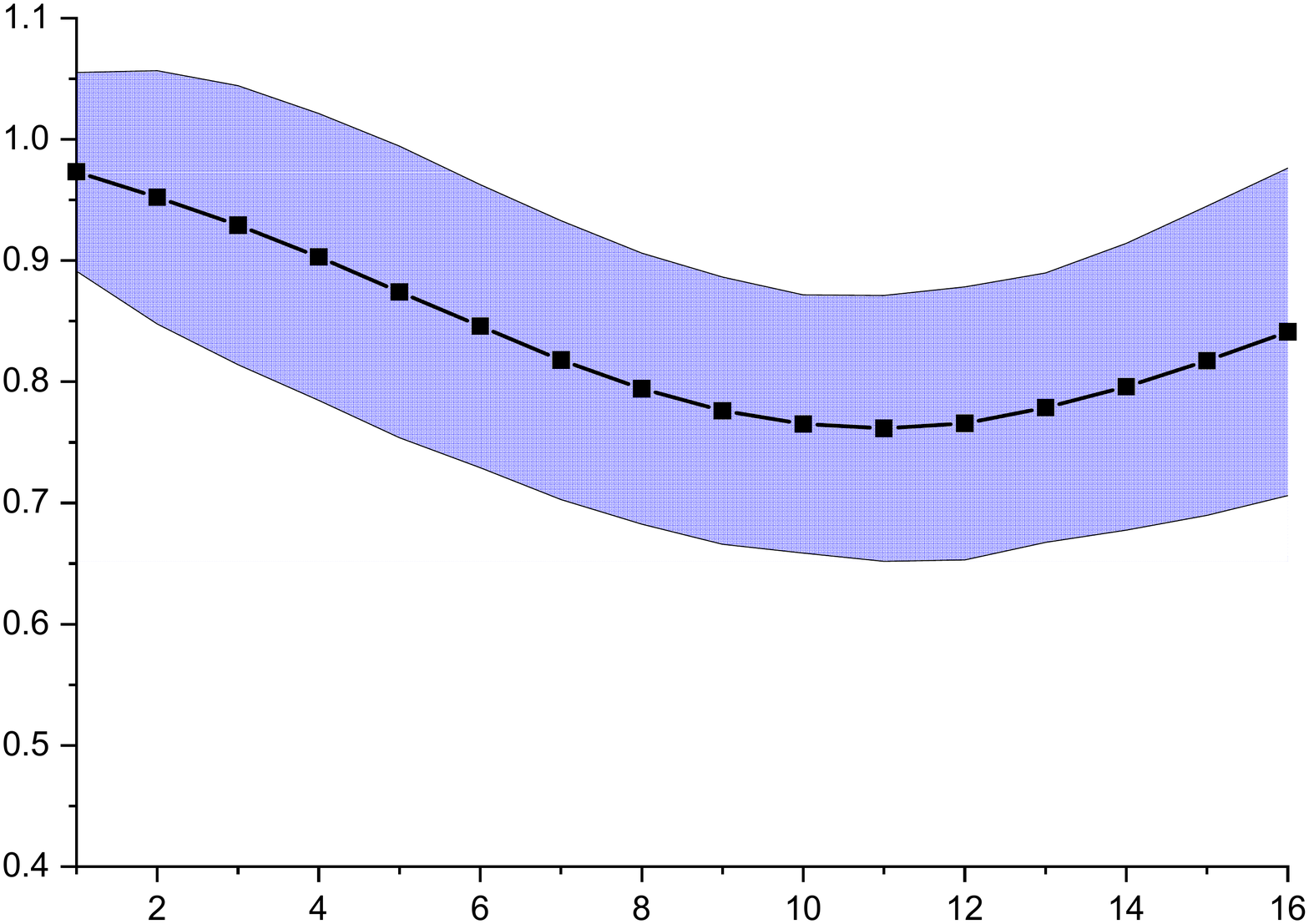}
\end{minipage}
}
\subfigure[  BIM]{
\begin{minipage}[b]{0.14\textwidth}
\includegraphics[width=1\linewidth]{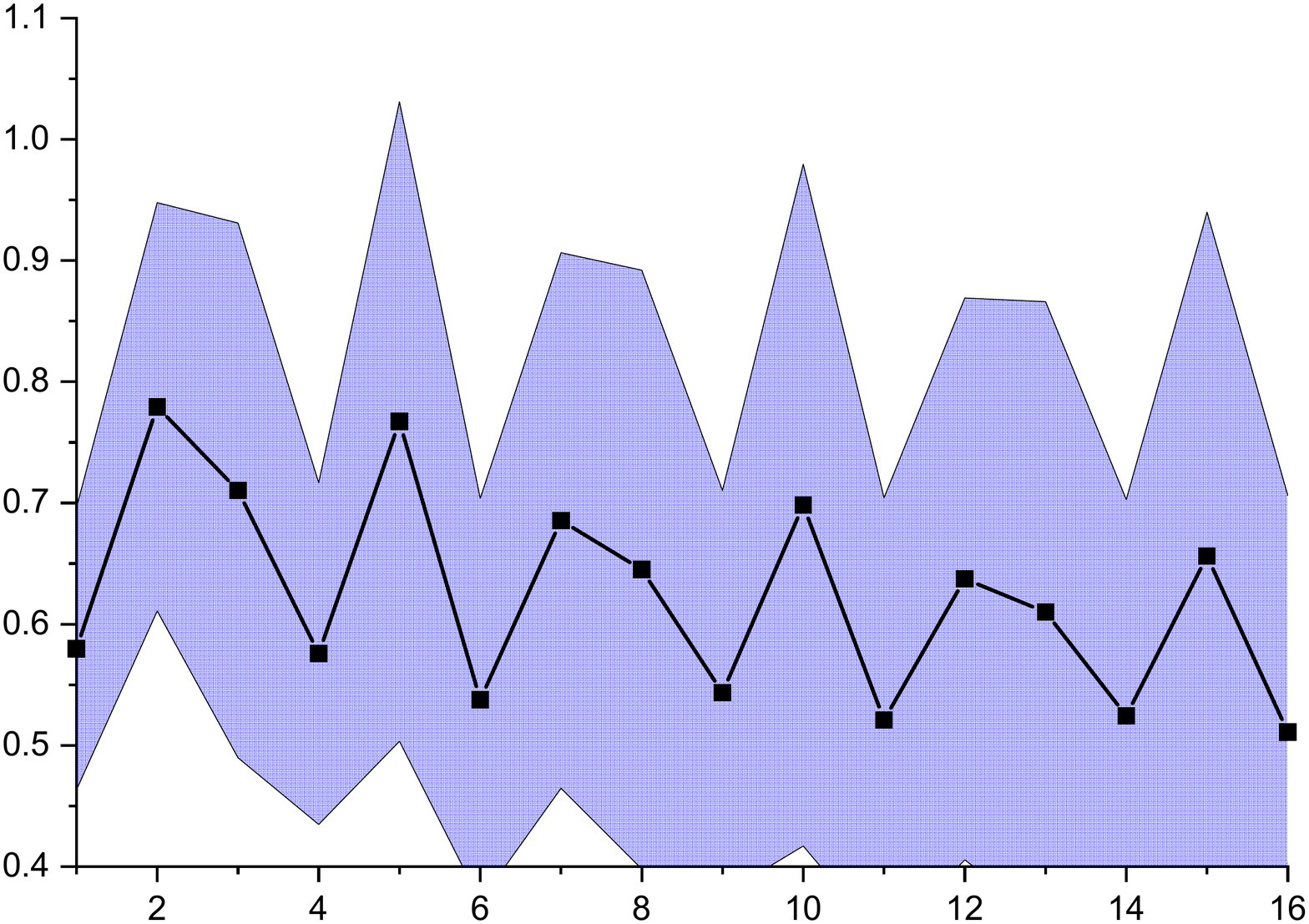}\vspace{4pt}
\includegraphics[width=1\linewidth]{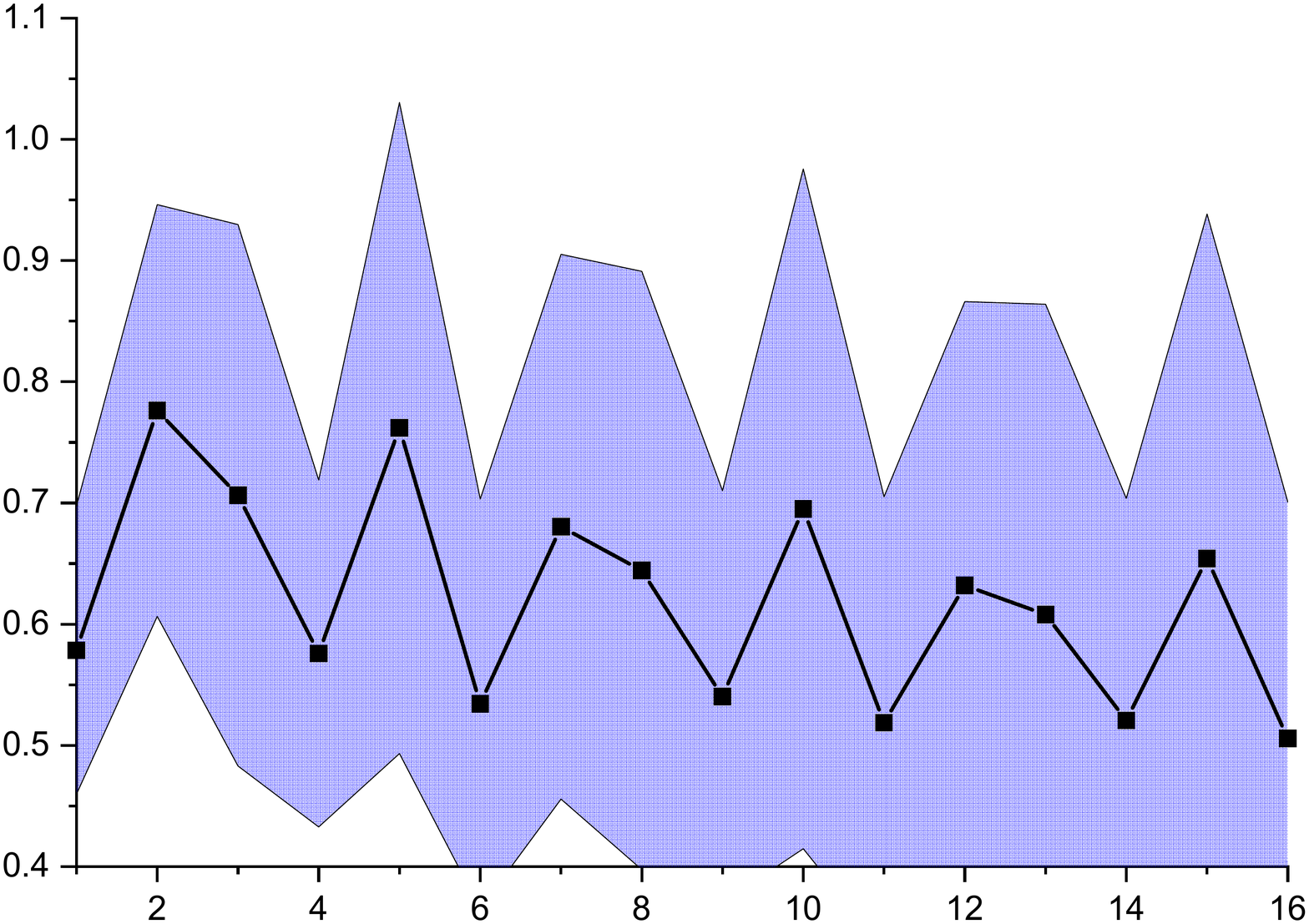}
\end{minipage}
}
\subfigure[  CW-L2]{
\begin{minipage}[b]{0.14\textwidth}
\includegraphics[width=1\linewidth]{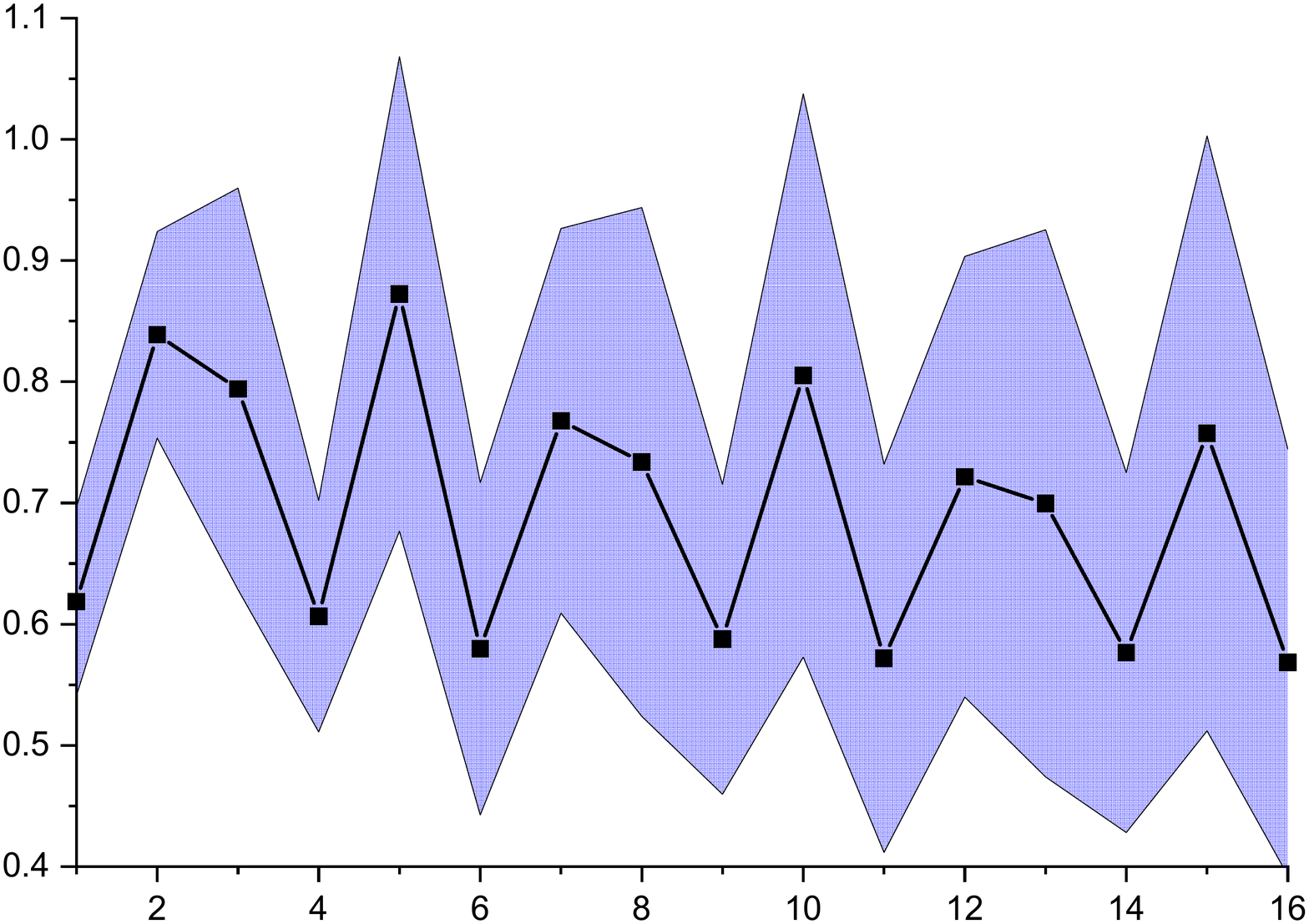}\vspace{4pt}
\includegraphics[width=1\linewidth]{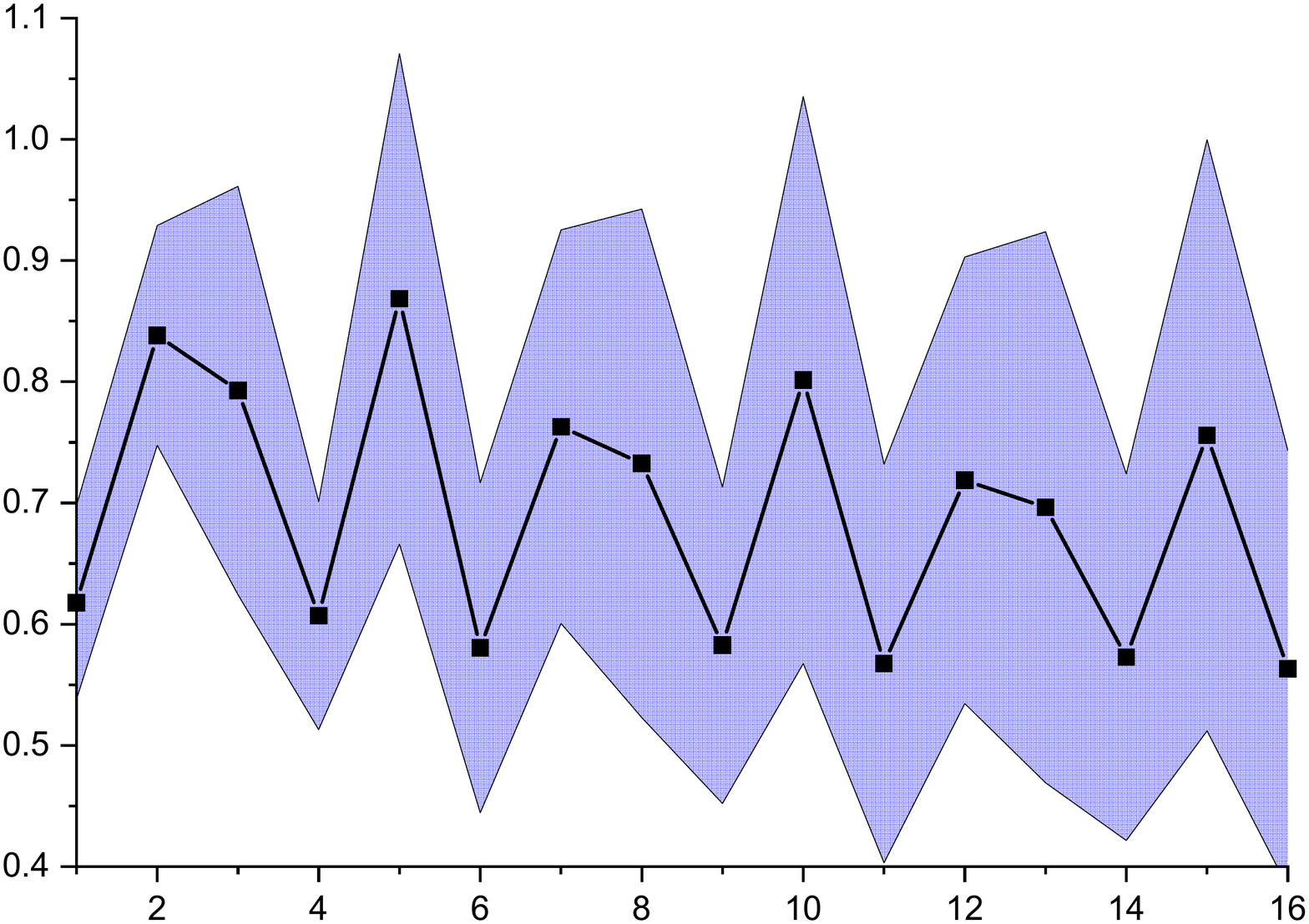}
\end{minipage}
}
\subfigure[  DeepFool]{
\begin{minipage}[b]{0.14\textwidth}
\includegraphics[width=1\linewidth]{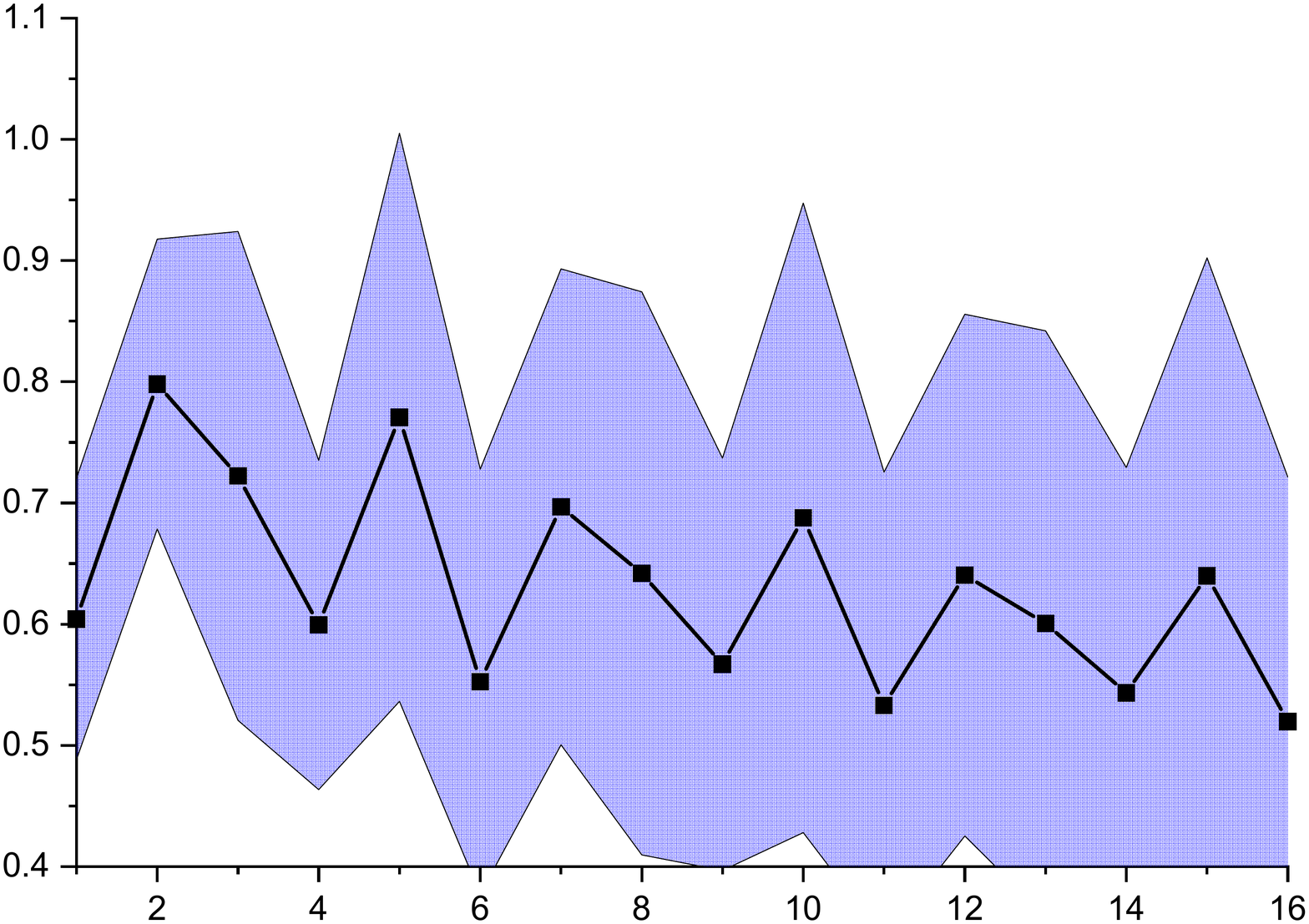}\vspace{4pt}
\includegraphics[width=1\linewidth]{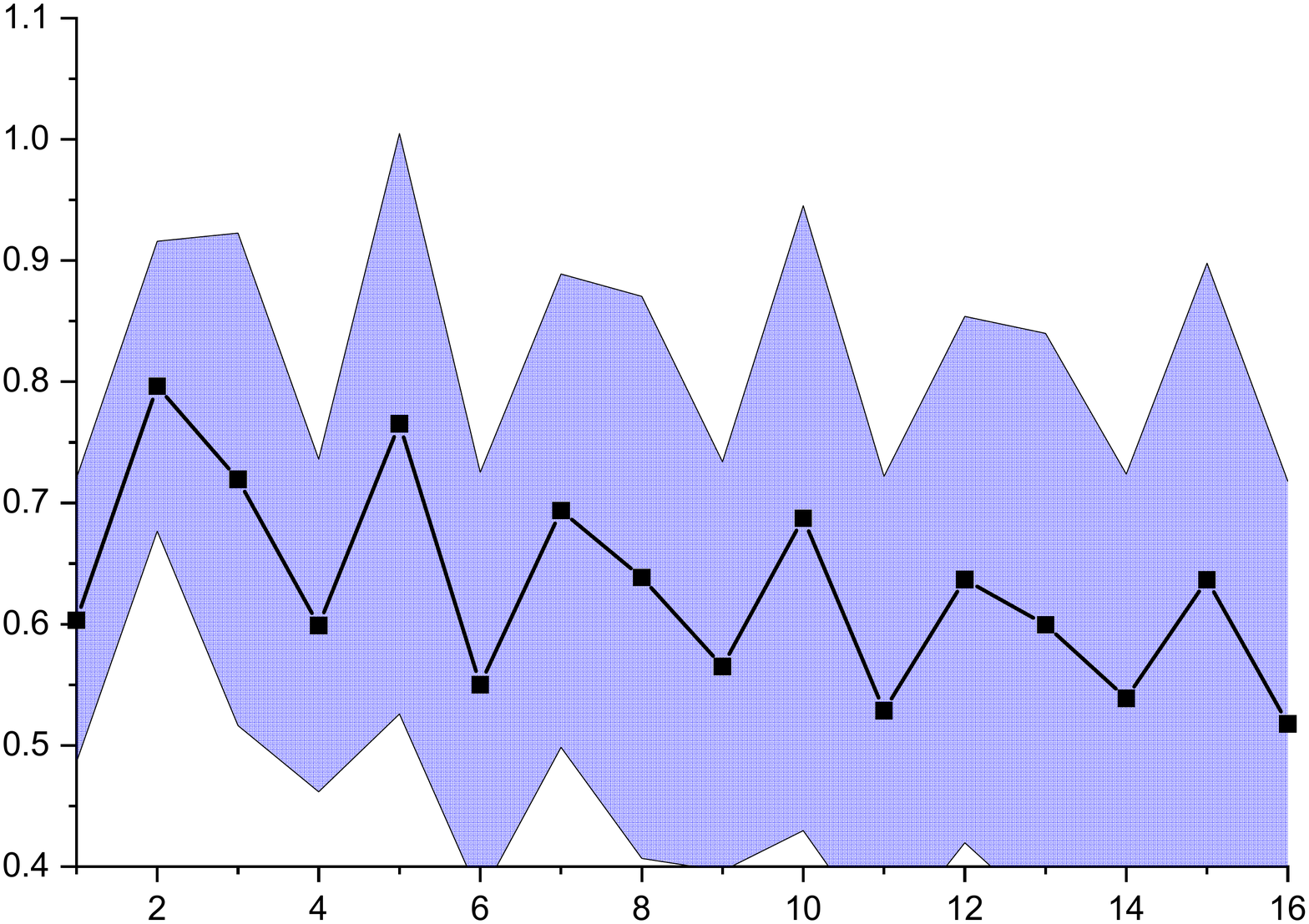}
\end{minipage}
}
\subfigure[  R-PGD]{
\begin{minipage}[b]{0.14\textwidth}
\includegraphics[width=1\linewidth]{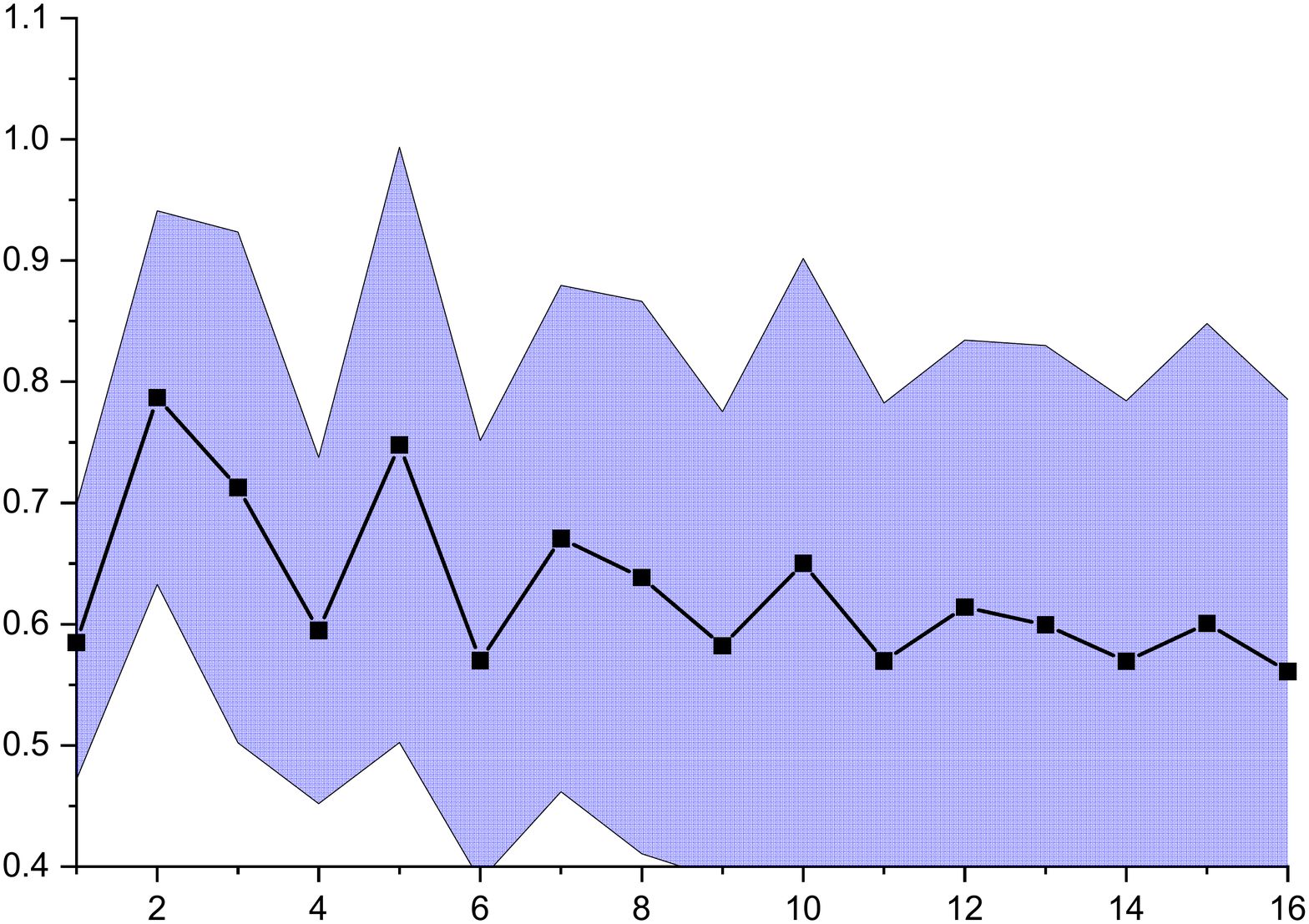}\vspace{4pt}
\includegraphics[width=1\linewidth]{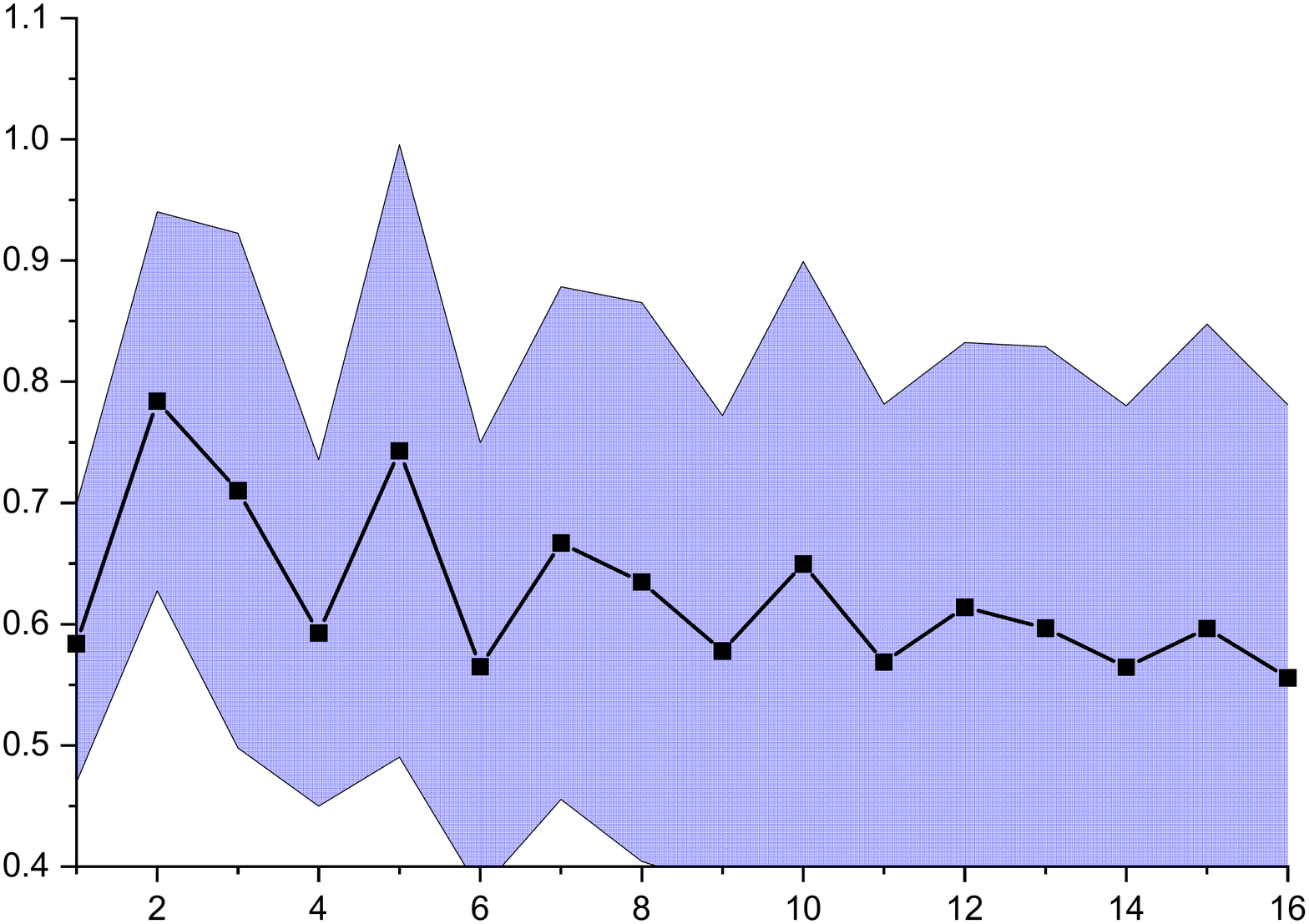}
\end{minipage}
}
}
\vspace{-3mm}
  \caption{ Statistics (mean $\pm$ standard deviation) of MBF coefficients on train (top row) and test (bottom row) set of SVHN. 
  }
  \label{fig:svhn MBF}
  \vspace{-4mm}
\end{figure}

\begin{figure}[htbp]
\centering
\scalebox{0.5}
{
\subfigure[ clean]{
\begin{minipage}[b]{0.14\textwidth}
\includegraphics[width=1\linewidth]{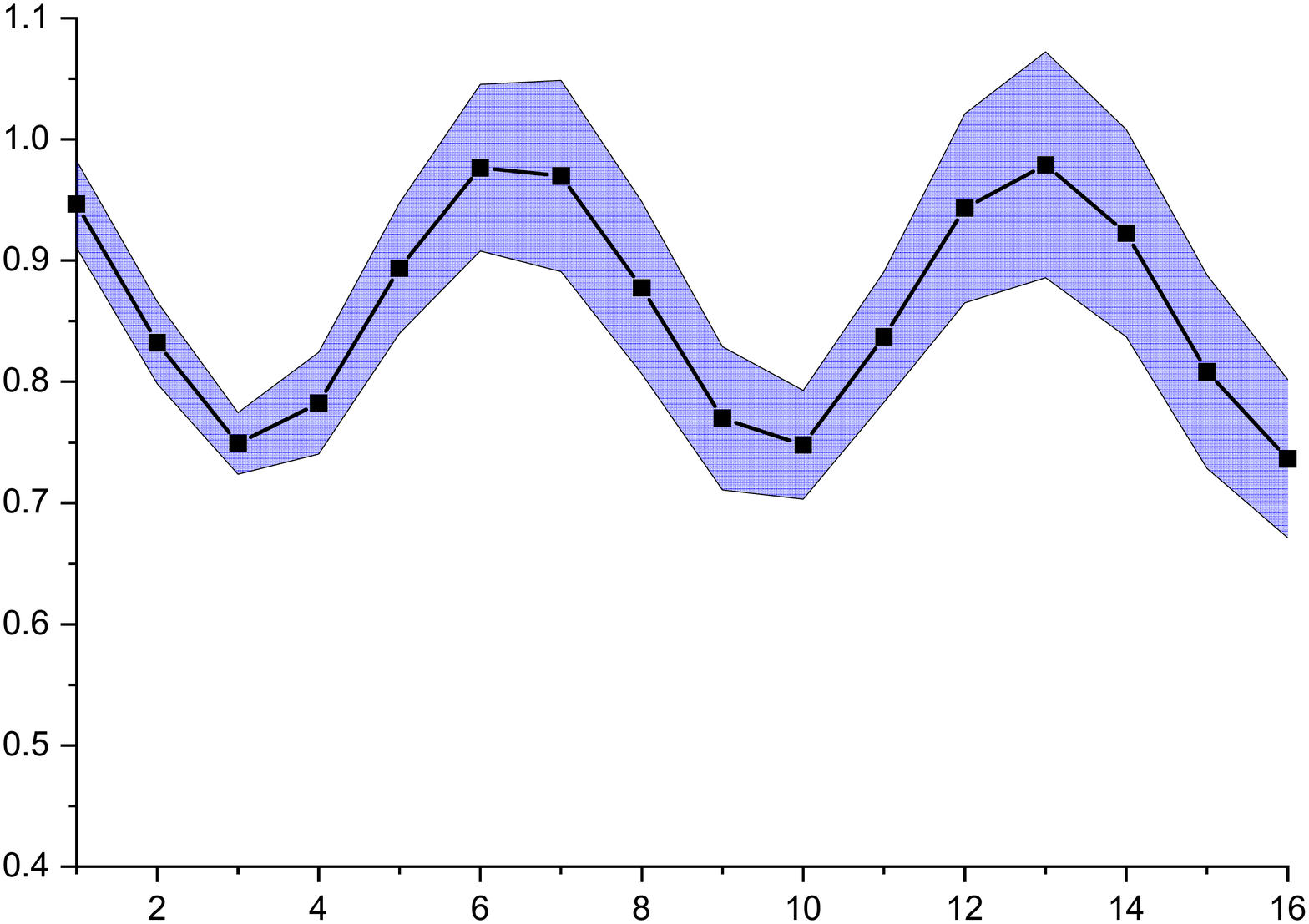}\vspace{4pt}
\includegraphics[width=1\linewidth]{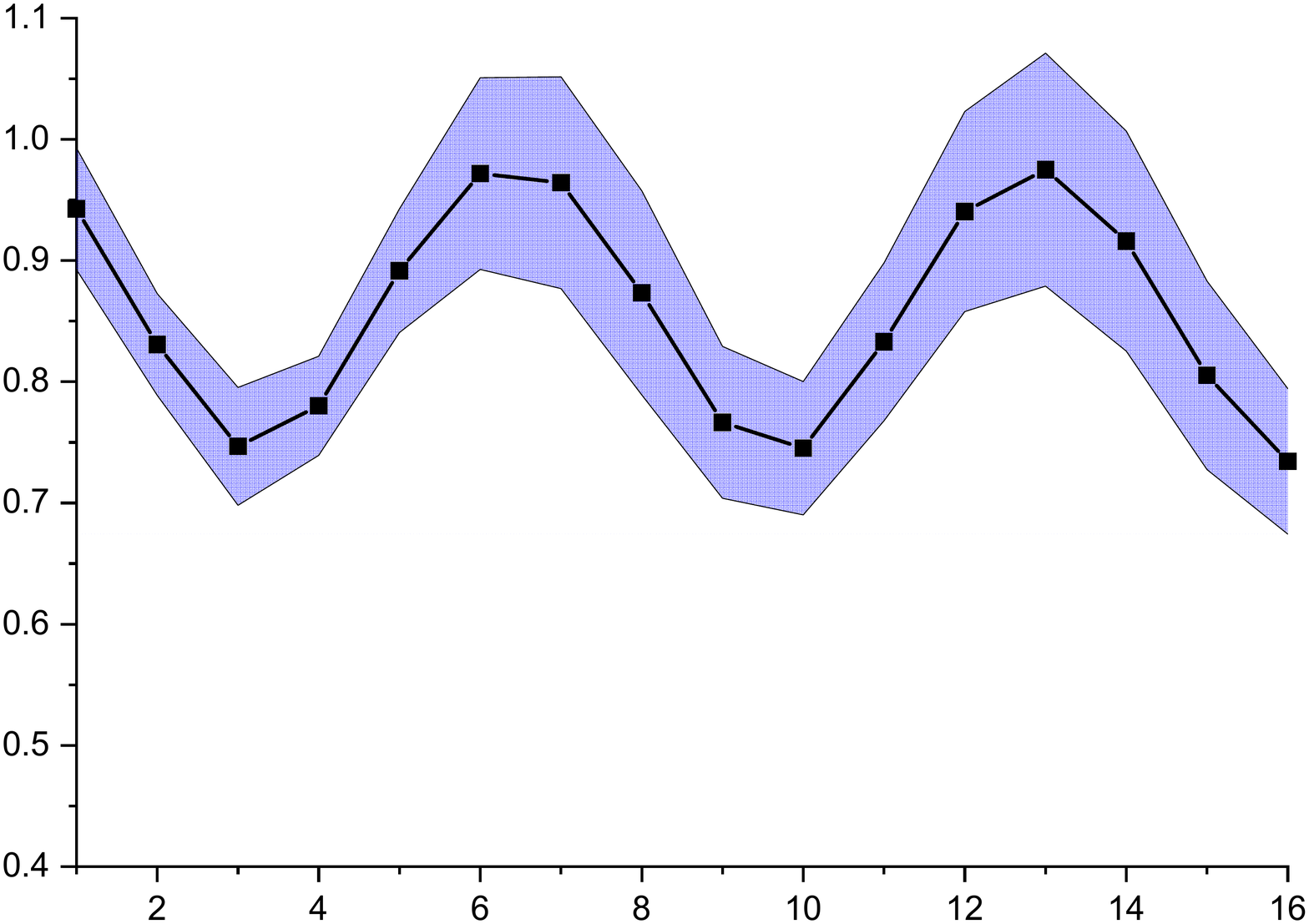}\vspace{4pt}
\includegraphics[width=1\linewidth]{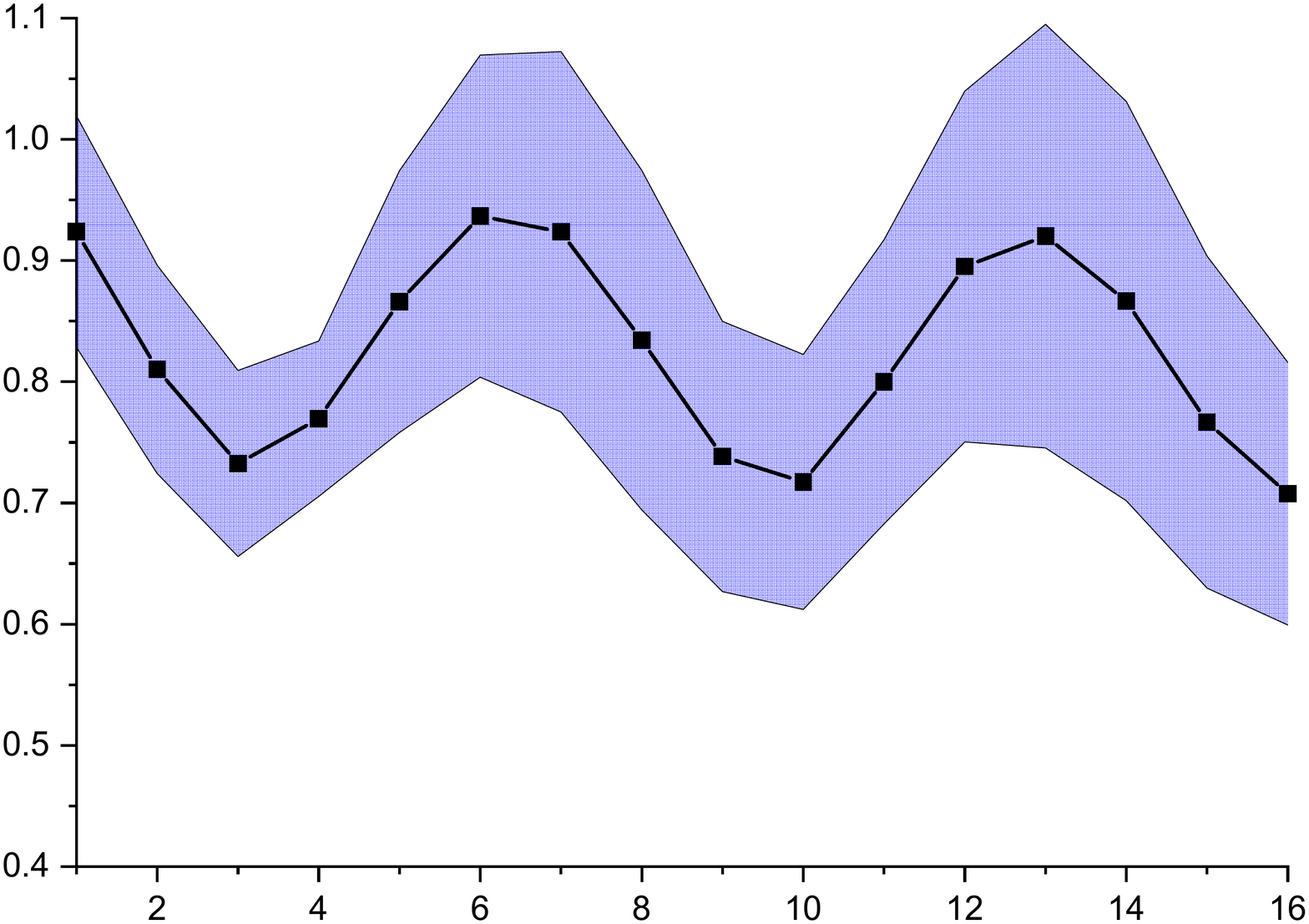}
\end{minipage}
}
\subfigure[ noisy]{
\begin{minipage}[b]{0.14\textwidth}
\includegraphics[width=1\linewidth]{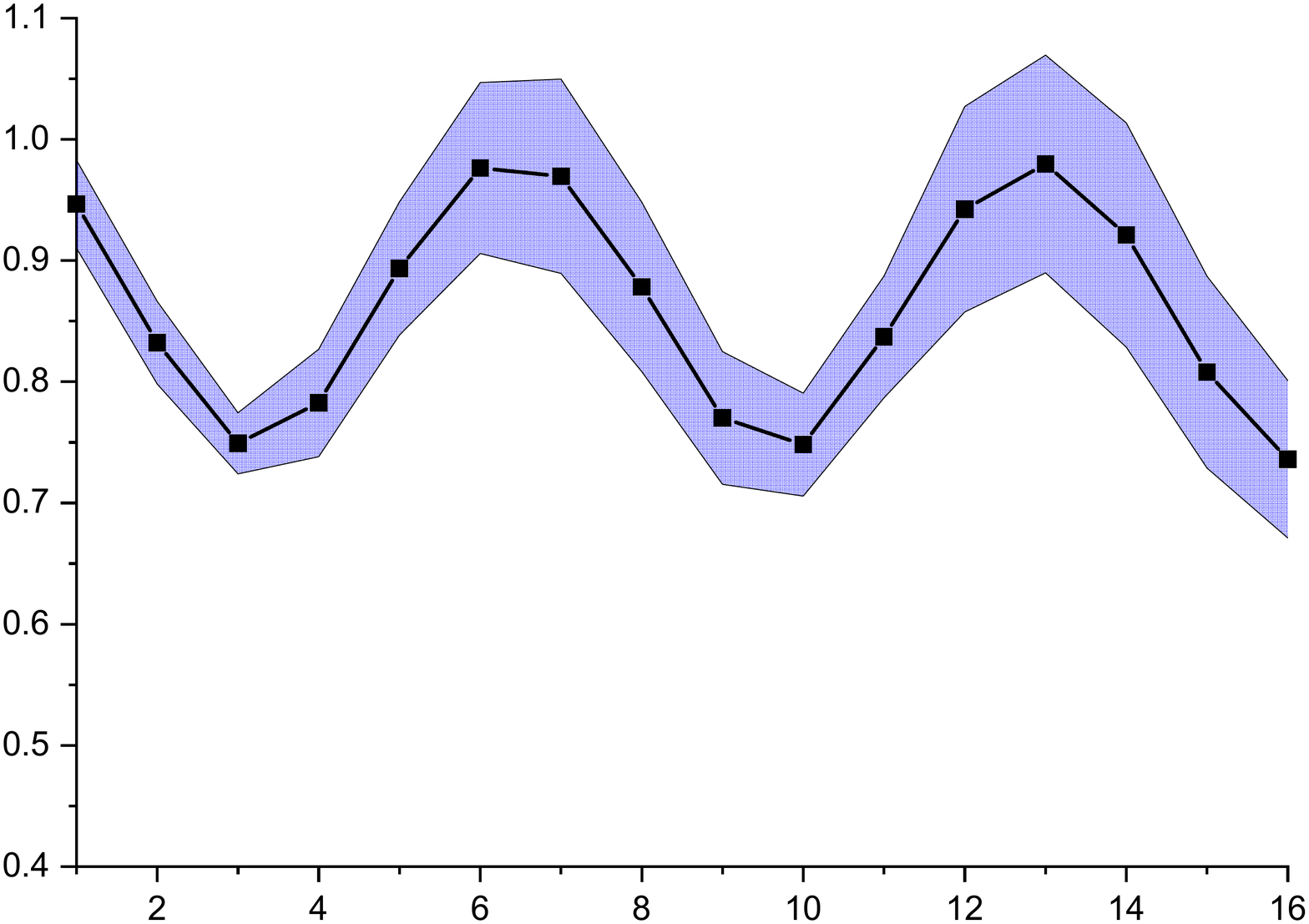}\vspace{4pt}
\includegraphics[width=1\linewidth]{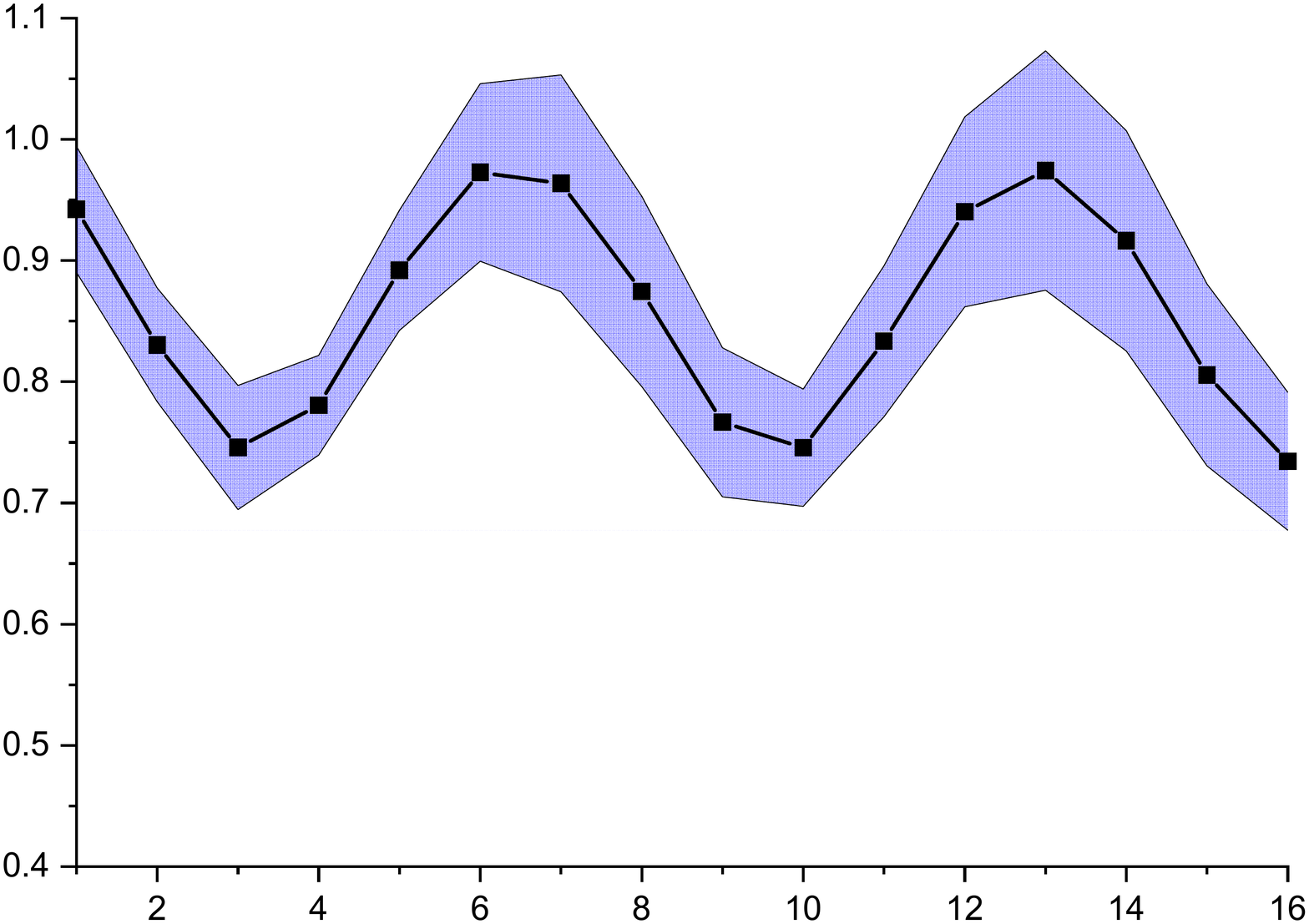}\vspace{4pt}
\includegraphics[width=1\linewidth]{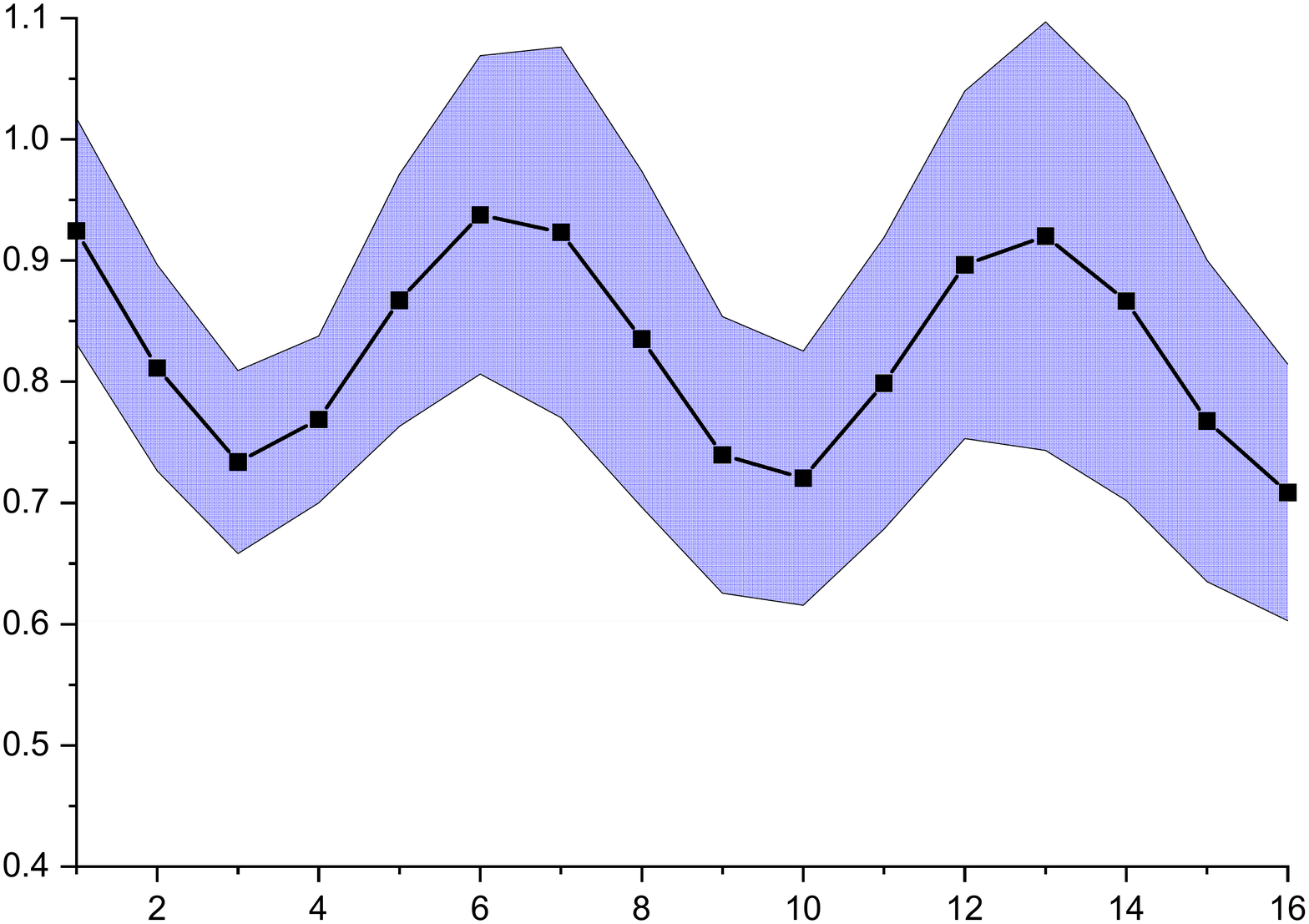}
\end{minipage}
}
\subfigure[  BIM]{
\begin{minipage}[b]{0.14\textwidth}
\includegraphics[width=1\linewidth]{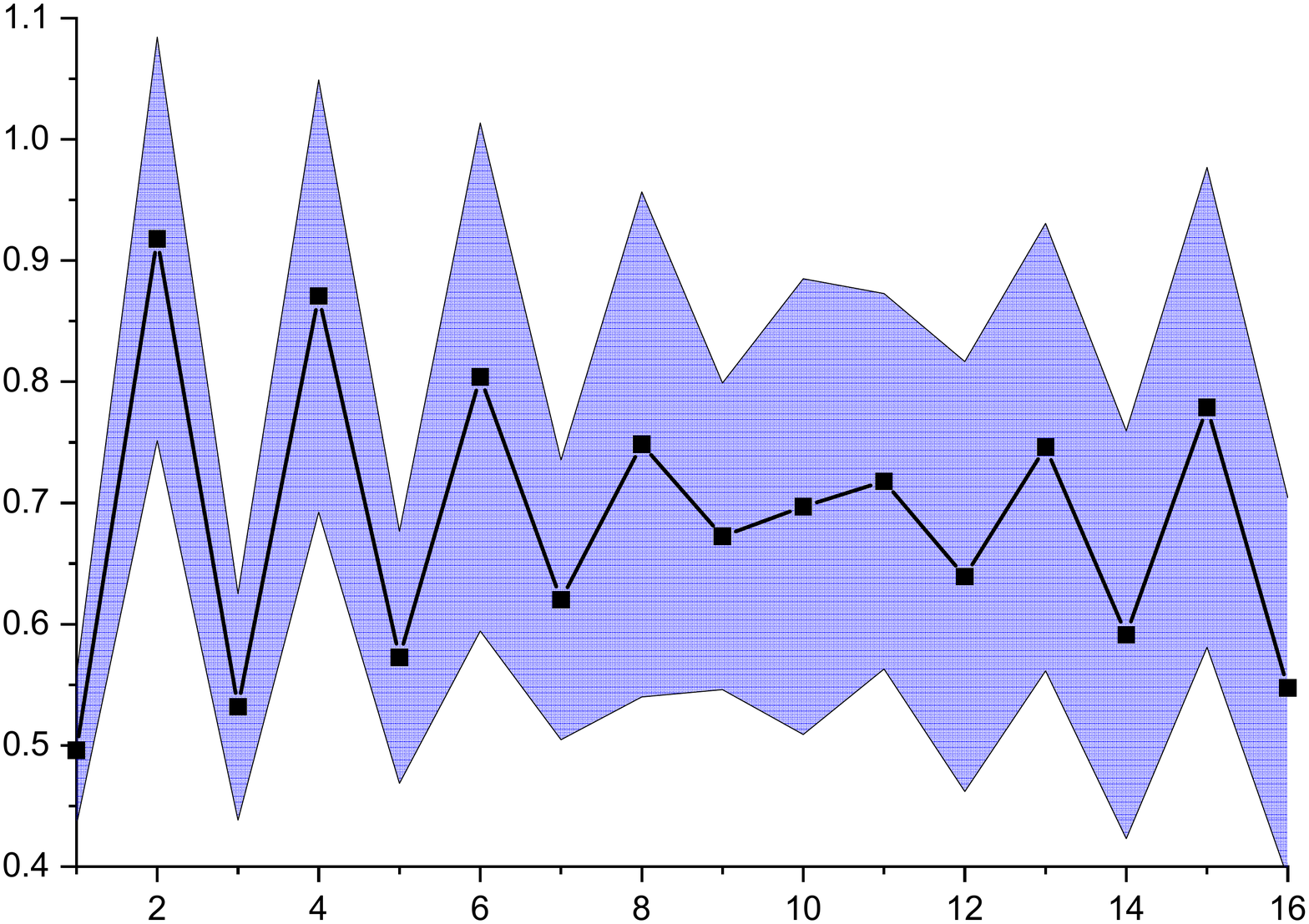}\vspace{4pt}
\includegraphics[width=1\linewidth]{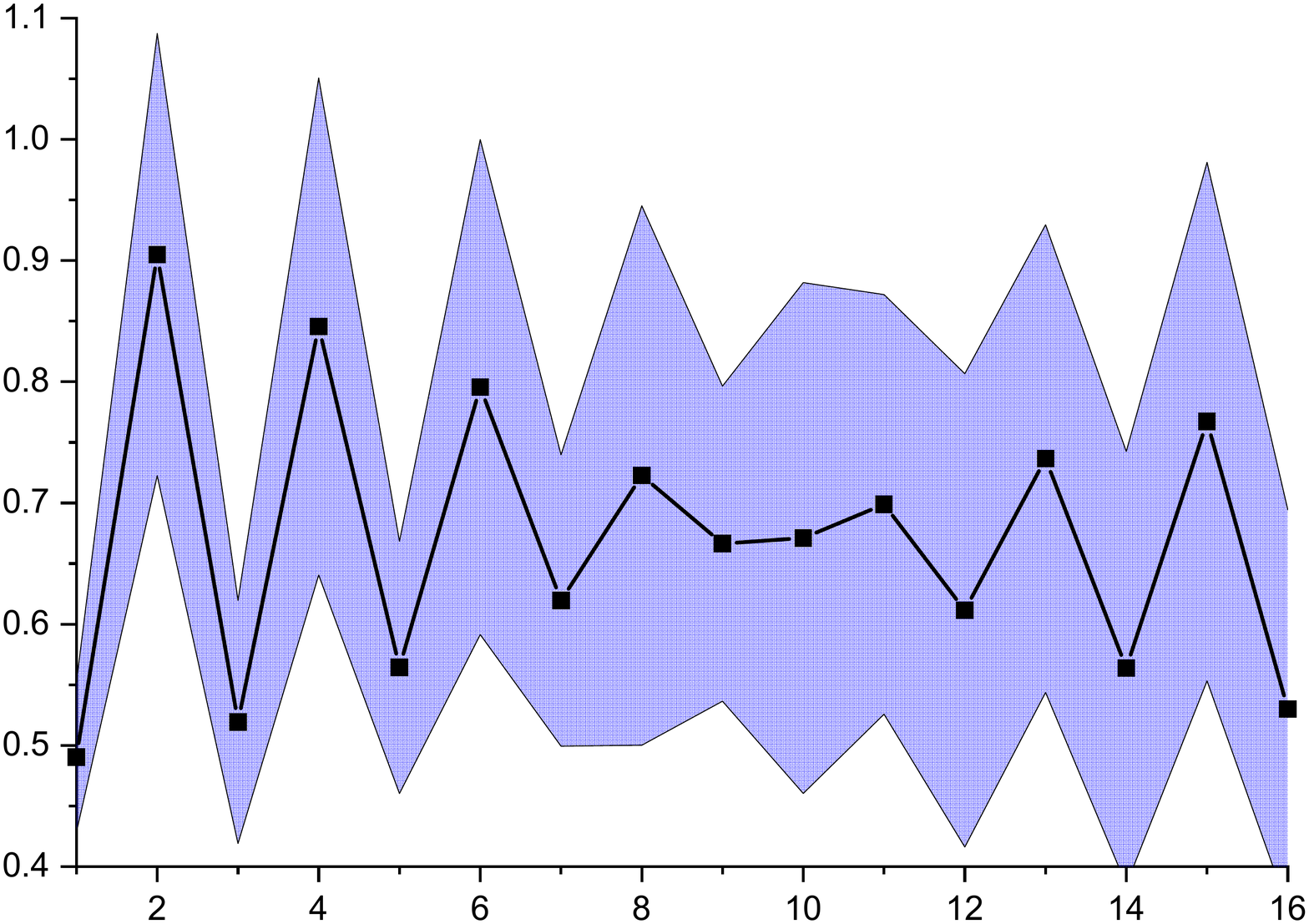}\vspace{4pt}
\includegraphics[width=1\linewidth]{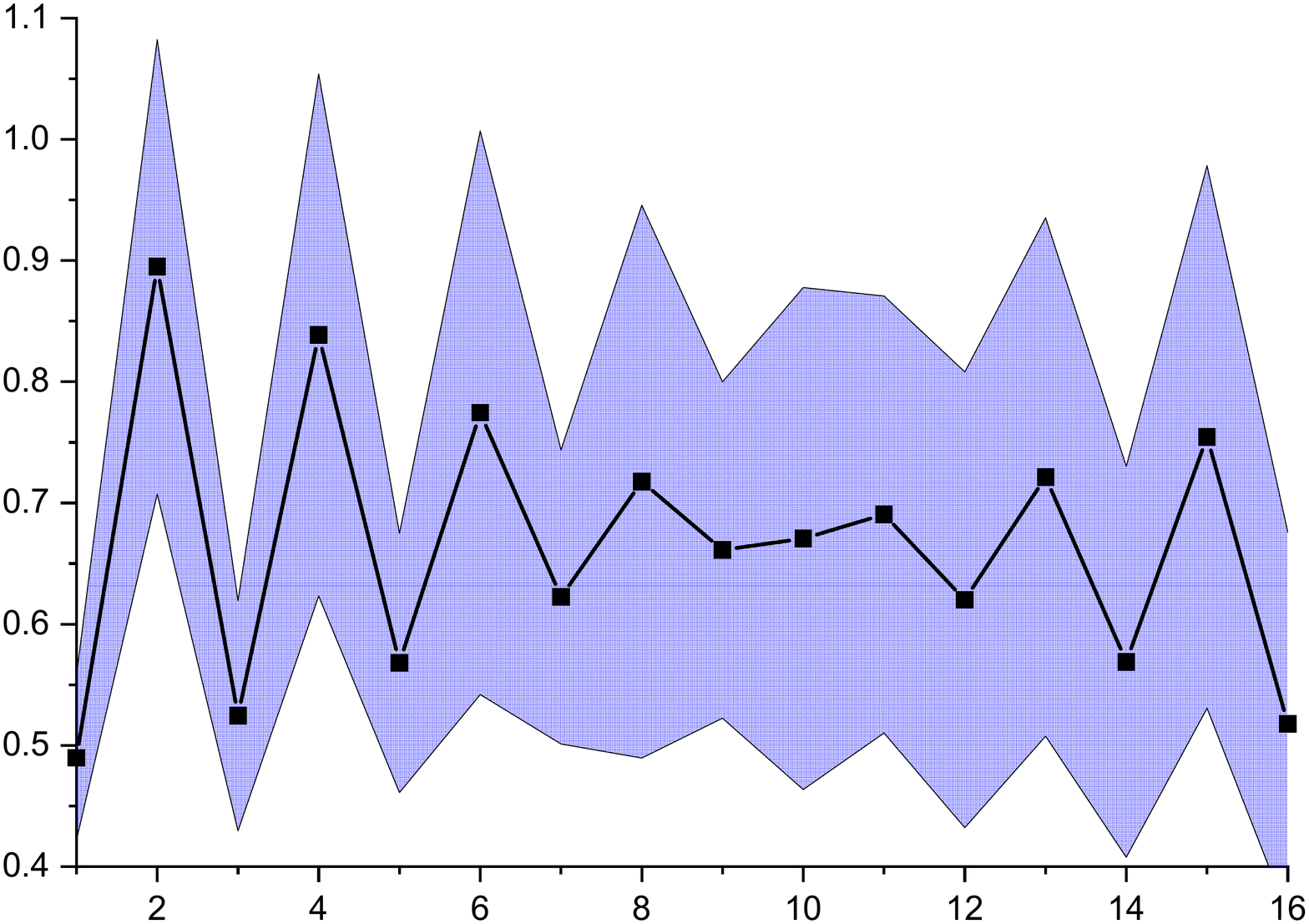}
\end{minipage}
}
\subfigure[  CW-L2]{
\begin{minipage}[b]{0.14\textwidth}
\includegraphics[width=1\linewidth]{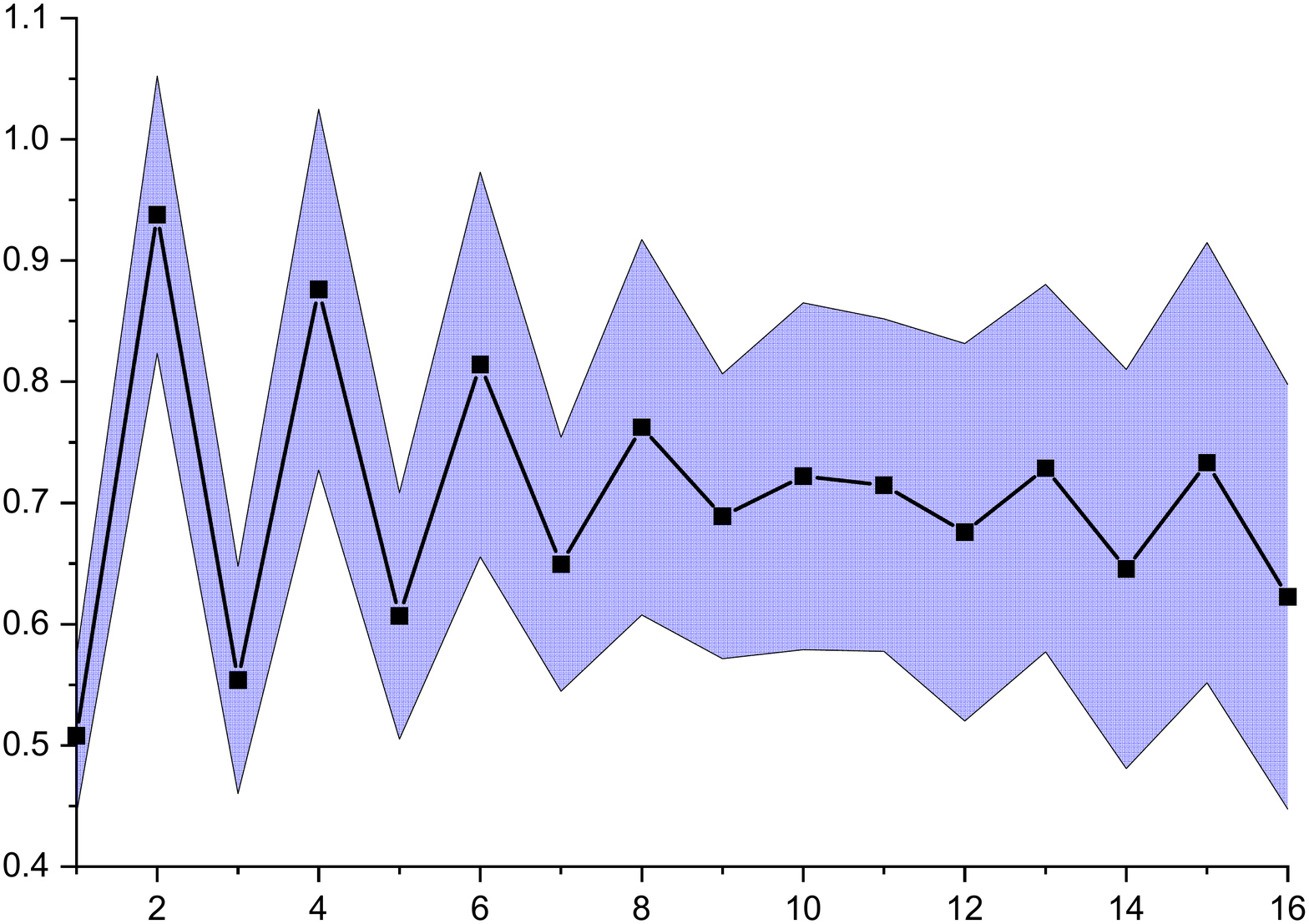}\vspace{4pt}
\includegraphics[width=1\linewidth]{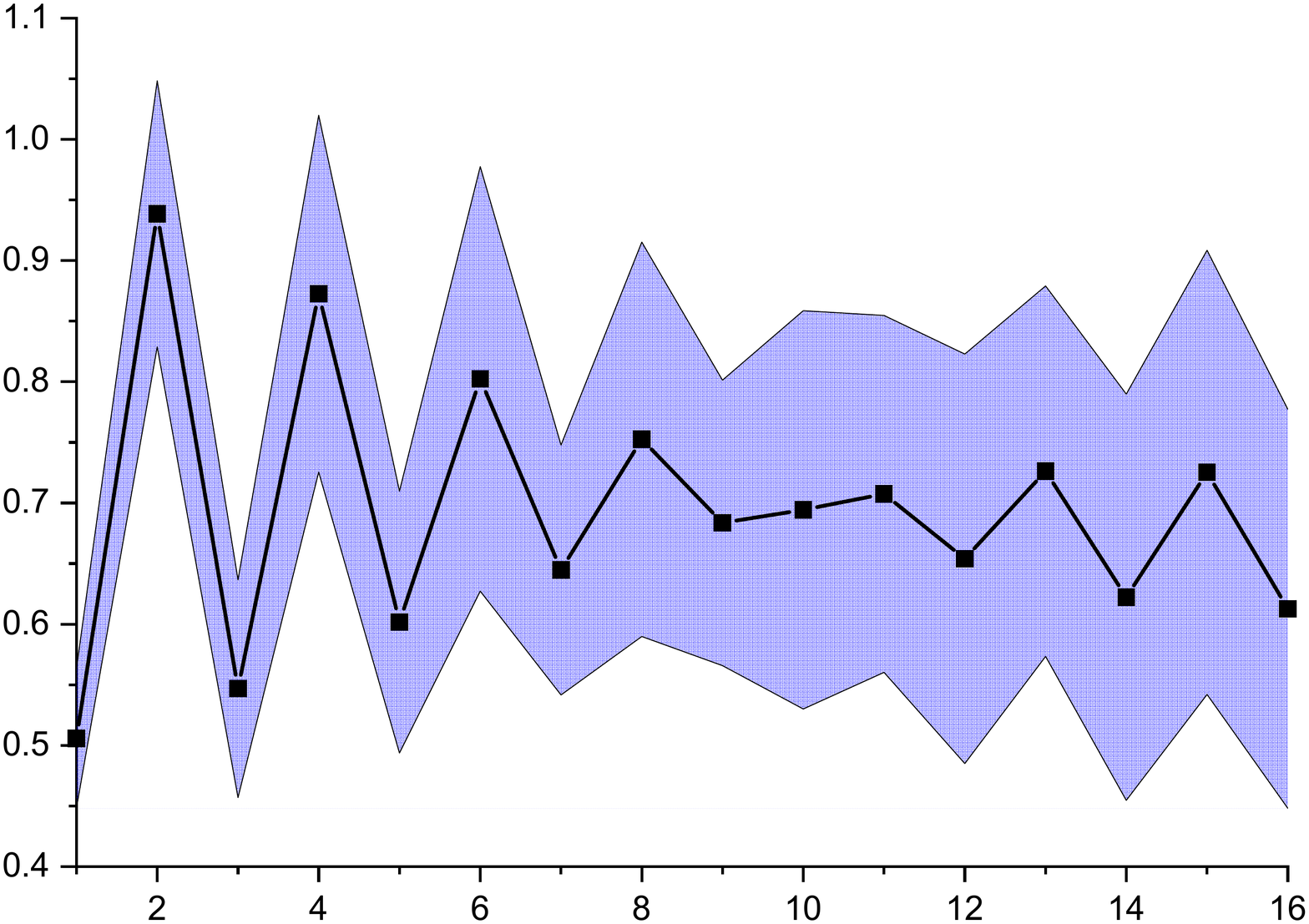}\vspace{4pt}
\includegraphics[width=1\linewidth]{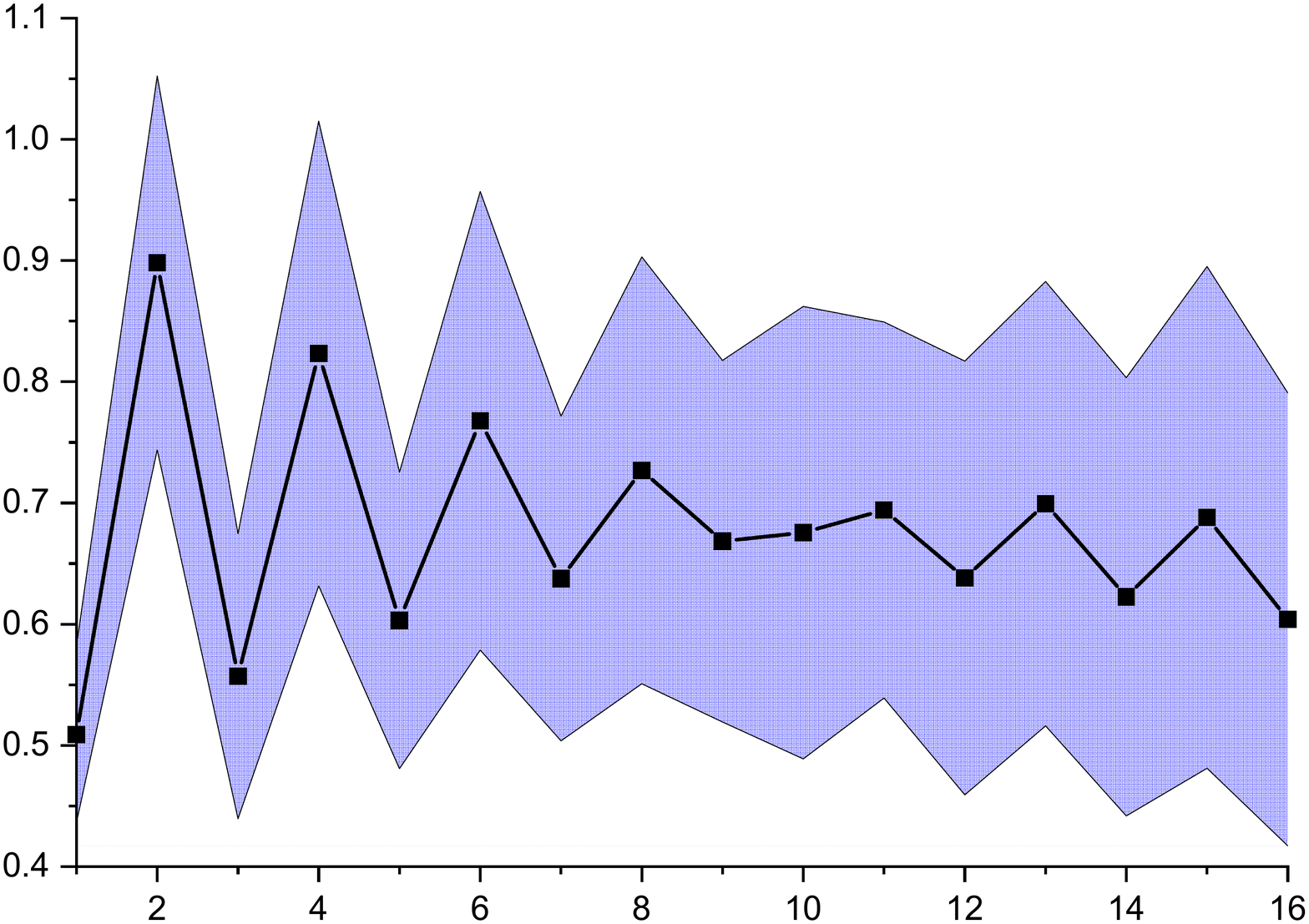}
\end{minipage}
}
\subfigure[  DeepFool]{
\begin{minipage}[b]{0.14\textwidth}
\includegraphics[width=1\linewidth]{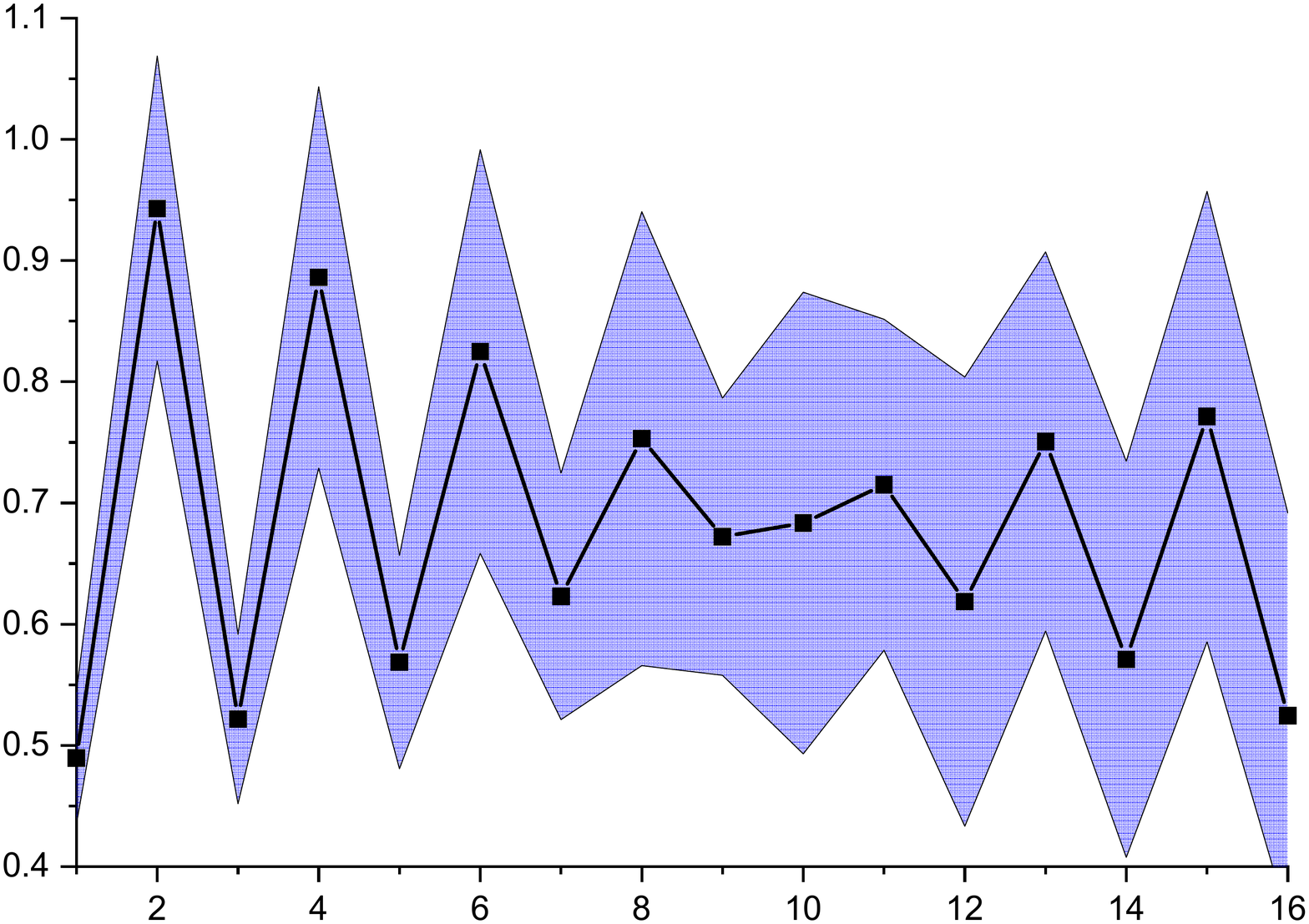}\vspace{4pt}
\includegraphics[width=1\linewidth]{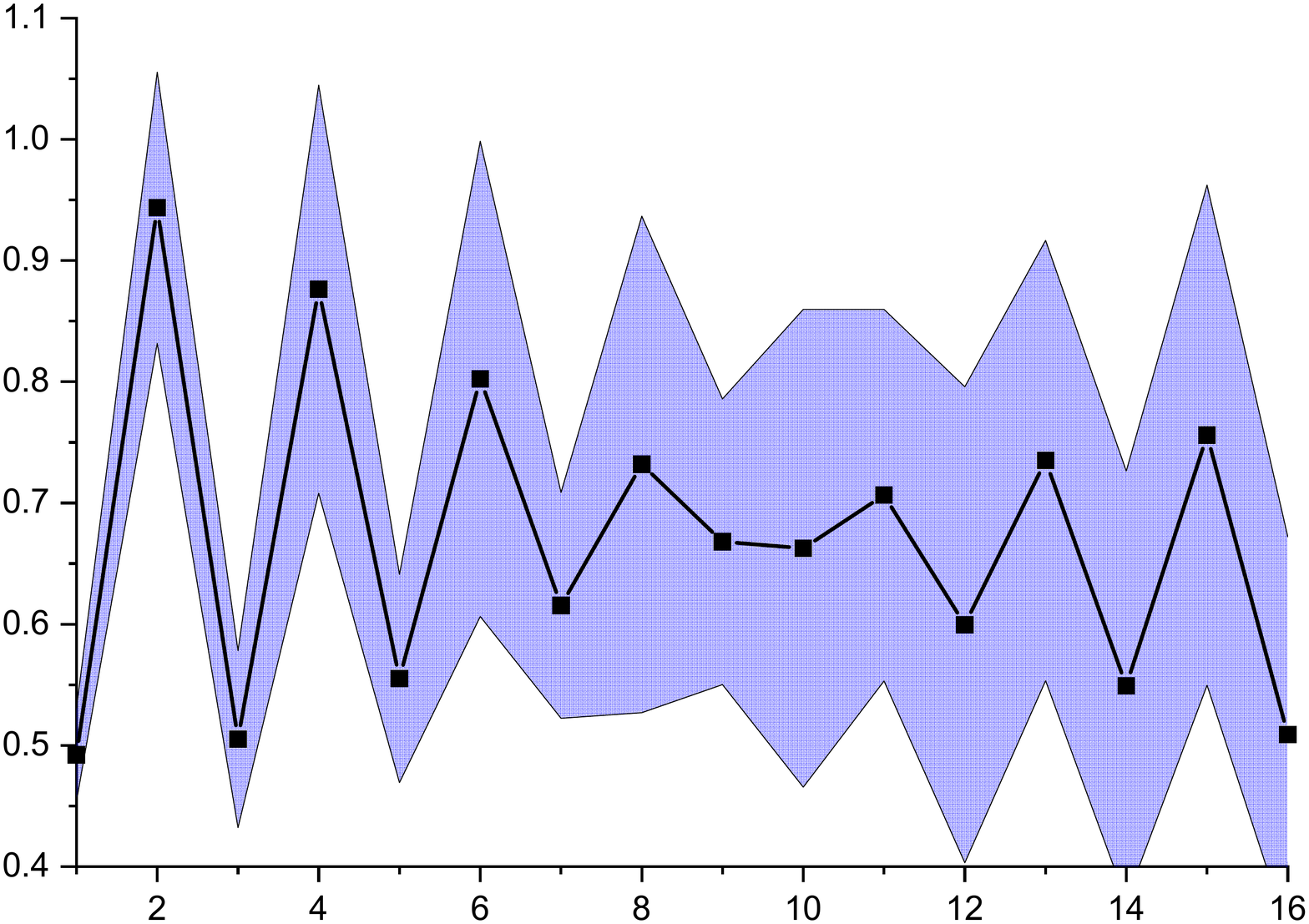}\vspace{4pt}
\includegraphics[width=1\linewidth]{figure/data-transfer-alexnet/alexnet-DeepFool.pdf}
\end{minipage}
}
\subfigure[  R-PGD]{
\begin{minipage}[b]{0.14\textwidth}
\includegraphics[width=1\linewidth]{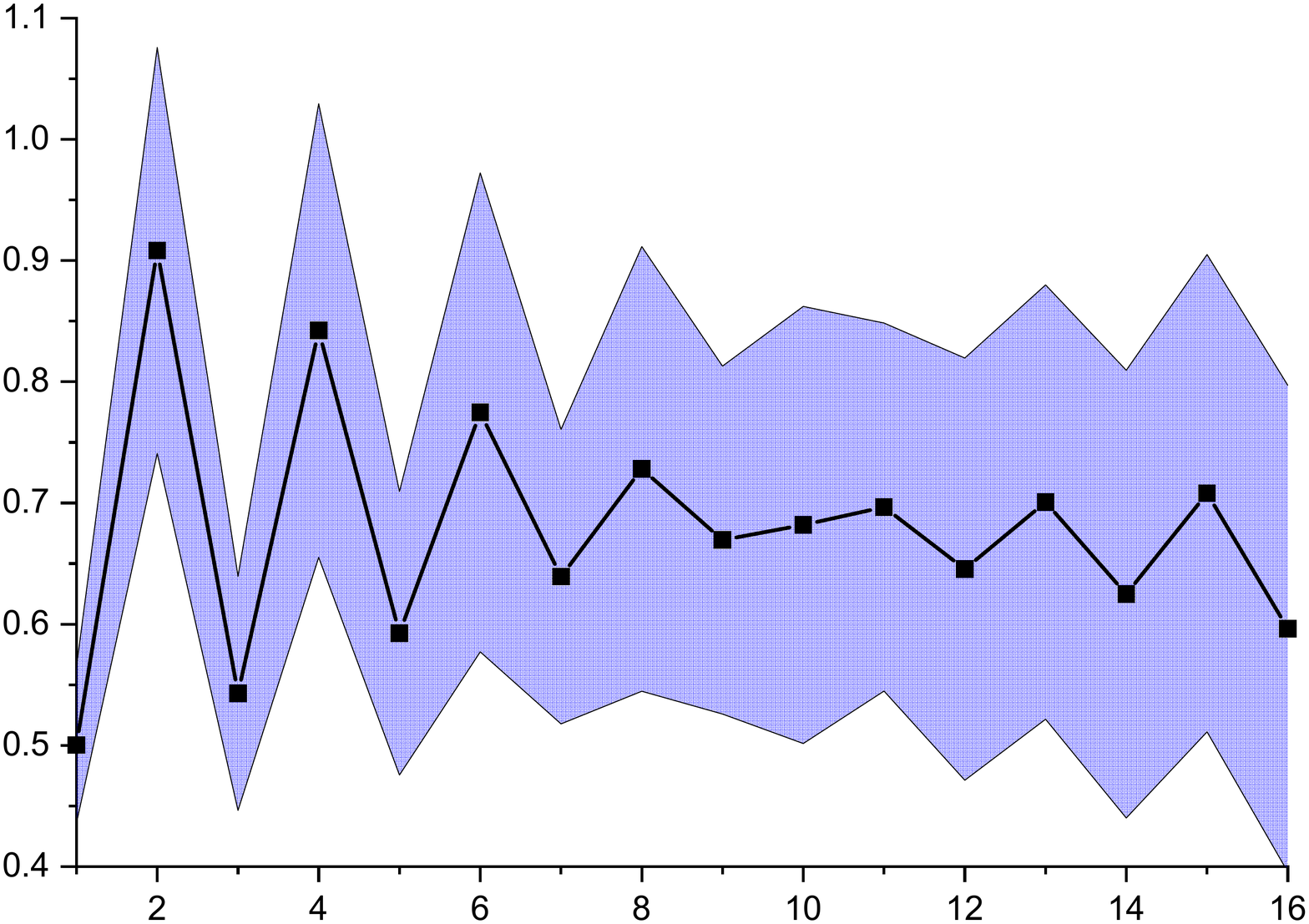}\vspace{4pt}
\includegraphics[width=1\linewidth]{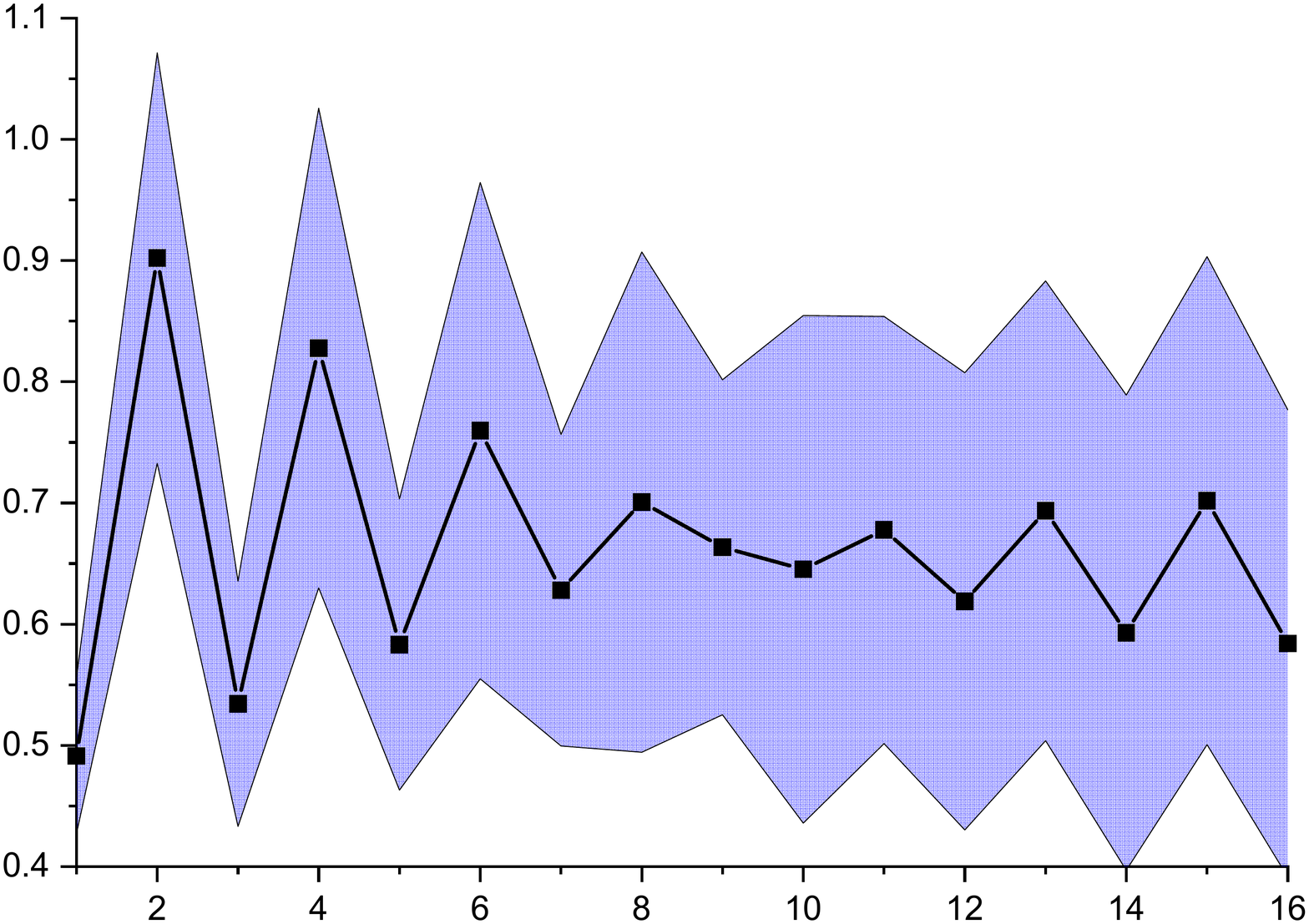}\vspace{4pt}
\includegraphics[width=1\linewidth]{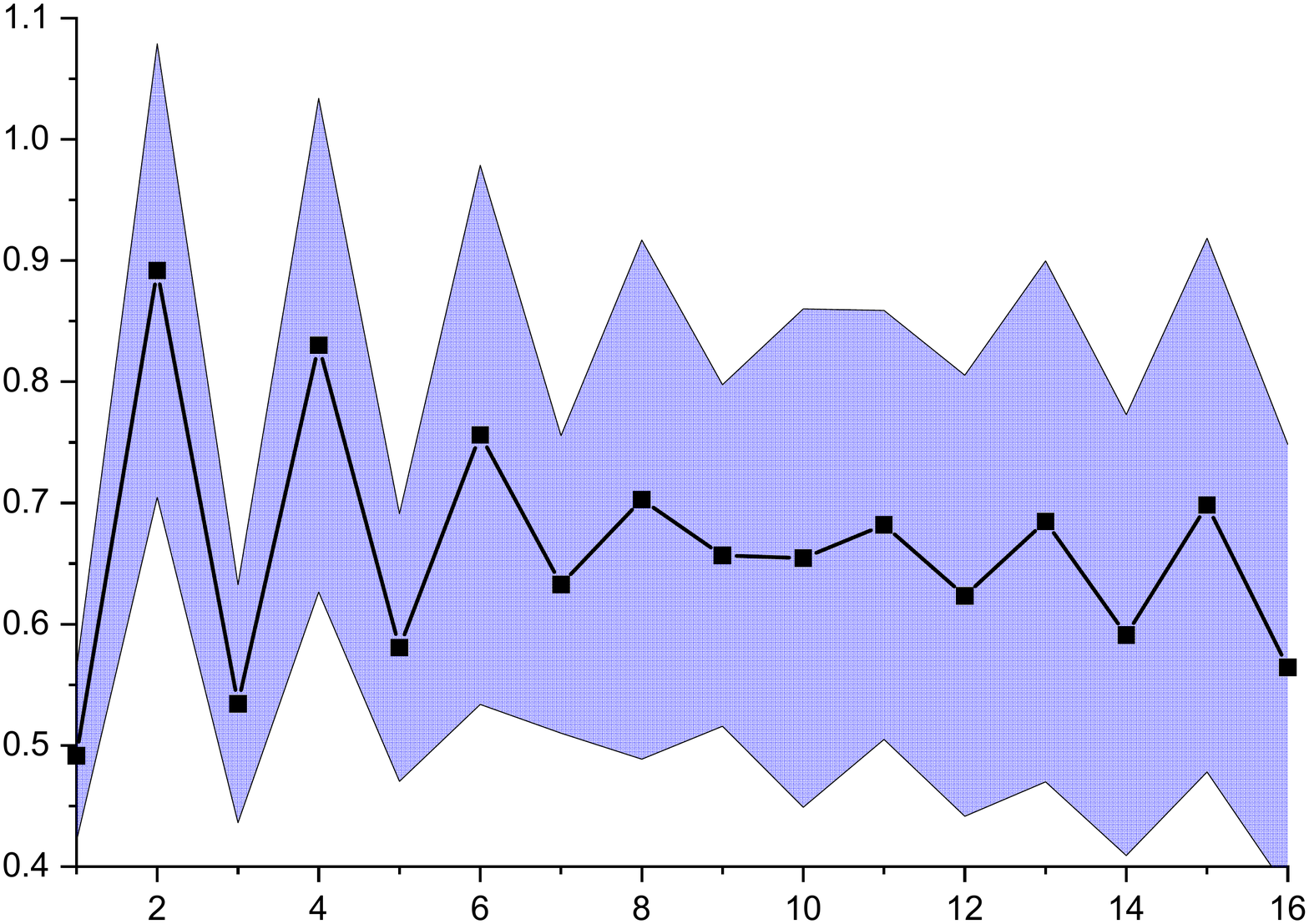}
\end{minipage}
}
}
\vspace{-3mm}
  \caption{ Statistics (mean $\pm$ standard deviation) of MBF coefficients on train (top row) and test (median row), and out-of-sample (bottom row) set of ImageNet-AlexNet. 
  }
  \label{fig:alexnet MBF}
  \vspace{-4mm}
\end{figure}

Here we present additional empirical analysis on more databases and networks, as shown in Tables \ref{Full Hypothesis test 1}, \ref{Full Hypothesis test 2.1} and \ref{Full Hypothesis test 2.2}.
The $p$-values in most cases also supports the conclusions obtained in the main manuscript.
We also present more visualizations in Figs. \ref{fig:cifar MBF}, \ref{fig:svhn MBF} and \ref{fig:alexnet MBF}.
These visualizations also demonstrate the distinct difference of MBF features between adversarial and benign examples.
\comment{
\hl{However, when the parameters of GGD are determined from the given samples, the $p$-values determined by KS test are invalid \cite{ks-test-1951}.
As a result, we have to specify the shape parameter of GGD as $2$  (}\ie  \hl{Gaussian distribution) and choose the Anderson-Darling test (AD test) \cite{AD_test} for Gaussian distribution with unknown mean and variance. The Anderson-Darling test is a modification of KS test, which used to test if a sample of data came from a population with a specific distribution. We implement AD test by the python function $statsmodels.stats.diagnostic.normal\_ad$\footnote{https://www.statsmodels.org/stable/generated/statsmodels.stats.\\diagnostic.normal\_ad.html}, which only supports the test of Gaussian distribution.
}
}
\renewcommand{\arraystretch}{1}
\begin{table*}
\centering
\caption{The $p$-value of KS hypothesis test among clean samples, noisy samples, and adversarial samples crafted by four methods, respectively.} \comment{The $p$-value of \hl{AD} hypothesis test among clean samples, noisy samples, and adversarial samples crafted by four methods, respectively. \hl{Most of the $p$-values exceed 5\%, which means we cannot reject the hypothesis that the posterior vectors of both adversarial and benign  examples follow Gaussian distribution}.
}
\begin{tabular}{>{\centering}p{0.15\textwidth} >{\centering}p{0.15\textwidth} |>{\centering}p{.08\textwidth} >{\centering}p{.08\textwidth}  >{\centering}p{.08\textwidth} >{\centering}p{.08\textwidth} >{\centering}p{.08\textwidth} >{\centering\arraybackslash}p{.08\textwidth}}   
\hline
& & clean & noisy & BIM & CW-L2 & DeepFool & R-PGD \\
\hline
\multirow{2}{*}{CIFAR-10}
&train set       &$0.074$ &$0.074$ &$0.158$ &$0.139$ &$0.155$ &$0.159$ \\
&test set        &$0.072$ &$0.072$ &$0.158$ &$0.138$ &$0.216$ &$0.216$\\
\hline
\multirow{2}{*}{SVHN}
&train set       &$0.104$ &$0.104$ &$0.249$ &$0.194$ &$0.222$ &$0.245$ \\
&test set        &$0.106$ &$0.106$ &$0.249$ &$0.193$ &$0.219$ &$0.239$\\
\hline
\multirow{2}{*}{ImageNet-AlexNet}
&train set       &$0.116$ &$0.116$ &$0.253$ &$0.235$ &$0.242$ &$0.253$ \\
&test set        &$0.117$ &$0.117$ &$0.261$ &$0.236$ &$0.243$ &$0.259$\\
\hline
\multirow{2}{*}{ImageNet-VGG-16}
&train set       &$0.114$ &$0.114$ &$0.236$ &$0.228$ &$0.240$ &$0.235$ \\
&test set        &$0.114$ &$0.114$ &$0.231$ &$0.228$ &$0.239$ &$0.232$\\  \hline
\end{tabular}
\label{Full Hypothesis test 1}
\vspace{-4mm}
\end{table*}

\renewcommand{\arraystretch}{1}
\begin{table*}[htbp]
\centering
\caption{The $p$-value of two-sample KS test among MBF coefficients of different types of examples. We cannot reject the hypothesis that the posterior vectors of both clean and non-malicious noisy images follow a same probability distribution, while clean and adversarial posterior vectors follow different probability distribution.}
\vspace{1mm}
{
\begin{tabular}{>{\centering}p{0.15\textwidth} >{\centering}p{0.09\textwidth} |>{\centering}p{.12\textwidth} >{\centering}p{.12\textwidth}  >{\centering}p{.16\textwidth} >{\centering\arraybackslash}p{.16\textwidth} }  
\hline
& & (clean, noisy) & (clean, BIM) & (BIM, DeepFool) & (CW-L2, R-PGD)  \\
\hline
\multirow{2}{*}{CIFAR-10} &train set      & $0.640$  & $0.000$  &$0.000$  &$0.003$\\
&test set       & $0.764$  & $0.000$  &$0.030$  &$0.000$\\
\hline
\multirow{2}{*}{SVHN} &train set      & $0.559$  & $0.000$  &$0.030$  &$0.000$\\
&test set       & $0.685$  & $0.000$  &$0.000$  &$0.000$\\
\hline
\multirow{2}{*}{ImageNet-AlexNet} &train set      & $1.000$  & $0.000$  &$0.002$  &$0.134$\\
&test set       & $0.993$  & $0.000$  &$0.477$  &$0.000$\\
\hline
\multirow{2}{*}{ImageNet-VGG-16} &train set      & $0.998$  & $0.000$  &$0.000$  &$0.236$\\
&test set       & $1.000$  & $0.000$  &$0.000$  &$0.225$\\
\hline
\end{tabular}}
\label{Full Hypothesis test 2.1}
\vspace{-3mm}
\end{table*}

\renewcommand{\arraystretch}{1}
\begin{table*}[htbp]
\centering
\caption{The $p$-value of KS test among MBF coefficients from different data sources. Since all the $p$-values exceed 5\%, we cannot reject the hypothesis that the posterior vectors of examples from both training dataset and testing dataset follow a same probability distribution.}
\vspace{1mm}
{
\begin{tabular}{>{\centering}p{0.16\textwidth} >{\centering}p{0.23\textwidth} |>{\centering}p{.055\textwidth} >{\centering}p{.055\textwidth}  >{\centering}p{.055\textwidth} >{\centering}p{.075\textwidth} >{\centering}p{.075\textwidth} >{\centering\arraybackslash}p{.085\textwidth} }
\hline
& & clean & noisy & BIM & CW-L2 & DeepFool & R-PGD \\ \hline
CIFAR-10& (train set, test set) & $0.120$ & $0.188$ &$0.333$ &$0.456$ &$0.584$ &$0.223$\\ \hline
SVHN& (train set, test set) & $0.494$ & $0.708$ &$0.523$ &$0.147$ &$0.491$ &$0.645$\\ \hline
\multirow{2}{*}{ImageNet-AlexNet}& (train set, test set) & $0.153$ & $0.148$ &$0.370$ &$0.605$ &$0.410$ &$0.386$\\
& (train set, out-of-sample) & $0.000$ & $0.000$ &$0.169$ &$0.101$  &$0.206$ &$0.158$\\ \hline
\multirow{2}{*}{ImageNet-VGG-16}& (train set, test set) & $0.839$ & $0.858$ &$0.368$ &$0.842$ &$0.246$ &$0.221$\\
& (train set, out-of-sample) & $0.049$ & $0.054$ &$0.094$ &$0.407$  &$0.737$ &$0.088$\\

\hline

\end{tabular}}
\label{Full Hypothesis test 2.2}
\vspace{-3mm}
\end{table*}

\renewcommand{\arraystretch}{0.95}
\begin{table*}[!t]
\centering
\caption{Comparison with Defense-GAN on detection results evaluated by AUROC score (\%) among nearly 4000 images from the test set of MNIST. The best results are highlighted in bold.}
\vspace{1mm}
{
\begin{tabular}{>{\centering}p{.28\textwidth} >{\centering}p{.28\textwidth}|>{\centering}p{.055\textwidth} >{\centering}p{.075\textwidth}>{\centering}p{.075\textwidth} >{\centering\arraybackslash}p{.085\textwidth}}
\hline
Iteration number in each GD run & Number of GD runs &BIM &CW-L2 &DeepFool &R-PGD
\\\hline
200 & 10 & 0.744 & 0.980 & 0.702 & 0.746 \\
100 & 10 & 0.689 & 0.973 & 0.650 & 0.690 \\
100 & 5  & 0.664 & 0.969 & 0.628 & 0.666 \\
100 & 2  & 0.612 & 0.959 & 0.586 & 0.614 \\\hline
\multicolumn{2}{c|}{MBF} & \textbf{1.000} & \textbf{0.991} &
\textbf{1.000} & \textbf{1.000}
\\\hline
\end{tabular}}
\label{Defense-GAN}
\vspace{0mm}
\end{table*}

\renewcommand{\arraystretch}{0.95}
\begin{table*}[!t]
\centering
\caption{Comparison with C1\&C2t/u on detection results evaluated by AUROC score (\%) among 334 images from the test set of MNIST. The best results are highlighted in bold.}
\vspace{1mm}
{
\begin{tabular}{p{.085\textwidth} >{\centering}p{.085\textwidth}>{\centering}p{.085\textwidth} >{\centering}p{.085\textwidth}>{\centering}p{.085\textwidth}| >{\centering}p{.085\textwidth}>{\centering}p{.085\textwidth} >{\centering}p{.085\textwidth}>{\centering\arraybackslash}p{.085\textwidth}}
\hline
&\multicolumn{4}{c|}{CW} & \multicolumn{4}{c}{PGD}
\\
& LR=0.001 & LR=0.01 & LR=0.03 & LR=0.1 & LR=0.001 & LR=0.01 & LR=0.03 & LR=0.1
\\\hline
C1 & 0.574 & 0.908 & 0.930 & 0.929 & 0.948 & 0.918 & 0.919 & 0.897
\\
C2t & 0.531 & 0.800 & 0.888 & 0.914 & 0.858 & 0.749 & 0.553 & 0.528
\\
C2u & 0.609 & 0.769 & 0.765 & 0.769 & 0.925 & 0.865 & 0.879 & 0.846
\\
C1\&C2t/u & 0.554 & 0.810 & 0.889 & 0.912 & 0.961 & 0.906 & 0.938 & 0.926
\\\hline
MBF & \textbf{0.892} & \textbf{0.995} & \textbf{0.995} & \textbf{0.994} & \textbf{0.992} & \textbf{0.976} & \textbf{0.977} & \textbf{0.976}
\\\hline
\end{tabular}}
\label{C12}
\vspace{0mm}
\end{table*}

\renewcommand{\arraystretch}{0.95}
\begin{table*}[!t]
\centering
\caption{Comparison on running time (ms) of crafting features of a single image among all compared methods and MBF.}
\vspace{1mm}
{
\begin{tabular}{>{\centering}p{0.17\textwidth} >{\centering}p{.10\textwidth}>{\centering}p{.10\textwidth} >{\centering}p{.10\textwidth}>{\centering}p{.15\textwidth} >{\centering}p{.10\textwidth}>{\centering\arraybackslash}p{.10\textwidth}}
\hline
&KD+BU &M-D &LID &Defense-GAN &C1\&C2t/u &MBF
\\\hline
CIFAR-10         & 11.3  & 16.7 & 110.8 & -   & -    & 369.0 \\
SVHN             & 13.6  & 72.9 & 241.6 & -   & -    & 322.1 \\
ImageNet-AlexNet & 150.9 & 1.4  & 96.0  & -   & -    & 234.3 \\
ImageNet-VGG-16  & 318.9 & 4.4  & 241.3 & -   & -    & 486.5 \\
MNIST            & -     & -    & -     & 590 & -    & 97.0 \\
CIFAR-10-VGG-19  & -     & -    & -     & -   & 576  & 380.2 \\
\hline
\end{tabular}}
\label{complexity}
\vspace{0mm}
\end{table*}

\section{More Comparison Experiments}

\subsection{Comparison with Defense-GAN}

Defense-GAN \cite{defensegan-iclr-2018} has been recently proposed to detect and restore adversarial examples.
Different with other compared methods in the main manuscript which directly extracted the detection features from the intermediate activation, Defense-GAN should firstly train a dataset-specific generative adversarial network (GAN). Then, the norm of the difference between the image generated by GAN and the detected image is computed as the detection feature. The image is considered as adversarial if the norm is large than a given threshold.
We adopt official code\footnote{
\text{https://github.com/kabkabm/defensegan/}
} of Defense-GAN  and train an generative model on train set of MNIST after $133,500$ iterations with the mini-batch size (note that the official code of Defense-GAN only supports three datasets, including MNIST, Fashion-MNIST, and CelebA). The trained model achieves an accuracy of 98.7\% on the test set of MNIST. We still utilize four attack methods, including BIM, CW-L2, DeepFool and R-PGD. After removing images that are misclassified and failed to attack, we pick $6000$ images to train SVMs based on our MBF features, and leave nearly $4000$ images for testing. There are two key hyper-parameters in Defense-GAN, including the iteration number in each gradient descent (GD) run, and the total number of GD runs. We have tried multiple settings of this two hyper-parameters and report the best results. The AUROC scores on nearly $4000$ test images are shown in Table \ref{Defense-GAN}. Our MBF method exceeds Defense-GAN by a large margin on quantitative performance.

\subsection{Comparison with C1\&C2t/u}

A recent paper called C1\&C2t/u \cite{NIPS2019_turning} discussed the property of adversarial perturbations around benign examples, and came up with two paradoxical criteria for detection. It is assumed that the adversarial example cannot satisfy this two criteria simultaneously, while the benign example could them both.
We also adopt official code\footnote{
\text{https://github.com/s-huu/TurningWeaknessIntoStrength/}
} of C1\&C2t/u. The pre-trained classification neural network is VGG-19 which achieves $93.4\%$ accuracy on 1000 images picked from CIFAR-10 \cite{cifar10}. 600 out of 934 correctly classified images are randomly chosen as training images for detection, and the remaining 334 images are used as testing images.
Besides, the learning rate (LR) in crafting white-box adversarial examples significantly affects the detection performance of C1\&C2t/u. Different values of LR are tested, and the corresponding results are reported in Table \ref{C12}. Note that C1, C2, C2u are different variants of C1\&C2t/u. Please refer to \cite{NIPS2019_turning} for details.
Our MBF method shows much better detection performance than C1\&C2t/u in all cases.

\section{Detailed Parameters of Adversarial Attack}

Four popular adversarial attack methods are adopted to craft adversarial examples, including basic iterative method (BIM), CarliniWagnerL2Attack (CW-L2), DeepFool, and random projected gradient descent (R-PGD). We emphasize detailed parameters of these attack strategies for reproducibility, which are
BIM (eps=$0.3$, stepsize=$0.05$, iterations=$10$),
CW-L2 (binary\_search\_steps=$5$, confidence=$0.0$, learning\_rate=$0.005$, max\_iterations=$1000$),
DeepFool (max\_steps=$100$),
R-PGD (eps=$0.3$, stepsize=$0.01$, iterations=$40$). All of the attack strategies are implemented by Foolbox.

\section{Computational Complexity}
{\bf Crafting discriminative feature}
Compared with KD+BU \cite{KD-BU-arxiv-2017}, LID \cite{LID-iclr-2018}, and M-D \cite{M-distance-nips-2018}, our MBF method requires no neighboring samples to craft discriminative feature for each input example. 
Recalling Eq. (8) in the main manuscript, the time complexity of crafting feature $\hat{\mathbf{a}}$ is $O(N)$, where $N=T\sum_{l=1}^{L}d_l$, where $d_l$ denotes the number of response entries in $l$-th layer, and $T$ indicates the demension of the extracted MBF features.
In contrast, the main bottleneck of KD+BU lies in the computation of  KernelDensity\footnote{https://scikit-learn.org/stable/modules/generated/sklearn\\.neighbors.KernelDensity.html}, which requires the internal response of the whole train set. As for LID, it describes the distance between one example and its k-nearest neighboring samples in the $d_l$-dim feature space of intermediate responses. The KNN algorithm can be quite time-consuming. In terms of M-D, it computes the class-conditional Gaussian distribution of the responses based on the whole train set, so the computation of mean vector and covariance matrix of the whole train set burdens most.

In terms of Defense-GAN \cite{defensegan-iclr-2018}, the time complexity of generate an virtual image is $O(LR)$, with $L$ being the iteration number in each gradient descent (GD) run, and $R$ being the total number of GD runs.
C1\&C2t/u turns out to be the slowest algorithm, since both C1 and C2 need to add perturbation to the input image iteratively until misclassified. Both feed-forward and back-forward propagation are executed during each iteration step.

The running time of our method and all competing methods are shown in Table \ref{complexity}.
Note that some entries in the table are missing, as the corresponding experiments are not conducted. For example, Defense-GAN is only evaluated on MNIST, while other databases are not adopted.

{\bf Testing with SVM}
Since we test each sample with a RBF-SVM, the computational complexity only depends on the dimension of input discriminative feature $d$, which is $O(d^2)$, as discussed in \cite{SVM_predict}. As a result, the running time of MBF detection is a little longer than other compared methods, because the feature dimension of MBF is $TL$, while other methods utilize a $L$-dim vector or even a scalar as feature. However, we have tested that the  difference is up to $0.05$ second and can be negligible in practice.

\nocite{langley00}

\comment{
\appendix
\section{}  
}
\comment{
\textbf{\emph{Do not put content after the references.}}
Put anything that you might normally include after the references in a separate
supplementary file.

We recommend that you build supplementary material in a separate document.
If you must create one PDF and cut it up, please be careful to use a tool that
doesn't alter the margins, and that doesn't aggressively rewrite the PDF file.
pdftk usually works fine.

\textbf{Please do not use Apple's preview to cut off supplementary material.} In
previous years it has altered margins, and created headaches at the camera-ready
stage.
}

\end{document}